\documentclass[preprint,11pt]{elsarticle}

\usepackage[top=0.75in,bottom=0.75in,right=0.75in,left=0.75in]{geometry}
\usepackage{hyperref}
\newcommand{\textot}[1]{\scalebox{0.95}{\texttt{#1}}}

\usepackage{enumitem}
\usepackage{scrextend}

\usepackage{pgfplots,xifthen,bm,tikz,tikz-cd,adjustbox,afterpage,algorithmic}
\usetikzlibrary{calc, positioning}
\usepackage{array}
\usepackage{eqparbox}

\usepackage{etoolbox}  
\makeatletter
\patchcmd{\algorithmic}{\addtolength{\ALC@tlm}{\leftmargin} }{\addtolength{\ALC@tlm}{\leftmargin}}{}{}
\newcommand\footnoteref[1]{\protected@xdef\@thefnmark{\ref{#1}}\@footnotemark}
\makeatother
\usetikzlibrary{patterns}
\usepackage{pgfplots}
\usepackage{float}
\usepackage{graphicx}

\definecolor{green1}{HTML}{004c00}
\definecolor{green2}{HTML}{009900}
\definecolor{green3}{HTML}{47da47}
\definecolor{green4}{HTML}{99ea99}
\definecolor{blue1}{HTML}{003566}
\definecolor{blue2}{HTML}{186fc0}
\definecolor{blue3}{HTML}{68aae7}
\definecolor{blue4}{HTML}{a1caf0}
\definecolor{red0}{HTML}{E74C4C}
\definecolor{red1}{HTML}{990000}
\definecolor{red2}{HTML}{DD0000}
\definecolor{red3}{HTML}{E74C4C}
\definecolor{red4}{HTML}{F19999}
\definecolor{yellow1}{HTML}{F1C239}
\definecolor{gray1}{HTML}{C8C8C8}
\definecolor{gray2}{HTML}{707070}
\definecolor{gray3}{HTML}{505050}
\definecolor{brown1}{HTML}{654321}

\usepackage{caption}
\usepackage{subcaption}
\usepackage{multirow}
\usepackage{booktabs}

\usepackage{amsmath}
\usepackage{arydshln}

\usepackage{pgfplotstable}

\usepackage[ruled,vlined]{algorithm2e} 
\usepackage{amsmath}
\usepackage{amsfonts}
\SetAlFnt{\footnotesize}
\SetKwInput{KwInit}{Initialization}

\usepackage{subfiles}

\newrobustcmd*{\mycircle}[1]{\tikz{\filldraw[draw=#1,fill=#1] (0,0) circle [radius=0cm] (0,0.1cm) circle [radius=0.06cm];}}

\newrobustcmd*{\mytriangle}[1]{\tikz{\filldraw[draw=#1,fill=#1] (0,0) circle [radius=0cm] (0,0.05cm) -- (0.1cm,0.05cm) -- (0.05cm,0.15cm);}}

\journal{Pattern Recognition}

\makeatletter
\def\@author#1{\g@addto@macro\elsauthors{\normalsize%
		\def\baselinestretch{1}%
		\upshape\authorsep#1\unskip\textsuperscript{%
			\ifx\@fnmark\@empty\else\unskip\sep\@fnmark\let\sep=,\fi
			\ifx\@corref\@empty\else\unskip\sep\@corref\let\sep=,\fi
		}%
		\def\authorsep{\unskip,\space}%
		\global\let\@fnmark\@empty
		\global\let\@corref\@empty  
		\global\let\sep\@empty}%
	\@eadauthor={#1}
}
\makeatother

\begin{document}

\begin{frontmatter}

\title{Instance exploitation for learning temporary concepts from sparsely labeled drifting data streams}


\author{\L{}ukasz Korycki\corref{mycorrespondingauthor}}
\ead{koryckil@vcu.edu}
\cortext[mycorrespondingauthor]{Corresponding author}

\author{Bartosz Krawczyk}
\ead{bkrawczyk@vcu.edu}

\address{Department of Computer Science, Virginia Commonwealth University, Richmond VA, USA}

\begin{abstract}
Continual learning from streaming data sources becomes more and more popular due to the increasing number of online tools and systems. Dealing with dynamic and everlasting problems poses new challenges for which traditional batch-based offline algorithms turn out to be insufficient in terms of computational time and predictive performance. One of the most crucial limitations is that we cannot assume having access to a finite and complete data set -- we always have to be ready for new data that may complement our model. This poses a critical problem of providing labels for potentially unbounded streams. In the real world, we are forced to deal with very strict budget limitations, therefore, we will most likely face the scarcity of annotated instances, which are essential in supervised learning. In our work, we emphasize this problem and propose a novel instance exploitation technique. We show that when: (i) data is characterized by temporary non-stationary concepts, and (ii) there are very few labels spanned across a long time horizon, it is actually better to risk overfitting and adapt models more aggressively by exploiting the only labeled instances we have, instead of sticking to a standard learning mode and suffering from severe underfitting. We present different strategies and configurations for our methods, as well as an ensemble algorithm that attempts to maintain a sweet spot between risky and normal adaptation. Finally, we conduct a complex in-depth comparative analysis of our methods, using state-of-the-art streaming algorithms relevant to the given problem.
\end{abstract}


\end{frontmatter}

\section{Introduction}
\label{sec:int}

Contemporary data has moved beyond being small and static collections of information. Nowadays, we need to tackle not only the massive volume of data sets, but also their velocity, as new data may arrive anytime, continuously flooding the system \cite{Ditzler:2015}. If we focus only on storing the information and analyzing it in an offline fashion, then we may face the problem known as data tombs \cite{Wren:2008}. This term is used to describe enormous data collections, holding potentially valuable knowledge that is never analyzed due to its size and time constraints. Even if we alleviated the data tomb problem by using high-performance computing environments, we would still be introducing a delay between the moment when information arrived and when it was analyzed. In such a setting, we cannot respond immediately and when we finish the analysis of our data collection it may turn out to be outdated.  In order to extract the most current and valuable knowledge modern data must be analyzed in (near) real-time. The most recent example of such dynamic data is the information about the spread of COVID-19 \cite{Staszkiewicz:2020}. Here, we are dealing with modeling a completely new event that at the given stage offers limited information. Any new data is therefore of extremely high value and must be incorporated into the predictive system as soon as possible. The underlying learning model must be capable of flexible adaptation, as even a small amount of data gathered in the last couple of hours may completely invalidate previously made predictions. COVID-19 showed why predictive models should be updated in an online manner and how predicted dynamics may change rapidly whenever a new batch of data becomes available \cite{Rustam:2020}. When dealing with such an impactful event, one cannot delay obtaining vital conclusions by using offline models. 

Such problems gave a rise to the field of data stream mining. Here, we assume that new information arrives continuously and our learning model must be capable of making quick decisions and incorporating new data on-the-fly. Streaming algorithms continuously update themselves in order to capture the most current state of a stream and its dynamics \cite{Cano:2019}. This is necessary, as data streams are subject to a phenomenon known as concept drift \cite{Lu:2019}. It assumes that the distribution and characteristics of analyzed data will change over time, generating temporarily valid concepts and rendering previously trained models irrelevant. When a drift occurs, old information stored in the model should be discarded and learning over a new concept should be prioritized. Due to the varying nature of concept drift, it may affect different properties of data (feature distribution, class labels, or proportion of instances in classes) under different rapidity of changes (sudden or gradual). These factors must dictate a way in which the model will adapt to drifts, making understanding the dynamics of change a vital task \cite{Goldenberg:2019, Goldenberg:2020}. 

Data stream mining poses additional challenges, other than creating accurate predictive models. We must assume working under time and memory constraints. Designing machine learning methods for data streams can be viewed as a multi-criteria optimization, where one aims at reaching a trade-off between several factors that define efficacy in streaming environments \cite{Krawczyk:2017}. As we cannot store the entire data stream, we should carefully choose which ones to keep and which ones to discard (or, in the case of online learning, assume that each instance must be discarded after a single pass). The trained model must quickly provide predictions for new arriving instances to avoid bottlenecks \cite{Ramirez-Gallego:2017}. Furthermore, when faced with a concept drift, we need not only to detect it as soon as possible but also to adapt the model to the new state of the stream with the lowest possible latency. The time between the occurrence of the concept drift and classifier re-training is known as a drift recovery rate \cite{Shaker:2015}. 

Reducing the time when our classifier is incompetent (i.e., not adapted to the new concept) should be seen as one of the main goals of learning from data streams. The speed of adaption can be affected by the training procedure of the classifier, as well as by the access to data from the new concept -- the more instances representative to the new concept we have, the more likely we are to reduce the recovery time. However, we must be aware that a fully supervised learning from data streams is an unrealistic scenario. It would require access to an oracle providing class labels for every single instance in the stream. As such labels usually come from a domain expert, we cannot assume that we can obtain all of them -- we are restricted by both the budget (i.e., time of the domain expert is costly) and human limitations (instances arrive with such a speed and in such a volume that is impossible for a human to tackle). A realistic scenario for data stream mining assumes either limited access to ground truth \cite{Zliobaite:2014}, or delayed labeling \cite{Plasse:2016}. This restricted access to labeled data strengthens another problem known as underfitting, making effective drift adaptation even more challenging.

In this paper, we address an extremely important challenge of data stream mining: how to improve learning from drifting data while having only strictly limited access to class labels. While many of the existing works focus on how to reasonably choose instances for labeling, in this work, we aim at providing further improvements based solely on the few labeled instances given to us and at no additional cost. Our assumption is that any obtainable improvement under strict labeling constraints is vital for the underlying classifier, offering faster adaptation to new concepts. The main contributions of our work include the following.

\begin{itemize}
\item \textbf{Enhancing drift adaptation with instance exploitation.} Our idea is to intensively exploit labeled instances at our disposal, aiming at reinforcing adaptation to non-stationary concepts and providing a faster classifier recovery. By reusing the labeled instances we should be able to avoid underfitting and offer faster and more accurate reactions to concept drift. 

\item \textbf{Exploitation techniques for improved drift adaptation.} We introduce three techniques for instance exploitation to empower active learning from sparsely labeled drifting data streams. They are based on a reactive sliding window that uses one of three probabilistic sampling techniques to select instances for a more aggressive exposure to the online classifier. 

\item \textbf{Ensemble architectures to avoid overfitting.} In order to minimize the risk of overfitting, we propose two simple, yet effective ensemble architectures. They are based on two paired learners, one preferring aggressive instance exploitation and the other learning in a standard way. Our architectures are capable of dynamic switching between these models, with an added procedure for improving the weaker of the two learners. 

\item \textbf{Flexible framework.} Our proposal is a universal wrapper that can be used to enhance any online active learning algorithm dedicated to data streams.

\item \textbf{Insights into the role of enhanced drift adaptation.} We offer an exhaustive experimental study on both artificial and real data streams, paired up with in-depth analysis that offers unique insights and better understanding of how to efficiently improve the adaptation to concept drift by using already labeled instances without any additional labeling costs. 
\end{itemize}

\section{Learning from streams}
\label{sec:streams}

Data stream is defined as a sequence ${<S_1, S_2, ..., S_n,...>}$, where each element $S_j$ is a new instance. In this paper we assume the (partially) supervised learning scenario with classification task and thus we will define each instance as $S_j \sim p_j(x^1,\cdots,x^d,y) = p_j(\mathbf{x},y)$, where $p_j(\mathbf{x},y)$ is a joint distribution of $j$-th instance, defined by a $d$-dimensional feature space and assigned to class $y$. Each instance is independent and drawn randomly from a probability distribution $\Psi_j (\mathbf{x},y)$.

\medskip
\noindent \textbf{Concept drift.} When all instances come from the same distribution, we deal with a stationary data stream. In real-world applications data very rarely falls under stationary assumptions \cite{Masegosa:2020}. It is more likely to evolve over time and form temporary concepts, being subject to concept drift \cite{Lu:2019}. This phenomenon affects various aspects of a data stream and thus can be analyzed from multiple perspectives. One cannot simply claim that a stream is subject to the drift. It needs to be analyzed and understood in order to be handled adequately to specific changes that occur \cite{Goldenberg:2019,Goldenberg:2020}. More precise approaches may help us achieving faster and more accurate adaptation \cite{Shaker:2015}. Let us now discuss the major aspects of concept drift and its characteristics. 

\medskip
\noindent \textbf{Influence on decision boundaries}. Firstly, we need to take into account how concept drift impacts the learned decision boundaries, distinguishing between real and virtual concept drifts \citep{Oliveira:2019}. The former influences previously learned decision rules or classification boundaries, decreasing their relevance for newly incoming instances. Real drift affects posterior probabilities $p_j(y|\mathbf{x})$ and additionally may impact unconditional probability density functions. It must be tackled as soon as it appears since it impacts negatively the underlying classifier. Virtual concept drift affects only the distribution of features $\mathbf{x}$ over time:

\begin{equation}
\widehat{p}_j(\mathbf{x}) = \sum_{y \in Y} p_j(\mathbf{x},y),
\label{eq:cd2}
\end{equation}
\smallskip

\noindent where $Y$ is a set of possible values taken by $S_j$. While it seems less dangerous than real concept drift, it cannot be ignored. Despite the fact that only the values of features change, it may trigger false alarms and thus force unnecessary and costly adaptations. 

\medskip
\noindent \textbf{Locality of changes}. It is important to distinguish between global and local concept drifts~\citep{Gama:2006}. The former one affects the entire stream, while the latter one affects only certain parts of it (e.g., regions of the feature space, individual clusters of instances, or subsets of classes). Determining the locality of changes is of high importance, as rebuilding the entire classification model may not be necessary. Instead, one may update only certain parts of the model or sub-models, leading to a more efficient adaptation.

\medskip
\noindent \textbf{Speed of changes}. Here we distinguish between sudden, gradual, and incremental concept drifts~\cite{Lu:2019}. 

\begin{itemize}

\medskip
\item \textbf{Sudden concept drift} is a case when instance distribution abruptly changes with $t$-th example arriving from the stream:

\begin{equation}
p_j(\mathbf{x},y) =
  \begin{cases}
   D_0 (\mathbf{x},y),       & \quad \text{if } j < t\\
    D_1 (\mathbf{x},y),  & \quad \text{if } j \geq t.
  \end{cases}
\label{eq:cd3}
\end{equation}

\medskip
\item \textbf{Incremental concept drift} is a case when we have a continuous progression from one concept to another (thus consisting of multiple intermediate concepts in between), such that the distance from the old concept is increasing, while the distance to the new concept is increasing:

\begin{equation}
p_j(\mathbf{x},y) =
  \begin{cases}
  D_0 (\mathbf{x},y),       &  \text{if } j < t_1\\
   (1 - \alpha_j) D_0 (\mathbf{x},y) + \alpha_j D_1 (\mathbf{x},y),       &\text{if } t_1 \leq j < t_2\\
    D_1 (\mathbf{x},y),  &  \text{if } t_2 \leq j
  \end{cases}
\label{eq:cd4}
\end{equation}

\noindent where

\begin{equation}
\alpha_j = \frac{j - t_1}{t_2 - t_1}.
\label{eq:cd5}
\end{equation}

\medskip
\item \textbf{Gradual concept drift} is a case where instances arriving from the stream oscillate between two distributions during the duration of the drift, with the old concept appearing with decreasing frequency:

\begin{equation}
p_j(\mathbf{x},y) =
  \begin{cases}
   D_0 (\mathbf{x},y),       &  \text{if } j < t_1\\
   D_0 (\mathbf{x},y),       &  \text{if } t_1 \leq j < t_2 \wedge \delta > \alpha_j\\
   D_1 (\mathbf{x},y),       &  \text{if } t_1 \leq j < t_2 \wedge \delta \leq \alpha_j\\
    D_1 (\mathbf{x},y),  &  \text{if } t_2 \leq j,
  \end{cases}
\label{eq:cd4}
\end{equation}

\noindent where $\delta \in [0,1]$ is a random variable. 

\end{itemize}

\noindent \textbf{Recurrence}. In many scenarios it is possible that a previously seen concept from $k$-th iteration may reappear D$_{j+1}$ = D$_{j-k}$ over time \cite{Sobolewski:2017}. One may store models specialized in previously seen concepts in order to speed up recovery rates after a known concept reemerges \cite{Guzy:2020}. 

\medskip
\noindent \textbf{Presence of noise}. Apart from concept drift, one may encounter other types of changes in data. They are connected with the potential appearance of incorrect information in the stream and known as blips or noise. The former stands for singular random changes in a stream that should be ignored and not mistaken for a concept drift. The latter stands for significant corruption in the feature values or class labels and must be filtered out in order to avoid feeding false \cite{Krawczyk:2018} or even adversarial information to the classifier \cite{Sethi:2018}. 

\medskip
\noindent \textbf{Feature drift}. This is a type of change that happens when a subset of features becomes, or stops to be, relevant to the learning task \cite{Barddal:2017}. Additionally, new features may emerge (thus extending the feature space), while the old ones may cease to arrive. 

\medskip
\noindent \textbf{Classifiers for drifting data streams.} In order to be able to adapt to evolving data streams, classifiers must either have explicit information on when to update their model, or use continuous learning to follow the progression of a stream. Concept drift detectors are external tools that can be paired with any classifier and used to monitor a state of the stream \cite{Barros:2018}. Usually, this is based on tracking the error of the classifier \cite{Pinage:2020} or measuring the statistical properties of data \cite{Korycki:2019}. Drift detectors emit two-level signals. The warning signal means that changes start to appear in the stream and recent instances should be stored in a dedicated buffer. This buffer is used to train a new classifier in the background. The drift signal means that changes are significant enough to require adaptation and the old classifier must be replaced with the new one trained on the most recent buffer of data. This reduces the cost of adaptation by lowering the number of times when we train the new classifier, but may be subject to costly false alarms or missing changes appearing locally or on a smaller magnitude. Alternative approaches assume using classifiers that are capable of learning in an incremental or online manner. Sliding windows storing only the most recent instances are very popular, allowing for natural forgetting of older instances \cite{Ramirez-Gallego:2017tsmc,Roseberry:2019}. The size of the window is an important parameter and adapting it over time seems to yield the best results \cite{Bifet:2007}. Online learners are capable of learning instance by instance, discarding data after it passed the training procedure. They are very efficient on their own, but need to be equipped with a forgetting mechanism in order not to endlessly grow their complexity \cite{Yu:2019}. Adaptive Hoeffding Trees \cite{Bifet:2009} and gradient-based methods \cite{Jothimurugesan:2018} are among the most popular solutions. 

\medskip
\noindent \textbf{Ensemble approaches.} Combining multiple classifiers is a very popular and powerful approach for standard learning problems \cite{Wozniak:2014}. The technique transferred seamlessly to data stream mining scenarios, where ensemble approaches have displayed a great efficacy \cite{Krawczyk:2017}. They not only offer improved predictive power, robustness, and reduction of variance, but also can easily handle concept drift and use it as a natural way of maintaining diversity. By encapsulating new knowledge in the ensemble pool and removing outdated models, one can assure that the base classifiers are continuously mutually complementary, while adapting to changes in the stream. There are two main approaches for ensemble design in streaming environments: (i) updating base classifiers and (ii) updating the ensemble set-up. The first approach assumes that our ensemble uses classifiers capable of incremental or online learning. New instances arriving from the stream are used to continuously update those classifiers, without adding or removing any models. This provides natural adaptation to changes in the stream and retaining knowledge from previous concepts. A diversity assurance method is required, as feeding the entire stream to all base classifiers would force them to converge to a similar, if not identical, model. Popular solutions are based on the usage of online versions of bagging \cite{Bifet:2010}, boosting \cite{Oza:2001}, Random Forest \cite{Gomes:2017}, or instance-based clustering \cite{Korycki:2018}. The second approach to the ensemble design assumes that we can add new classifiers to the ensemble pool and that outdated or irrelevant classifiers can be removed via a pruning procedure. Classifiers are combined with weights reflecting the time they spent in the ensemble and their current accuracy \cite{Kolter:2007}. Accuracy Weighted Ensemble \cite{Wang:2003} and Accuracy Updated Ensemble \cite{Brzezinski:2014} enable natural encapsulation of changes in data in a form of new classifiers and allow using any base learner, as incremental mode is not required. However, methods using the dynamic ensemble set-up react slower to concept drift than their online counterparts, as they need to gather enough of new instances for a classifier training. As both approaches have their advantages and drawbacks, hybrid solutions emerge. Kappa Updated Ensemble \cite{Cano:2020} is a most recent example, where all base classifiers are updated in an online manner to offer high reactivity to concept drift, while new classifiers are trained in the background and used to replace the current ones whenever changes in the data are too severe for online updates to be handled. This allows tackling both sudden and gradual changes in streaming data.

\section{Limited budget and underfitting}
\label{sec:framework}

While a plethora of new approaches and algorithms have been proposed for continual learning from data streams, the predominant part of them share a common weakness -- they are based on a naive assumption that we have unrestrained access to labeled instances that represent a given problem in a sufficient way \cite{Krawczyk:2017}. Authors usually omit the aspect of acquiring annotated data for their methods and focus on improving their capacity to adapt, given a sufficient amount of supervision. Such an assumption is highly unfeasible in real-life scenarios \cite{Iosifidis:2019}. Instance labeling is usually connected with a monetary cost, which quickly becomes prohibitive if one wants to label a potentially infinite stream and does not have an "unlimited" budget. Even if the cost is not an issue, human experts cannot work non-stop and require a certain amount of time to analyze each instance, while being subject to various limitations, such as availability and throughput \cite{Plasse:2016, Souza:2015}. Finally, living in the era of deep learning, we must acknowledge the fact that such models require significantly higher amounts of labeled data than their shallow counterparts \cite{Roh:2018}. This has lead to a significant growth of labeling services \cite{Zhang:2016}. If we encounter such problems while working with offline models, then we should expect them to be even more severe in online systems.

Analogously to the traditional batch-based machine learning approaches, this critical obstacle has been addressed by some semi-supervised and unsupervised methods, which aim at alleviating the lack of supervision by utilizing unlabeled instances. The former group is represented mainly by algorithms based on the incremental cluster-then-label approach \cite{Ditzler:2011, Castellano:2017} and online self-labeling \cite{Korycki:2018}. Other methods involves graph-based label propagation techniques \cite{Wagner:2018} or online co-training \cite{Sousa:2017}. The unsupervised methods present a different, proactive approach, in which models try to anticipate changes in decision boundaries without labeled instances. This involves capturing some general characteristics of a stream just from initially available labeled data and simulating their most probable evolution over time \cite{Dyer:2014}, or very interesting attempts to model dynamics of a decision space \cite{Kumagai:2018}. While utilizing unlabeled data seems perfectly adequate for scenarios involving streams, it may be surprising why the actual number of works using it is so limited. It is possible that obtaining significant improvements from unreliable unlabeled data is very difficult in dynamic environments. Indeed, one of the main assumptions of semi-supervised learning is that in order to work correctly unlabeled data have to be drawn from the same, stable distributions as observed labeled instances \cite{Chapelle:2010}. This may be prohibitive in many streaming scenarios, especially if we add the fundamental cluster and smoothness requirements. In fact, results from some publications suggest that either feasible improvements from unlabeled data are barely significant or very unstable and highly dependent on characteristics of a specific stream \cite{Korycki:2017}.

As a result of the potential weak spots, it is not a coincidence that a significant attention was given to methods shifted towards supervised approaches, which are more likely to provide a reliable information regardless of a state of a stream. One of the attractive solutions is to focus on a careful selection of instances to be labeled -- namely active learning \cite{Lughofer:2017}. In static scenarios, those methods look for the most representative objects which can be used for an effective concept modeling. In dynamic environments, a crucial aspect of a proper stream labeling is a swift adaptation to changes in class characteristics \cite{Zliobaite:2014}. It is not only about finding the most valuable objects for one batch of data, but also about proper exploration of a decision space in order to find subspaces that are no longer valid and to sample from those regions a sufficient number of instances. Dynamic active learning ensures a more efficient discovery of changes and faster recovery rate after a concept drift, which prevents a classifier from prolonged drops in performance \cite{Shaker:2015}. Most popular strategies use adaptive thresholds on classifier certainty (or support functions) \cite{Zliobaite:2014} or density information \cite{Mohamad:2018b}. The problem of obtaining class labels becomes even more difficult when dealing with imbalanced class distributions in streams \cite{Zhang:2016, Korycki:2019ali}.

\begin{figure}[h]
	\centering
	\includegraphics[width=0.65\linewidth]{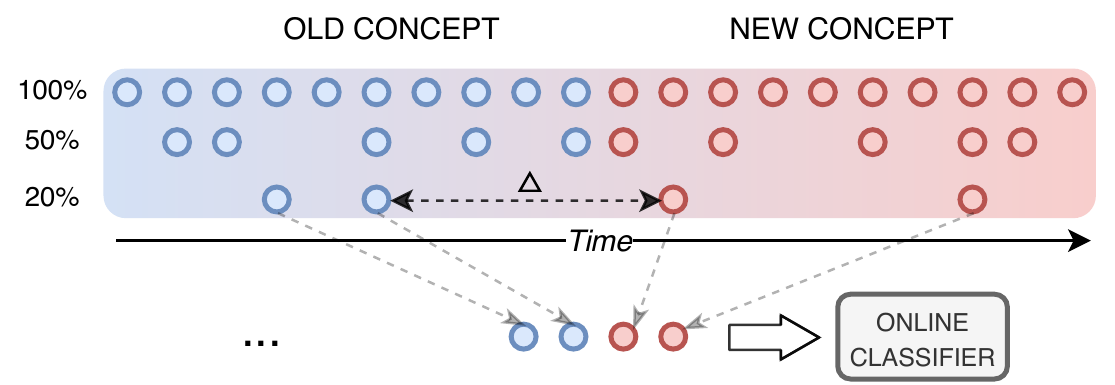}
	\caption{The sparse labeling problem.}
	\label{fig:sparse}
\end{figure}

All of the existing active learning strategies are significantly impaired when the label query budget (i.e., the number of instances allowed to be labeled) is small. In many real problems labeling even as little as $1\%$ of instances from the stream may be too costly. As a consequence, even if a strategy picks only valuable objects to be labeled and used, the effect of the single update may still be insufficient. Furthermore, it is crucial to remember that when we deal with evolving data streams and only low budgets are available, we usually get very sparse snapshots of observed concepts. Arriving labeled objects often come from distant parts of the stream, so if the data is non-stationary, they likely represent different distributions constituting temporary concepts (Fig. \ref{fig:sparse}), especially when a rate of incoming instances is low. 

As a result of these two observations, it is very likely that, while working with dynamic streams and a substantially limited labeling budget, we will encounter the underfitting problem. In fact, we may even completely overlook some of the ephemeral concepts, leaving them unnoticed. Thus, the very few labeled objects we obtain are essential for keeping our models up-to-date and we should exploit them as much as possible to avoid the waste of potential benefits. Finally, since active learning methods can be seen as exploration methods and semi-supervised ones as regularization-exploitation, approaches combining both of them are a promising research direction \cite{Dyer:2014, Korycki:2017}. However, due to the mentioned problems with dynamic environments and using unsupervised input, in this work, we focus on investigating how significant improvements can be obtained solely from the exploitation of scare but reliable supervised information, provided with actively selected instances. To the best of our knowledge, it is the first such an approach to the problem.

\section{Enhancing adaptation}
\label{sec:risky}

\subsection{Idea}

\begin{figure}[h]
	\centering
	\begin{subfigure}{0.24\linewidth}{
			\includegraphics[width=\linewidth,trim=2cm 2cm 2cm 2cm,clip]{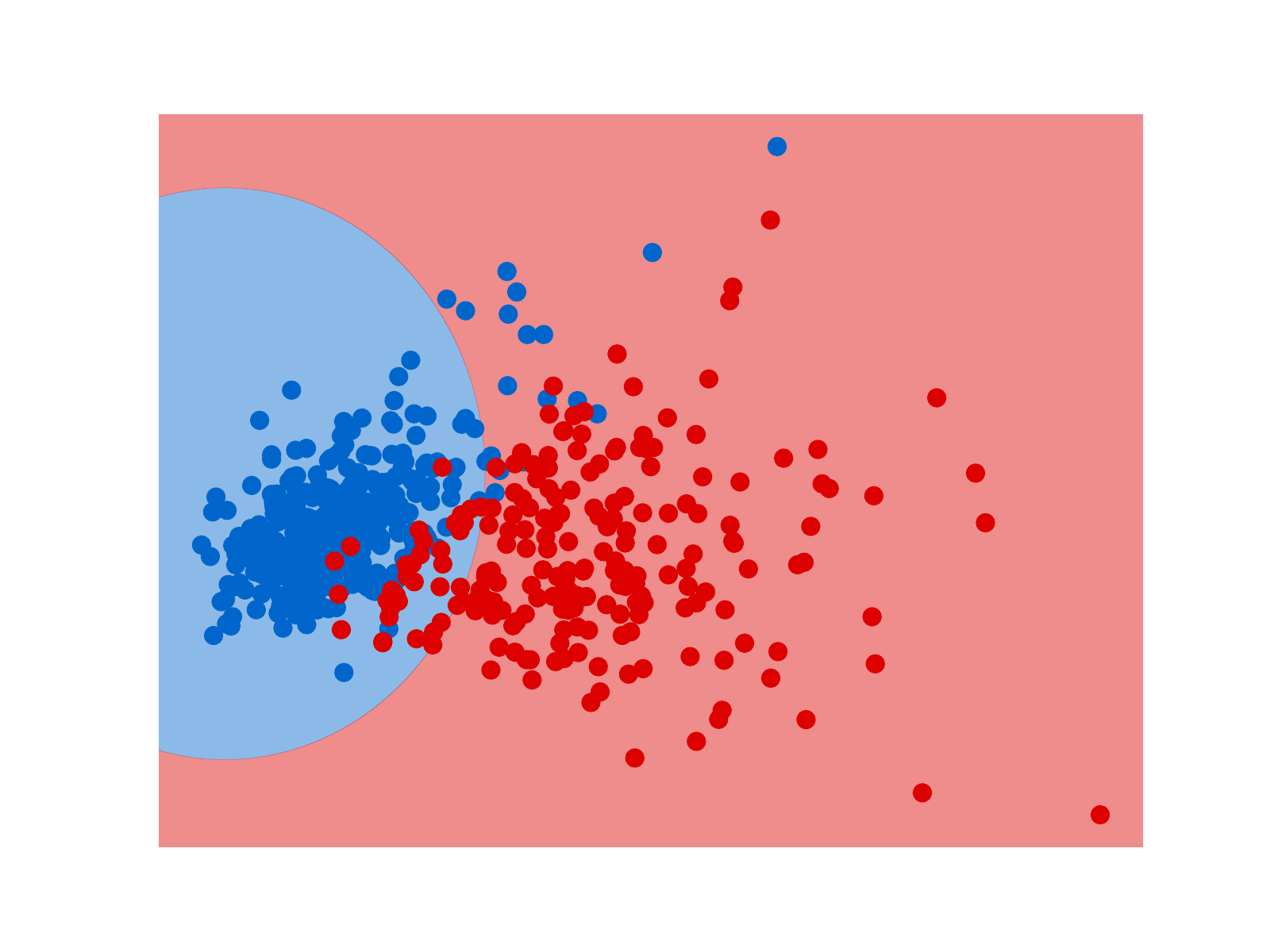}
			\subcaption{Before drift}}
	\end{subfigure}\hspace*{10pt}%
	\begin{subfigure}{0.24\linewidth}
		{\includegraphics[width=\linewidth,trim=2cm 2cm 2cm 2cm,clip]{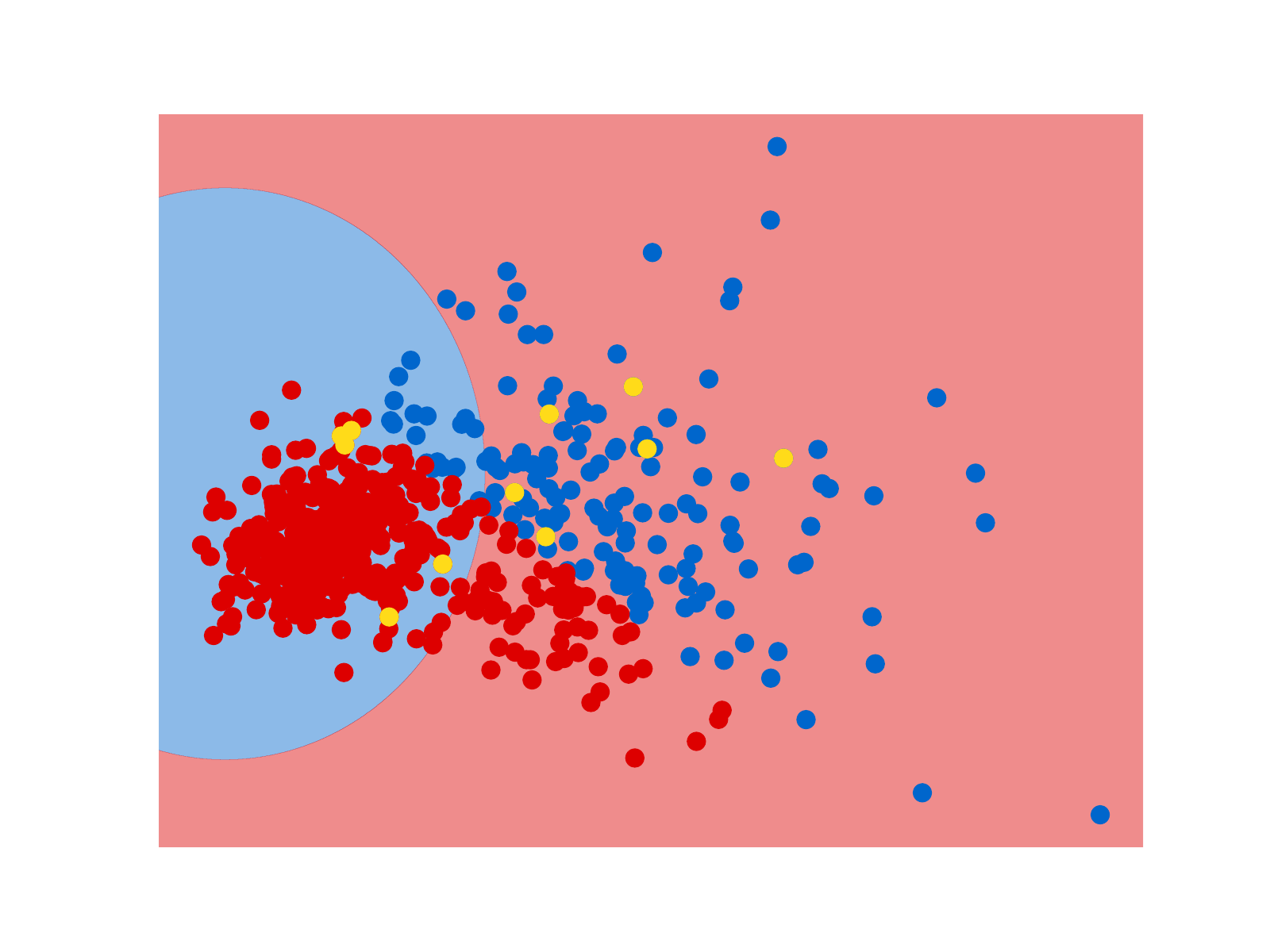}
			\subcaption{After drift, $\lambda=0$}}
	\end{subfigure}\hspace*{10pt}%
	\begin{subfigure}{0.24\linewidth}
		{\includegraphics[width=\linewidth,trim=2cm 2cm 2cm 2cm,clip]{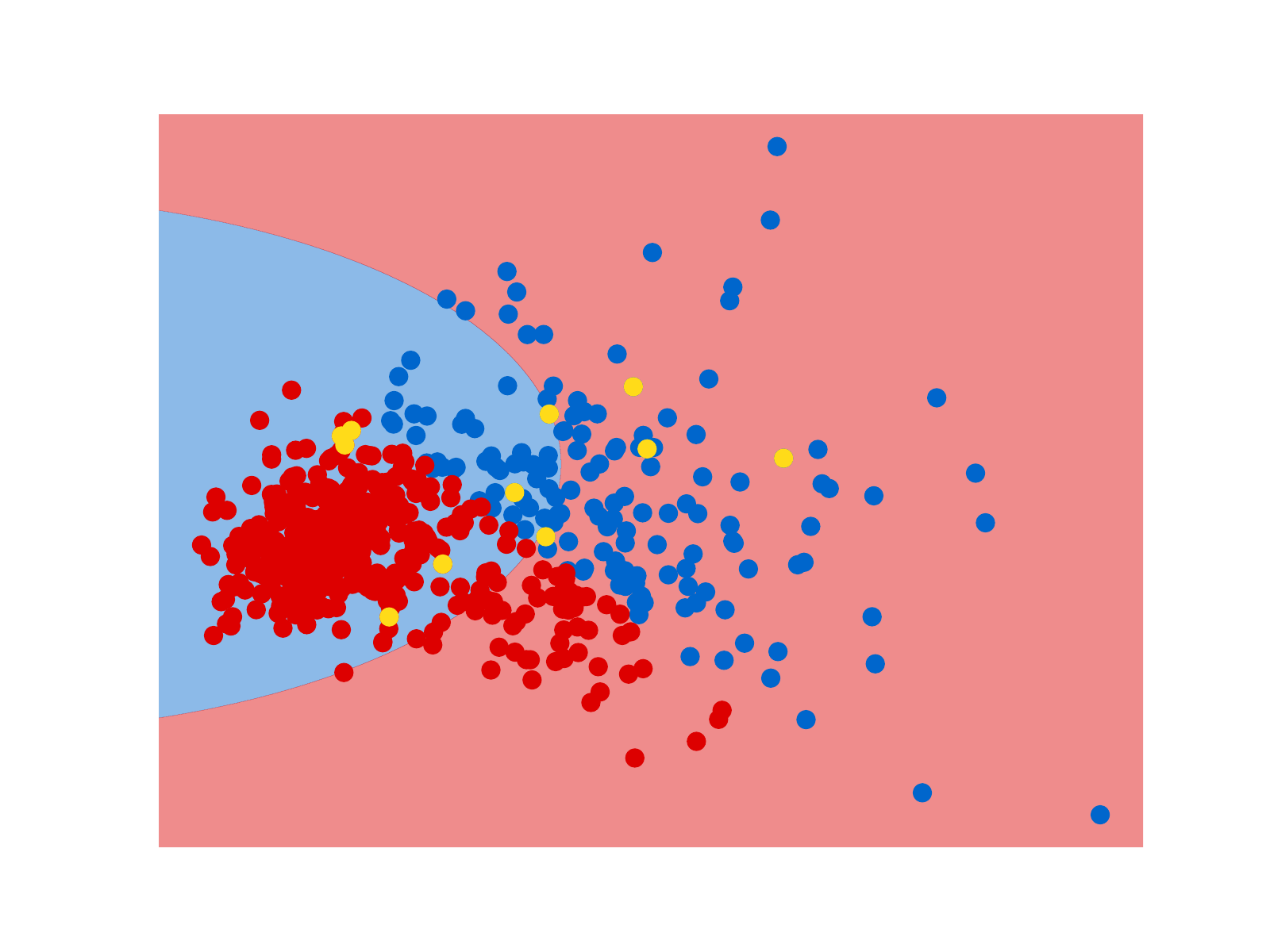}
			\subcaption{$\lambda=10$}}
	\end{subfigure}\vspace*{3pt}\\
	\begin{subfigure}{0.24\linewidth}
		{\includegraphics[width=\linewidth,trim=2cm 2cm 2cm 2cm,clip]{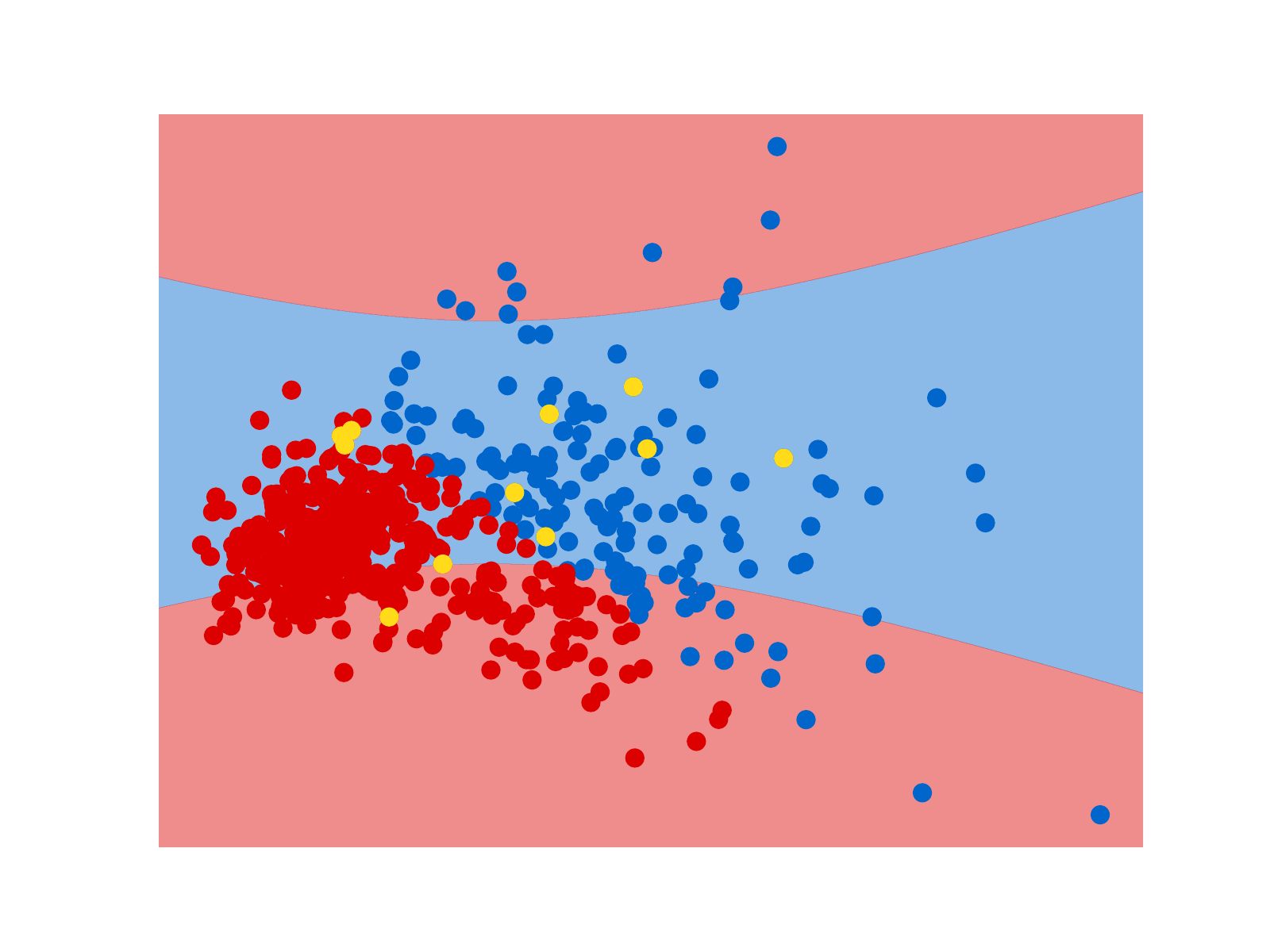}
			\subcaption{$\lambda=50$}}
	\end{subfigure}\hspace*{10pt}%
	\begin{subfigure}{0.24\linewidth}
		{\includegraphics[width=\linewidth,trim=2cm 2cm 2cm 2cm,clip]{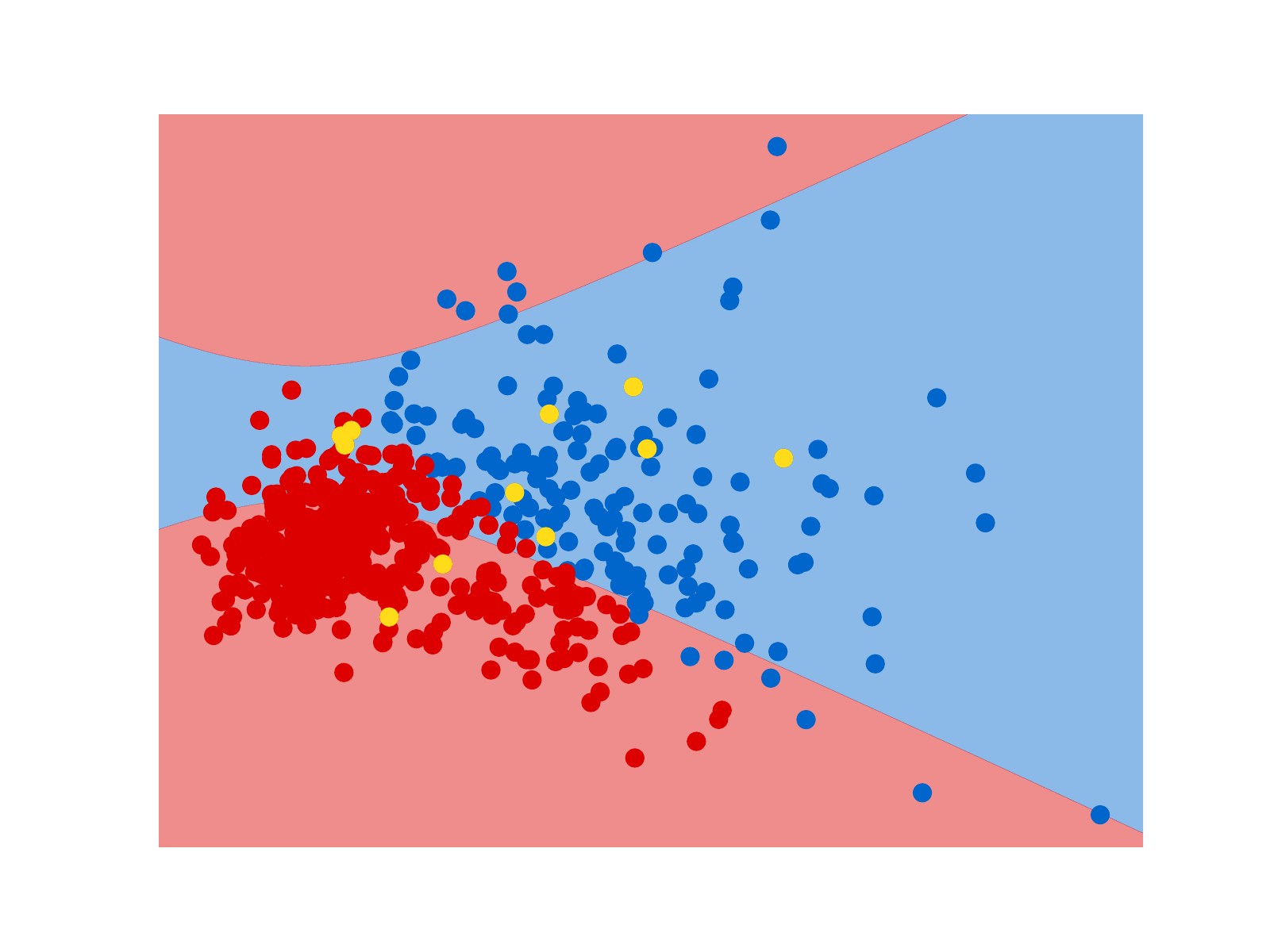}
			\subcaption{$\lambda=100$}}
	\end{subfigure}\hspace*{10pt}%
	\begin{subfigure}{0.24\linewidth}
		{\includegraphics[width=\linewidth,trim=2cm 2cm 2cm 2cm,clip]{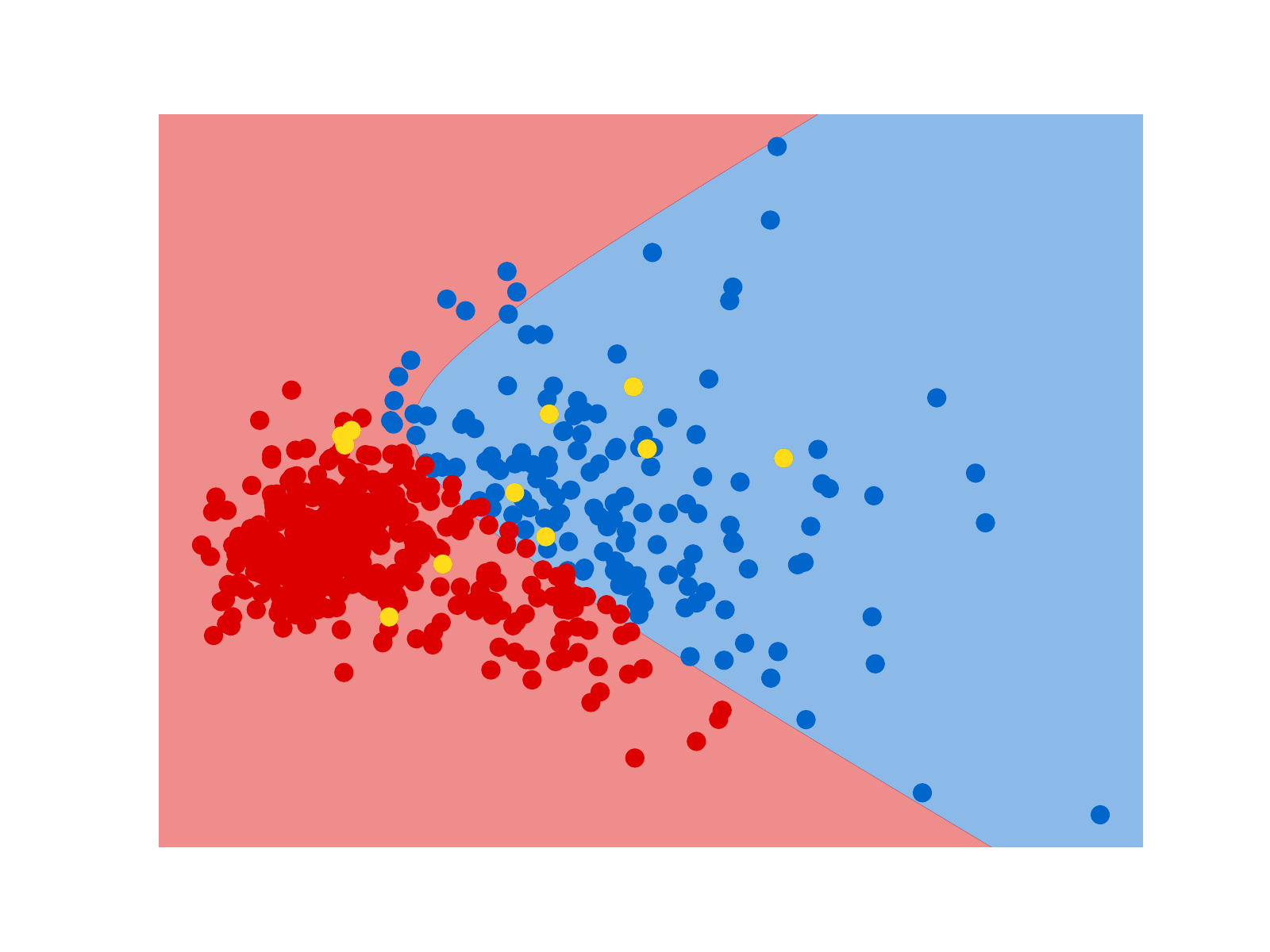}
			\subcaption{$\lambda=200$}}
	\end{subfigure}%
	\caption{Decision boundaries created by an incremental classifier, while tackling a concept drift, after learning on $\lambda$ duplicates of the selected instances.}
	\label{fig:vis}
\end{figure}

Our hypothesis is that when dealing with temporary and evolving concepts under strictly limited access to labeled data, it is reasonable to aggressively exploit the data we currently have in order to overcome underfitting and obtain efficient up-to-date models. Let us consider the following example. In Fig. \ref{fig:vis} we can see an adaptation process for a classifier while recognizing two classes (marked as red and blue points). Fig. \ref{fig:vis}a presents a state just before a drift. The classifier has learned the previous concept with sufficient accuracy, however, after the change (Fig. \ref{fig:vis}b) its model becomes practically useless. There is a need to update it, but due to a limited budget only few labeled instances selected by the active learning strategy are available (yellow). Since we deal with data streams, we can assume that the data points are currently stored in a sliding window. 

In the standard active learning approach, the chosen objects are used only once ($\lambda=0$) and the effect is unacceptable -- the class boundary did not change at all. This observation is connected with the nature of many incremental parameters maintained by the classifiers. Their internal estimators, cannot be updated efficiently using only few new values. The most straightforward solution is to reuse the limited labeled instances several times to influence the metrics in a more significant manner and boost the adaptation. We can clearly see that after some number of repetitions $\lambda$ the decision boundary is reshaping towards a proper model -- we exploit the acquired knowledge (Fig. \ref{fig:vis}c - \ref{fig:vis}f). The final result is highly dependent on the representativeness of the instances selected by active learning, but this has been already investigated in other publications \cite{Zliobaite:2014}. While we are aware of the fact that such an approach may potentially cause overfitting to the few instances, our assumption is that it is less important than the underfitting problem encountered before the risky adaptation was performed.

Such a wrapper can work with many online base learners, especially with those dedicated to evolving data streams. The currently most important online algorithm -- Adaptive Hoeffding Tree (AHT) -- may benefit from such a method, since its internal splits are rearranged only if a significantly large amount of data differs from a previous concept \cite{Bifet:2009}. In the case of the stochastic gradient descent classifier (SGD) or neural networks, while exposing wrongly classified instances several times, we can speed-up the gradient-based optimization of a loss function. Of course, while using such algorithms we can also directly focus on the criterion, for example by adjusting learning rate or momentum, however, such solutions are limited only to the gradient-based classifiers \cite{Zhu:2017cls}. Finally, Na\"ive Bayes may also work with the introduced method, since its internal estimators can be updated incrementally. It is also worth noting that we may avoid reusing instances several times by employing weight-sensitive classifiers, which can be straightforwardly adapted for this approach. In this work, due to their greater flexibility, we focus only on the more generic instance-based algorithms.

This approach can be associated with a mechanism similar to experience replay, which is a method of tackling catastrophic forgetting in deep neural networks \cite{Masson:2019}. The fundamental difference is that instead of reusing instances for stabilizing models of incoming classes, we want to erase outdated knowledge more quickly and practically "overfit" an online model to current concepts. The method is also related to resampling, known from imbalance learning, which also utilizes some instances multiple times \cite{Krawczyk:2016imb}.

One may notice that another popular approach, that allows for the most current instances to significantly affect a maintained model, is to simply retrain the model from scratch, using a collected batch of data \cite{Gama:2014}. However, while such methods work well with sudden concept drifts, they are not suitable for incremental changes. In addition, forgetting the whole priorly gathered information may be a wrong idea if the budget is limited and if we take into consideration the fact that streams may be characterized by hybrid drifts or changes affecting only some parts of the analyzed space. In the example above (Fig. \ref{fig:vis}), we would lose all the information about red points at the bottom, which were not influenced by the concept drift. We do not want to completely forget previous knowledge, just quickly and incrementally adapt to the current situation when lacking labels. One well-known way to overcome this problem is to use ensembles that consist of classifiers constructed on the basis of the most current batches \cite{Wang:2003}. Still, such an approach is much more time and memory consuming than single-classifier frameworks. Another problem is that highly limited budgets may be prohibitive in the context of building a diverse pool of classifiers. Finally, all the strategies using the retraining approach, are highly dependent on effective concept drift detectors, which increase the overall complexity of a system and are often affected by the limited access to labeled instances as well.

\subsection{Instance exploitation}
\label{sec:frame}

We propose a simple, generic, single-classifier framework that uses aggressive online updates based on instances selected for labeling by an active learning strategy. Our implementation utilizes a reactive sliding window that stores the most recent labeled instances and exposes them several times to the online classifier in order to exploit the knowledge that comes with them. We present three different probabilistic sampling methods that determine how the instances are selected from the window. The general assumption is that while active learning will effectively explore a decision space, by selecting the most representative seeds, the instance exploitation wrapper will sufficiently utilize them in order to enhance the adaptation of a classifier using only few labels. That should help us tackle the underfitting problem.

\subsubsection{Framework}
The generic wrapper framework is presented in Alg. \ref{alg:framework}. When a new instance $x$ appears, the actual budget spending $\hat{b}$ is checked. Since the data stream is infinite by definition, the current spending has to be estimated as a ratio of objects that have been already labeled to the total number of registered instances \cite{Zliobaite:2014}. If the value is below an available budget $B$, we use the active learning strategy to check if the instance is worth labeling. Some of the online strategies that can be used here were presented in \cite{Zliobaite:2014}. The methods are focused on selecting instances close to the decision boundary, based on the uncertainty of a classifier. The idea is that those regions are most likely to be affected by drifts. In order to make the strategies more exploratory and sensitive to changes in other subspaces, randomization techniques can be utilized. One of the methods implementing a hybrid comprehensive querying is the RandVar algorithm. In addition to the complex sampling, the strategy uses a variable threshold balancing budget spending. We select this method as our default strategy due to its theoretically sound design.

After receiving a positive response, the object is queried (we obtain its true label $y$), the current labeling expenses are increased and the classifier is updated with the labeled instance. Then, we activate our instance exploitation strategy, which selects indices $I$ of previously acquired labeled instances in order to reuse them and boost the adaptation process.

\begin{algorithm}[h]
	\BlankLine
	\KwData{labeling budget $B$, \textot{ActiveLearningStrategy}, \textot{ExploitationStrategy}, window size $\omega_{max}$}
	\KwResult{classifier $L$ at every iteration}
	\KwInit{$\hat{b} \gets 0$, $I \gets []$, $W \gets []$}
	\Repeat{stream ends}{
		receive incoming instance $x$\;
		\uIf{$\hat{b} < B$ {\normalfont and} \upshape\textot{ActiveLearningStrategy} $(x) = true$}{
			request the true label $y$ of instance $x$\;
			update labeling expenses $\hat{b}$\;
			update classifier $L$ and sliding window $W$ with $(x,y)$\;
			
			\BlankLine
			$I \gets $ \textot{ExploitationStrategy} $(W)$\;
			\ForEach{$i \in I$}{
				update classifier $L$ with $w_i \in W$\;
			}	
		}
	}
	\caption{Combining active learning with instance exploitation.}
	\label{alg:framework}
\end{algorithm}

Since we want to adapt a predictive model to current concepts, our exploitation methods work with a batch of the most recent instances. We maintain a sliding window $W$ that is updated with every labeled instance we receive, before using an exploitation strategy. We define the window $W$ as a sequence of $\omega \in \langle1,\omega_{max}\rangle$ instances $w_i \in W$, where $i \in \langle1,\omega\rangle$, the oldest instance is $w_1$ and the newest one is $w_{\omega}$. It is important to remember that for very low budgets, wide windows may not be able to represent actual concept properly, since the relatively small number of labeled instances will cover only few batches. For example, if $B=1\%$, $\omega_{max}=1000$ and a stream consists of 100 000 instances, then the window will not slide at all, since all labeled instances will cover exactly one such window. Therefore, in order to have a reactive sliding window, when using a low budget or dealing with a low-rate stream, its size must be adequate.

\subsubsection{Exploitation strategies}
\label{sec:strategies}

Our exploitation methods consists of strategies that simply reuse instances selected from a sliding window. In practice, they generate a multiset of indices $I = \{i~|~i \in \langle1,\omega\rangle\}$, where $|I| = \lambda$. The $\lambda$ parameter determines the intensity of the exploitation process. A single index is selected using the formula $i = \phi(\omega, r) = \lceil r\omega \rceil$, where $r$ is a random variable $r \sim \mathcal{P}(0,1)$. As a result, each strategy is modeled by a probability distribution $\mathcal{P}(0,1)$ -- it determines on which parts of the window a strategy focuses. We may see the sliding window as a probabilistic or fuzzy one.

\begin{figure}[h]
	\centering
	\begin{subfigure}{0.32\linewidth}{
			\includegraphics[width=\linewidth]{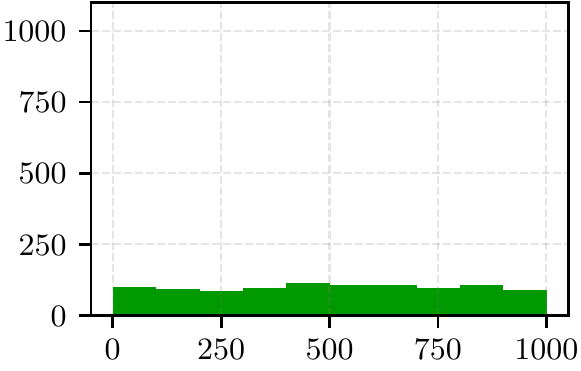}
			\subcaption{Uniform window}}
	\end{subfigure}
	\begin{subfigure}{0.32\linewidth}
		{\includegraphics[width=\linewidth]{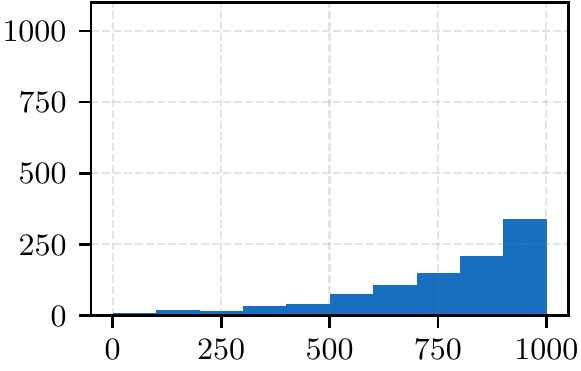}
			\subcaption{Exponential window, $\gamma=4$}}
	\end{subfigure}
	\begin{subfigure}{0.32\linewidth}
		{\includegraphics[width=\linewidth]{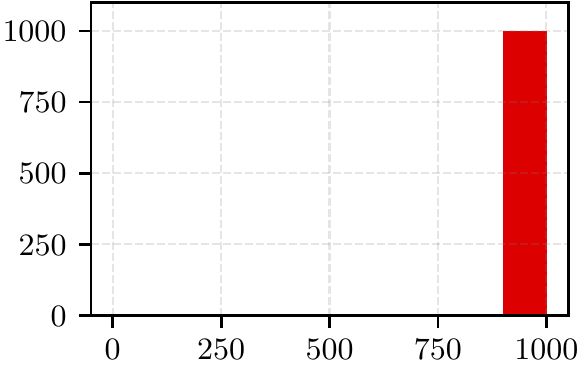}
			\subcaption{Single exposition}}
	\end{subfigure}
	\caption{Histograms for selected window indices (x-axis) depending on a strategy.}
	\label{fig:strats}
\end{figure}

\bigskip\bigskip
\noindent \textbf{Uniform Window} (\textbf{UW}) -- the most straightforward solution is to select the instances uniformly, which results in each instance having an equal chance of being chosen (Fig. \ref{fig:strats}a). The probability value is given by the uniform distribution, so a single object has $1/\omega$ chance of being used again, where $\omega$ is a number of instances within the window after a new object is added. Therefore, the indices $i$ are drawn by the strategy using $r = u \sim \mathcal{U}(0,1)$, where $u$ is a random variable sampled from the uniform distribution $\mathcal{U}(0,1)$. This approach keeps adapting a classifier to a wider context of the maintained batch of data and by doing so it may reduce the risk of falling into local minima. On the other hand, if the sliding window is insufficiently reactive, this method may fail at handling more rapid changes, since, in such case, the strategy will keep providing outdated instances. We assume that this method may work well while dealing with gradual concept drifts, when instances for both older and newer concepts should contribute to updates. The complexity for this method is solely based on the intensity $\lambda$, so it is $O(\lambda)$.\\

\noindent \textbf{Exponential Window} (\textbf{EW}) -- instead of selecting instances with an equal probability, we can focus more on the newer objects, while leaving a non-zero sampling rate for the rest of the window (Fig. \ref{fig:strats}b). We model this behavior, using the normalized exponential distribution $r = e \sim \mathcal{E}_{n}(0,1)$. Utilizing a truncated exponential transform, we have $e = -ln(u)/\gamma$, where $\gamma$ is the parameter of the exponential distribution and $u \sim \mathcal{U}(0,1)$ \cite{Eisenberg:2008}. This strategy assumes that a sliding window may imperfectly represent the current concepts, e.g, by having an inadequate size or by insufficient exploitation of the newer instances. It attempts to handle this problem on its own by increasing the chances for the more recent data. Theoretically, this approach may be adequate for more rapid incremental changes and, at the same time, acceptable for gradual and sudden concept drifts. We set $\gamma=4$ as our default value, which provides a functionality balanced between the two other strategies. The complexity is the same as for the previous approach -- $O(\lambda)$.\\

\noindent \textbf{Single Exposition} (\textbf{SE}) -- the most extreme method is reusing only the latest instance $w_{\omega}$ (Fig. \ref{fig:strats}c). Obviously, probabilities for such a strategy are: $P[i=\omega]=1$ and $P[i\neq\omega]=0$, so formally $I = \bigcup_{\lambda \in \mathbb{N}}\{\omega\}$. One of the advantages of this method is that, in practice, we do not even have to maintain a sliding window. It is enough to update the classifier with the instance several times and proceed to the next incoming objects. The strategy is a fully online one. While it has indisputable advantages regarding processing performance, it is also possible that such intense focus on a single instance may more likely lead to detrimental overfitting, which is opposite to the problem we want to solve (we spend more time on this aspect in the next sections). On the other hand, it may be a good choice for sudden drifts. The complexity of this strategy is also $O(\lambda)$.

\medskip

\subsubsection{Dynamic parameters}
\label{sec:dynamic}

It is easy to notice that two the most important parameters of the strategies are: intensity $\lambda$ and the size of the sliding window $\omega_{max}$. Generally, we may want to keep $\lambda$ relatively high for low budgets, low-rate data and unstable temporary concepts to compensate limited exploration, and low for the opposite, since having an abundance of labeled data and a sufficiently updated model, we can rely on the mechanism less. In addition, we have to control the computation time. When it comes to $\omega_{max}$, we most likely should do the opposite -- keep the size of the window low for limited budgets, low-rate data and drifting concepts, and high for more labeled instances and stable streams. These two heuristics should give us more reactivity to changes (plasticity) in the former case and provide us with more comprehensive and stable generalization (stability) in the latter one. In other words, depending on the situation we may either need to entirely focus on tackling underfitting, or to be a bit more careful no to end up with overfitting.

In our work, we empirically analyze how the values of the parameters should change depending on an amount of labeled data, as well as we provide a dynamic control technique for adjusting the values adequately to the state of a stream in a learning process. The idea is that since a model should quickly adapt to new data when it suffers from underfitting or after a concept drift, we can control $\lambda$ and $\omega_{max}$ based on a value of a current error, which should be high in both situations and which is a reliable state-of-the-art input in many drift detectors \cite{Barros:2018}. Therefore, if $\epsilon \in \langle0,1\rangle$ is a currently registered error for a model learning from a stream, for the intensity we can use:
\begin{equation}\label{eq:lambda}
\lambda(\epsilon) = \epsilon \lambda_{max}
\end{equation}

\noindent and for the size of a window:
\begin{equation}\label{eq:omega}
\omega'_{max}(\epsilon) = (1 - \epsilon) \omega_{max},
\end{equation}

\noindent where $\lambda_{max}$ and $\omega_{max}$ are fixed maximum values selected for a given budget. 

Obviously, a new problem emerges -- how should we determine the current value of $\epsilon$? The most straightforward solution is to calculate an error within a sliding window. However, in such case, we will, again, struggle with finding a proper window size $\omega_\epsilon$ that should depend on the amount of available data and a state of a stream. Fortunately, this problem has been already addressed by the well-known ADWIN algorithm \cite{Bifet:2007}, which provides an adaptive window capable of calibrating its size adequately to incoming data. At its core, the algorithm checks for each subwindow $W_0$ and $W_1$ whether $|\epsilon_{W_0} - \epsilon_{W_1}| > \theta_{cut}$, where $\theta_{cut}$ is a significance threshold based on the Hoeffding bound \cite{Hoeffding:1963} using a specified confidence value $\alpha_\theta$\footnote{Its actual implementation optimizes the computations, allowing updates with O(1) amortized and O(logW) worst-case time \cite{Bifet:2007}. Therefore, it is a very efficient algorithm for streaming data.}. If the condition holds, the algorithm shrinks the window in order to keep only the most recent consistent values, which are used for the calculation of the estimated average $\epsilon_{ADW}$. We use this indicator as a control signal in our methods, so we have $\epsilon = \epsilon_{ADW}$. Finally, since ADWIN provides the optimal window size in an online way, we can set $\omega_{max} = \omega_{ADW}$. We empirically show that all of those choices are reasoned, also in our scenario involving strict limitations on supervision.

\subsection{Alleviating overfitting}	
The presented approach to handling underfitting while learning on a budget has one significant potential weakness -- it may turn one problem into its opposite. Since our method prefers aggressive updates using few labeled instances, it becomes more likely that at some point we may encounter the overfitting problem instead of underfitting, even if we try to adequately adjust the dynamic parameters $\lambda$ and $\omega_{max}$ or when we choose a safer exploitation strategy. To alleviate it, we propose an ensemble of two paired learners, in which one of them performs risky adaptation with instance exploitation ($L_r$) and the another one learns in a standard way, without any additional enhancement ($L_s$). Such a simple ensemble can learn in two modes.

\medskip
\noindent \textbf{Switching} -- in this mode, we maintain both learners in parallel (Alg. \ref{alg:switch-learn}) and use only the currently better one for prediction, based on their respective errors $\epsilon_{r}$ and $\epsilon_{s}$ (Alg. \ref{alg:switch-pred}). The assumption is that for some parts of a stream the learner intensively exploiting instances will take the risk in the right direction, providing an efficient up-to-date model more quickly, while for the other parts (mostly more stable) it will fall into local minima, giving way to the basic learner. Since the latter tries to learn in a different way, it is possible that its perspective will allow it to temporarily outperform its counterpart, preventing overfitting.

\begin{algorithm}[h]
	\BlankLine
	\KwIn{incoming instance $\mathbf{x}$, true label $y$, risky base learner $L_{r}$, standard base learner $L_{s}$, \textot{ExploitationStrategy}}
	\KwResult{up-to-date models $L_{r}$, $L_{s}$}
	\KwInit{$\epsilon_{r} \gets 0$, $\epsilon_{s} \gets 0$}
	
	update $\epsilon_{r}$ and $\epsilon_{s}$\;
	update $L_{r}$ with ($\mathbf{x}$, y) using \textot{ExploitationStrategy} (Algorithm 1)\;
	update $L_{s}$ with ($\mathbf{x}$, y) \;
	
	\caption{The switching learning mode.}
	\label{alg:switch-learn}
\end{algorithm}

\begin{algorithm}[h]
	\BlankLine
	\KwData{incoming instance $\mathbf{x}$ risky base learner $L_{r}$, standard base learner $L_{s}$}
	\KwResult{prediction $\hat{y}$}
	
	\uIf{$\epsilon_{r} < \epsilon_{s}$}{
		$\hat{y} \gets L_{r}(\mathbf{x})$\;
	}\uElse {
		$\hat{y} \gets L_{s}(\mathbf{x})$\;
	}
	\Return $\hat{y}$\;
	
	\caption{The switching and elevating prediction mode.}
	\label{alg:switch-pred}
\end{algorithm}

\noindent \textbf{Elevating} -- in the alternative mode, we not only temporarily switch between better approaches (Alg. \ref{alg:switch-pred}), but also elevate the worse model (replacing it with the better one), if a difference between them is significant according to a chosen test $\delta$($\epsilon_{r}$, $\epsilon_{s}, \alpha_e$) (Alg. \ref{alg:elev-learn}), where $\alpha_e$ determines a significance level of the test. In this case, the idea is that instead of waiting for one model to catch up with learning, we simply instantly bring it up to speed by replacing with the more effective model. One should be careful, however, since having a better model for one timestamp does not mean that it will be easier to adapt it to a new concept from there.

\begin{algorithm}[h]
	\BlankLine
	\KwIn{incoming instance $\mathbf{x}$, true label $y$, risky base learner $L_{r}$, standard base learner $L_{s}$, \textot{ExploitationStrategy}, significance level $\alpha_e$}
	\KwResult{up-to-date models $L_{r}$, $L_{s}$}
	\KwInit{$\epsilon_{r} \gets 0$, $\epsilon_{s} \gets 0$}
	
	update $\epsilon_{r}$ and $\epsilon_{s}$\; 
	\uIf{$\delta$($\epsilon_{r}$, $\epsilon_{s}, \alpha_e$) $=$ True}{
		replace the worse model L and error $\epsilon$ with their counterparts\;
	}
	update $L_{r}$ and $L_{s}$ as in Algorithm 2\;
	
	\caption{The elevating learning mode.}
	\label{alg:elev-learn}
\end{algorithm}

To test whether two classifiers are significantly different we can use the Welch's test, which assumes that two populations (for $\epsilon_{r}$ and $\epsilon_{s}$) have unequal variances \cite{Welch:1947}. We find this assumption reasonable in our case. The statistic $t$ required for the test is obtained using the formula:
\begin{equation}
t = \dfrac{\epsilon_{r} - \epsilon_{s}}{\sqrt{\dfrac{\sigma_r^2}{N_r} + \dfrac{\sigma_s^2}{N_s}}},
\end{equation}

\noindent where required average errors $\epsilon$, their variance $\sigma$ and the numbers of samples $N$ can be easily calculated within a sliding window, including ADWIN, which we use as a default estimator. The same values are required for the degrees of freedom $\nu$, which can be approximated using the Welch-Satterthwaite equation:
\begin{equation}
\nu \approx \dfrac{\left(\dfrac{\sigma_r^2}{N_r} + \dfrac{\sigma_s^2}{N_s}\right)^2}{\dfrac{\sigma_r^4}{N_r^2\nu_r} + \dfrac{\sigma_s^4}{N_s^2\nu_s}},
\end{equation}

\noindent where $\nu_r = N_r - 1$ and $\nu_s = N_s - 1$ are the degrees of freedom associated with variance estimates. Since the Welch's test can be easily implemented in our setting, we use it as our default test.

It is worth noting that the effectiveness of both modes highly depends on the precision of determining the true value of a current error, especially when supervision is strictly limited. This problem has been already mentioned in Sec. \ref{sec:dynamic} and we empirically investigate it in the context of the proposed ensembles.

\section{Experimental study}
\label{sec:exp}

In this section, we present a comprehensive empirical evaluation of the proposed approaches. Our general goal was to provide a complex and, at the same time, an in-depth analysis of the strategies and their parameters in different settings. We focused on showing how limiting the labeling budget affects learning from data streams with or without the exploitation wrappers. The specific research questions we asked are given as follows.

\begin{itemize}
	\item \textbf{RQ1:} Does the instance exploitation improve classification while learning from sparsely labeled non-stationary data streams?
	
	\item \textbf{RQ2:} How the proposed methods should be configured depending on the available labeling budget and stability of a stream?
	
	\item \textbf{RQ3:} Is the ensemble technique, using the standard and risky classifier, capable of alleviating potential overfitting problems that may occur as a result of using the approach? Does it improve the overall performance?
	
	\item \textbf{RQ4:} Are our methods competitive when compared with some of the state-of-the-art classifiers dedicated to data streams?
\end{itemize}

In order to improve reproducibility of this study, the source code for the experiments presented in the following sections, along with information about benchmarks and details of configurations, have been uploaded to our public repository: \href{https://www.github.com/mlrep/ie-20}{\textit{github.com/mlrep/ie-20}}.

\subsection{Data streams}	
\label{sec:data}

For the purpose of the experiments, we utilized two groups of data stream benchmarks. Each of them was dedicated to a different goal of this study.

\medskip
\noindent\textbf{Synthetic streams}. The first one consists of streams that were created using artificial drift generators. Such benchmarks are commonly used while evaluating algorithms on dynamic data \cite{Barros:2018, Cano:2020}. Their well-defined characteristics can be utilized to evaluate the algorithms in very specific situations, for example, when adapting to stable or unstable concepts, or to different types of drifts. The streams were created using tools available in MOA \cite{Bifet:2010moa}, which provides some state-of-the-art generators that synthesize drifting streams by simulating transitions between different concepts. They are based on the following sigmoidal formula:
\begin{equation}
f(t) = 1/(1 + e^{-s(t-t_0)}),
\end{equation}

\noindent where $s$ controls the duration of change and $t_0$ is a peak of it. There are several different concepts that can be generated -- gaussian clusters (RBF), decision spaces modeled by a random decision tree (TREE), linearly separated subspaces (SEA) or concepts defined by some fixed formulas (STAG). There is also a simulation of a rotating hyperplane (HYPER), which does not use the function given above. We generated 14 synthetic streams (Tab. \ref{tab:data}), using different types of concepts and drifts. The latter include very sudden, less or more gradual and very slow changes. For the HYPER concept we used mediocre incremental rotations (defined by the rate parameter $\rho$). This diversification of benchmarks should provide a reliable generalization of observations.

\begin{table}[h]
	\caption{Summary of the used synthetic (left) and real (right) data streams.}
	\centering
	\hspace*{1.25cm}\scalebox{0.8}{
	\begin{subtable}{\textwidth}
		\begin{tabular}[H]{lcccccc}
			Name & \#Inst & \#Att & \#Cls & Drift & \#Drifts & Noise\\
			\midrule
			RBF1 & 1m & 15 & 5 & 100 & 3 & 0.05\\
			RBF2 & 1m & 15 & 5 & 10k & 3 & 0.05\\
			RBF3 & 1.2m & 15 & 5 & 50k & 2 & 0.05\\
			RBF4 & 1.2m & 15 & 5 & 100k & 2 & 0.05\\
			TREE1 & 1m & 15 & 5 & 100 & 3 & -\\
			TREE2 & 1m & 15 & 5 & 10k & 3 & -\\
			TREE3 & 1.2m & 15 & 5 & 50k & 2 & -\\
			TREE4 & 1.2m & 15 & 5 & 100k & 2 & -\\
			SEA1 & 600k & 3 & 2 & 100 & 3 & 0.05\\
			SEA2 & 600k & 3 & 2 & 10k & 3 & 0.05\\
			STAG1 & 600k & 3 & 2 & 100 & 3 & -\\
			STAG2 & 600k & 3 & 2 & 10k & 3 & -\\
			HYPER1 & 500k & 15 & 5 & $\rho=0.001$ & - & 0.05\\
			HYPER2 & 500k & 15 & 5 & $\rho=0.01$ & - & 0.05\\
			\bottomrule
		\end{tabular}
	\end{subtable}}%
	\hspace*{-5cm}\scalebox{0.8}{
		\begin{subtable}{\textwidth}
		\begin{tabular}[H]{lccc}	
			Name & \#Inst & \#Att & \#Cls\\
			\midrule
			Activity & 10 853 & 43 & 8\\
			Activity-Raw & 1 048 570 & 3 & 6\\
			Connect4 & 67 557 & 42 & 3\\
			Cover & 581 012 & 54 & 7\\
			Crimes & 878 049 & 3 & 39\\
			DJ30 & 138 166 & 8 & 30\\
			EEG & 14 980 & 14 & 2\\
			Elec & 45 312 & 8 & 2\\
			Gas & 13 910 & 128 & 6\\
			Poker & 829 201 & 10 & 10\\
			Power & 29 929 & 2 & 24\\
			Sensor & 2 219 804 & 5 & 57\\	 		
			Spam & 9 324 & 499 & 2\\
			Weather & 1 8158 & 8 & 2\\	 		
			\bottomrule
		\end{tabular}
	\end{subtable}}
	\label{tab:data}
\end{table}

Due to the deterministic character of the synthetic streams, we decided to use them in the first part of our experiments, in which we thoroughly investigate the influence of different configurations and the potential usability of the presented heuristics.

\medskip
\noindent\textbf{Real streams}. While the artificial data provides a well-defined environment for evaluation, real world streams may differ from them in some unspecified, unknown aspects. There is still very little research done on general dynamics of streams, therefore, it is difficult to say how realistic are the concepts and drifts generated by the artificial sources. Due to this reason, it is always important to evaluate streaming algorithms on real examples. In our experiments, we used 14 state-of-the-art real streams that are widely used in data stream mining \cite{Bifet:2010, Cano:2020}. Analogously to the synthetic streams, the chosen set represent a variety of different problems with diverse characteristics (Tab. \ref{tab:data}). Most of the selected streams are snapshots of some continuous observations registered in industrial systems or laboratories. All of them can be found either in the UCI repository, Kaggle competitions or in popular related publications. 

As opposed to the synthetic data, most of the real streams are not transparent when it comes to their internal characteristics. It means that usually we cannot say when exactly a drift occurs or what its specific type is. On the other hand, they provide a more realistic general validation. Because of that, we utilized the real streams in the second part of our experiments, conducting the final comparison between our methods (tuned in the previous step), baseline models without instance exploitation and some state-of-the-art streaming classifiers. By doing so, we also avoided fitting parameters of our algorithms to a specific testing set and fairly evaluated proposed default settings.

\subsection{Set-up}
\label{sec:set}

In this subsection, we present details of the set-up used in the experiments. For any very specific parameters or subroutines, please, refer to the source code given in the repository in Sec. \ref{sec:exp}.

\medskip
\noindent\textbf{Overview.} The experiments were conducted in the following order. First, using the synthetic streams, we evaluated how the exploitation strategies should be configured. We checked how different values of intensity ($\lambda_{max}$) and window size ($\omega_{max}$) affect the learning process. We also investigated whether adjusting these values dynamically provides any improvements in terms of performance and utilized resources. As the last part of the tuning stage, we analyzed the behavior of the supporting ensembles, focusing on the correctness of deciding which approach (standard or risky) is currently the better one, depending on the significance threshold ($\alpha_e$). Finally, we prepared the final comparison between the proposed algorithms, baseline and other well-known classifiers learning in a conservative way. It was conducted using the real streams. Last but not least, since at its core our work emphasizes the problem of limiting the access to supervision, the whole analysis was considered mainly from the perspective of having different labeling budgets ($B$). 

\medskip
\noindent\textbf{Evaluated parameters.} Below, we enclose values of parameters that were considered in our experiments. We evaluated all the values given in the third column. The last one presents other parameters that were predetermined before evaluating the main parameter.

\begin{table}[h]
	\caption{Summary of the parameters' values evaluated in the tuning experiments.}
	\centering
	\scalebox{0.8}{
	\begin{tabular}{llll}
		Parameter & Description & Values & Other \\
		\midrule
		$\lambda$ & intensity & \{1,10,100,1k\} & $\omega_{max} = \omega_{ADW}$, $\epsilon = \epsilon_{ADW}$ \\
		$\omega_{max}$ & sliding window size & \{10,100,1k,10k\} & $\lambda=100$, $\epsilon = \epsilon_{ADW}$ \\
		\hdashline
		$\alpha_e$ & elevating significance & \{0.01,0.05,0.1,0.2\} & \begin{tabular}[c]{@{}l@{}}$\lambda_{AHT}=\{1,1,1,10,10,10\},$ \\ $\lambda_{SGD}=100$ (both dynamic)\\ $\omega_{max} = \omega_{ADW}$, $\epsilon = \epsilon_{ADW}$\end{tabular}\\
		\bottomrule
	\end{tabular}}
\end{table}

While the choice of the default value for $\omega_{max}$ and the input error $\epsilon$ was somehow justified theoretically and empirically \cite{Bifet:2007}, we chose $\lambda$ either arbitrarily for the initial tests ($\omega_{max}$), or based on empirical observations, for the experiments involving ensembles ($\alpha_e$), which should use reasonably tuned base learners. Due to significant differences, we decided to set individual $\lambda_{max}$ for both classifiers. Multiple values in the last column mean that different settings were used for different labeling budgets. In our experiments, we considered $B=\{50\%,20\%,10\%,5\%,1\%\}$ and $B=100\%$ where it was necessary, including the final comparison. In addition to the given fixed values, we used $\alpha_{\theta}=\{0.05, 0.1, 0.1, 0.2, 0.2\}$ and 0.002 for $B=100\%$ (its default value) in all cases. It is the only parameter that has to be set for ADWIN. For the final comparison, we extracted recommended default configurations, based on the obtained results. We present them after the tuning phase, at the beginning of the final evaluation.

\medskip
\noindent\textbf{Classifiers.} We used two different base learners: AHT and SGD, in order to show that our framework is, indeed, able to work as a flexible wrapper. Since these classifiers use completely different learning approaches, they seem to be good candidates for providing a more general insight. As our baseline (Base) we used AHT and SGD combined with RandVar (ALRV) -- the active learning strategy mentioned in Sec. \ref{sec:frame}. Due to the fact that we utilize the same algorithm as the default query strategy in our framework, the proposed methods differ from the baseline only by applying the exploitation strategy. During the final comparison, beside of the mentioned baseline, we tested two additional variants using other active learning approaches: random query (ALR) and selective sampling (ALS) \cite{Zliobaite:2014}. Furthermore, since our methods involve ensemble techniques, we compared them with some of the state-of-the-art committees for streaming data: online bagging (OBAG) \cite{Oza:2001}, incremental bagging with ADWIN (ABAG), adaptive online boosting (ADOB) \cite{Carvalho:2014}, DWM \cite{Kolter:2007}, LNSE \cite{Polikar:2011}, AUC \cite{Brzezinski:2014} (only for AHT due to the implementation constraints) and AWE \cite{Wang:2003} (only for SGD). It is worth noting that the given ensembles represent diverse approaches to adaptation, as mentioned in Sec. \ref{sec:streams}, which, to the best of our knowledge, have not been evaluated in the context of limited supervision. All of the mentioned classifiers used recommended parameter settings given in MOA.

\medskip
\noindent\textbf{Metrics.} Most of our experiments measured the predictive performance of the considered algorithms. For this purpose, we collected kappa values in two ways. The first one involved obtaining averaged global results for whole streams (aggregated predictions), based on the test-then-train procedure \cite{Gama:2014}. The second approach was focused on registering temporal performance within a sliding window, using the prequential evaluation technique \cite{Bifet:2015}. In the case of synthetic streams, it allowed us to distinguish between stable and drift periods, and to report them separately. It is important to keep in mind that the global average tends to be biased towards stable periods, which are longer in most of the streams. Due to this reason, we also reported the average of the two separated metrics, which is more balanced. Analogously to the discussion made in Sec. \ref{sec:dynamic}, we used the ADWIN-based metrics, as recommended in \cite{Bifet:2015}. In the final phase of the experiments, we employed the Bonferroni-Dunn rank test ($\alpha=0.05$) to analyze the statistical generalization and significance of results. Finally, we find more specific metrics used in this study self-explanatory. They are briefly introduced with their presentation.

\subsection{Results and discussion}
\label{sec:res}

As we mentioned in the previous sections, the first part of the study presents different configurations and versions of the proposed algorithms, evaluated using fully controlled synthetic streams. It should provide the reader with some intuition about what to expect from those techniques under specific levels of supervision. After that, we proceed to the final evaluation and conclusions that can be drawn based on all the conducted experiments.

\subsubsection{Intensity}

\smallskip
\noindent\textbf{Fixed intensity}. The first evaluation focused on studying how intense the instance exploitation should be, meaning, how many labeled instances from the sliding window should be sampled at each step, after receiving a new object. 

In Tab. \ref{tab:fix_gen} we can see kappa values averaged over all streams for the given budgets and both classifiers. One distinctive observation is that for different classifiers we can observe significantly different preferences. While with AHT majority of the best results were obtained for $\lambda < 100$, SGD worked best with the most intensive exploitation for $\lambda=1000$. It suggests that, in general, AHT better adapts to streams even under limited supervision, while SGD severely suffers from underfitting, leaving plenty of room for improvements.

\begin{table}[h]
	\caption{Average kappa for AHT and SGD using fixed intensities.}
	\vspace*{-0.1cm}
	\centering
	\scalebox{0.8}{
		\begin{subtable}{\textwidth}\leftskip-15pt
			\begin{tabular}[H]{l >{\centering\arraybackslash} m{1.1cm} >{\centering\arraybackslash} m{1.1cm} >{\centering\arraybackslash} m{1.1cm} >{\centering\arraybackslash} m{1.1cm} >{\centering\arraybackslash} m{1.1cm}}
				\textbf{AHT}& $50\%$ & $20\%$ & $10\%$ & $5\%$ & $1\%$\\
				\midrule
				Base & 0.8556 & 0.8311 & 0.8103 & 0.7850 & 0.6916\\
				\hdashline
				UW-1 & 0.8680 & 0.8446 & 0.8202 & 0.7944 & 0.7129\\
				UW-10 & \textbf{0.8708} & \textbf{0.8533} & \textbf{0.8333} & 0.8115 & 0.7318\\
				UW-100 & 0.8571 & 0.8428 & 0.8255 & \textbf{0.8144} & \textbf{0.7324}\\
				UW-1k & 0.8359 & 0.8213 & 0.8022 & 0.7863 & 0.7059\\
				\hdashline
				EW-1 & 0.8668 & 0.8461 & 0.8252 & 0.8008 & 0.7226\\
				EW-10 & \textbf{0.8720} & \textbf{0.8567} & \textbf{0.8380} & \textbf{0.8280} & 0.7669\\
				EW-100 & 0.8555 & 0.8427 & 0.8347 & 0.8191 & \textbf{0.7753}\\
				EW-1k & 0.8445 & 0.8326 & 0.8184 & 0.8009 & 0.7559\\
				\hdashline
				SE-1 & \textbf{0.8698} & \textbf{0.8420} & \textbf{0.8270} & \textbf{0.8040} & \textbf{0.7330}\\
				SE-10 & 0.6763 & 0.6648 & 0.6631 & 0.6532 & 0.6630\\
				SE-100 & 0.4128 & 0.3986 & 0.3901 & 0.3853 & 0.3799\\
				SE-1k & 0.2438 & 0.2352 & 0.2289 & 0.2110 & 0.2295\\
				\bottomrule
			\end{tabular}
			\hspace*{0.25cm}
			\begin{tabular}[H]{l >{\centering\arraybackslash} m{1.1cm} >{\centering\arraybackslash} m{1.1cm} >{\centering\arraybackslash} m{1.1cm} >{\centering\arraybackslash} m{1.1cm} >{\centering\arraybackslash} m{1.1cm}}
				\textbf{SGD}& $50\%$ & $20\%$ & $10\%$ & $5\%$ & $1\%$\\
				\midrule
				Base & 0.4089 & 0.2892 & 0.2229 & 0.2018 & 0.1750\\
				\hdashline
				UW-1 & 0.4734 & 0.3806 & 0.2933 & 0.2245 & 0.1731\\
				UW-10 & 0.6047 & 0.5243 & 0.4721 & 0.4088 & 0.2401\\
				UW-100 & 0.6854 & 0.6557 & 0.6252 & 0.5774 & 0.4350\\
				UW-1k & \textbf{0.7069} & \textbf{0.6978} & \textbf{0.6826} & \textbf{0.6663} & \textbf{0.5573}\\
				\hdashline
				EW-1 & 0.4774 & 0.3840 & 0.2944 & 0.2239 & 0.1812\\
				EW-10 & 0.6078 & 0.5355 & 0.4798 & 0.4139 & 0.2259\\
				EW-100 & 0.6934 & 0.6670 & 0.6372 & 0.5975 & 0.4532\\
				EW-1k & \textbf{0.7148} & \textbf{0.7037} & \textbf{0.6955} & \textbf{0.6817} & \textbf{0.5943}\\
				\hdashline
				SE-1 & 0.4791 & 0.3851 & 0.3002 & 0.2335 & 0.1786\\
				SE-10 & 0.6143 & 0.5460 & 0.4866 & 0.4195 & 0.2310\\
				SE-100 & \textbf{0.6824} & 0.6594 & 0.6341 & 0.5964 & 0.4594\\
				SE-1k & 0.6760 & \textbf{0.6710} & \textbf{0.6654} & \textbf{0.6537} & \textbf{0.6024}\\
				\bottomrule
			\end{tabular}
	\end{subtable}}
	\label{tab:fix_gen}
\end{table} 

Another pattern that the results suggest is that for AHT using UW and EW there is a trend in relation between required intensity and the available labeling budget. Indeed, if we look closer and analyze Fig. \ref{fig:fix_gen_aht}, which presents the ratio between our strategies and the baseline, we will notice that as supervision is getting more and more limited, we gain more and more from exploiting the labeled instances we have (from about 1.05 to 1.2 on average). Furthermore, one should notice that for lower budgets below $B=10\%$, the difference between safer exploitation ($\lambda\leq10$) and more risky one ($\lambda\geq100$) becomes much more significant in favor of the latter approach. Finally, the gain that comes with applying the strategies is higher during drifts (1-1.4), when underfitting is more likely to occur regardless of the budget, than during stable periods (1.02-1.15).

One can also notice that the EW strategy (in Fig. \ref{fig:fix_gen_aht}, the red line represents the best value for a given budget) was slightly better than UW, especially for the lowest budget $B=1\%$. It is not surprising since in such scenarios even a reactive sliding window may contain some relatively old instances, while it is more important to focus on the strictly limited amount of the most recent information. The EW strategy provides this additional focus.

\begin{figure}[h]
	\centering
	\leftskip0pt\begin{subfigure}{0.33\textwidth}
		\begin{tikzpicture}
		
		\begin{axis}
		[
		title={STABLE},
		title style={yshift=-1.5ex},
		width=1.05\textwidth,
		height=.75\textwidth,
		ymin=0.98,
		ymax=1.17,
		ytick={1.0,1.05,1.1,1.15},
		ymajorgrids=true,
		ylabel={UW / Base},
		ylabel near ticks,
		y tick label style={
			font=\scriptsize,
			/pgf/number format/.cd,
			fixed,
			fixed zerofill,
			precision=2,
			/tikz/.cd
		},
		ylabel style={font=\scriptsize, at={(-0.15,0.5)}},
		extra y ticks=1.0,
		extra y tick labels={},
		extra y tick style={
			ymajorgrids=true,
			ytick style={
				/pgfplots/major tick length=0pt,
			},
			grid style={
				gray,
				dashed,
				/pgfplots/on layer=axis foreground,
			},
		},
		grid style={dashed,white!90!black},
		xtick style={draw=none},
		xtick=data,
		xticklabels={50,20,10,5,1},
		xlabel style={font=\footnotesize},
		x tick label style={font=\scriptsize},
		]
		
		\node[font=\tiny] at (axis cs: 0.5,0.99) {$Base$};
		
		\addplot [	
		draw=blue1,smooth,mark=*,mark options={blue1,scale=0.7}
		] coordinates {(0,1.018998997689418) (1,1.0277155310039696) (2,1.0490074143381483) (3,1.0458012631631572) (4,1.1203560229371907)};
		
		\addplot [	
		draw=blue2,smooth,mark=*,mark options={blue2,scale=0.7}
		] coordinates {(0,1.02173355983242) (1,1.03439822414923) (2,1.0412168064211) (3,1.05171352655329) (4,1.09805196307381)};
		
		\addplot [	
		draw=blue3,smooth,mark=*,mark options={blue3,scale=0.7}
		] coordinates {(0,1.02835473749238) (1,1.03707722494688) (2,1.03862119501353) (3,1.04102472694165) (4,1.10136265161407)};
		
		\addplot [	
		draw=blue4,smooth,mark=*,mark options={blue4,scale=0.7}
		] coordinates {(0,1.02063912500603) (1,1.01947973910339) (2,1.01661625065474) (3,1.01168288883821) (4,1.04304212107833)};
		
		\addplot [	
		draw=red2,smooth,mark=triangle*,mark options={red2,scale=0.7}, line width=0.8pt
		] coordinates {(0,1.02535090767702) (1,1.03900182353767) (2,1.04670071121853) (3,1.05881646158835) (4,1.15380172924896)};
		
		\end{axis}
		
		\end{tikzpicture}%
		\hspace*{0.05cm}
		\begin{tikzpicture}
		
		\begin{axis}
		[
		title={DRIFT},
		title style={yshift=-1.5ex},
		width=1.05\textwidth,
		height=.75\textwidth,
		ymin=0.9,
		ymax=1.5,
		ytick={1.0,1.2,1.4},
		ymajorgrids=true,
		ylabel near ticks,
		y tick label style={
			font=\scriptsize,
			/pgf/number format/.cd,
			fixed,
			fixed zerofill,
			precision=2,
			/tikz/.cd
		},
		ylabel style={font=\footnotesize, at={(-0.15,0.5)}},
		extra y ticks=1.0,
		extra y tick labels={},
		extra y tick style={
			ymajorgrids=true,
			ytick style={
				/pgfplots/major tick length=0pt,
			},
			grid style={
				gray,
				dashed,
				/pgfplots/on layer=axis foreground,
			},
		},
		grid style={dashed,white!90!black},
		xtick style={draw=none},
		xtick=data,
		xticklabels={50,20,10,5,1},
		xlabel style={font=\footnotesize},
		x tick label style={font=\scriptsize}
		]
		
		\node[font=\tiny] at (axis cs: 3.5,0.965) {$Base$};
		
		\addplot [	
		draw=blue1,smooth,mark=*,mark options={blue1,scale=0.7}
		] coordinates {(0,0.978399873420752) (1,1.00307544799404) (2,1.09019558397225) (3,1.18807525929443) (4,1.24260388463702)};
		
		\addplot [	
		draw=blue2,smooth,mark=*,mark options={blue2,scale=0.7}
		] coordinates {(0,1.01673357060812) (1,1.05032751004499) (2,1.12840138307144) (3,1.26727481218819) (4,1.24920436139242)};
		
		\addplot [	
		draw=blue3,smooth,mark=*,mark options={blue3,scale=0.7}
		] coordinates {(0,1.02325177069129) (1,1.05502341486412) (2,1.1170825176423) (3,1.23714027222904) (4,1.16031703711935)};
		
		\addplot [	
		draw=blue4,smooth,mark=*,mark options={blue4,scale=0.7}
		] coordinates {(0,1.01404851775822) (1,1.02230476532372) (2,1.05195348426952) (3,1.06303060991769) (4,1.01898597092849)};
		
		\addplot [	
		draw=red2,smooth,mark=triangle*,mark options={red2,scale=0.7}, line width=0.8pt
		] coordinates {(0,1.03566061624146) (1,1.07400617693083) (2,1.15693553751837) (3,1.28790834393723) (4,1.37676286772662)};
		
		\end{axis}
		
		\end{tikzpicture}%
		\hspace*{0.05cm}
		\begin{tikzpicture}
		
		\begin{axis}
		[
		title={AVG},
		title style={yshift=-1.5ex},
		width=1.05\textwidth,
		height=.75\textwidth,
		ymajorgrids=true,
		ylabel near ticks,
		y tick label style={
			font=\scriptsize,
			/pgf/number format/.cd,
			fixed,
			fixed zerofill,
			precision=2,
			/tikz/.cd
		},
		ylabel style={font=\footnotesize, at={(-0.15,0.5)}},
		extra y ticks=1.0,
		extra y tick labels={},
		extra y tick style={
			ymajorgrids=true,
			ytick style={
				/pgfplots/major tick length=0pt,
			},
			grid style={
				gray,
				dashed,
				/pgfplots/on layer=axis foreground,
			},
		},
		grid style={dashed,white!90!black},
		xtick style={draw=none},
		xtick=data,
		xticklabels={50,20,10,5,1},
		xlabel style={font=\footnotesize},
		x tick label style={font=\scriptsize},
		]
		
		\node[font=\tiny] at (axis cs: 3.5,0.985) {$Base$};
		
		\addplot [	
		draw=blue1,smooth,mark=*,mark options={blue1,scale=0.7}
		] coordinates {(0,0.99296050637618) (1,1.01093876040169) (2,1.05013863278576) (3,1.09092821442937) (4,1.12930728121526)};
		
		\addplot [	
		draw=blue2,smooth,mark=*,mark options={blue2,scale=0.7}
		] coordinates {(0,1.01947145275659) (1,1.04145777760176) (2,1.07802410206436) (3,1.13579372927443) (4,1.15236739180079)};
		
		\addplot [	
		draw=blue3,smooth,mark=*,mark options={blue3,scale=0.7}
		] coordinates {(0,1.02604604105512) (1,1.04503063151006) (2,1.07174573618256) (3,1.11752007143713) (4,1.12254744761912)};
		
		\addplot [	
		draw=blue4,smooth,mark=*,mark options={blue4,scale=0.7}
		] coordinates {(0,1.0176573867221) (1,1.02073173644576) (2,1.03153480683573) (3,1.03171119208809) (4,1.03439773221355)};
		
		\addplot [	
		draw=red2,smooth,mark=triangle*,mark options={red2,scale=0.7}, line width=0.8pt
		] coordinates {(0,1.02988857371431) (1,1.05350829501919) (2,1.093239284871) (3,1.14616826317283) (4,1.23392106549886)};
		
		\end{axis}
		
		\end{tikzpicture}%
	\end{subfigure}\\
	\begin{subfigure}{0.33\textwidth}
		\begin{tikzpicture}
		
		\begin{axis}
		[
		title={STABLE},
		title style={yshift=-1.5ex},
		width=1.05\textwidth,
		height=.75\textwidth,
		ymajorgrids=true,
		ylabel={SE / Base},
		ylabel near ticks,
		y tick label style={
			font=\scriptsize,
			/pgf/number format/.cd,
			fixed,
			fixed zerofill,
			precision=2,
			/tikz/.cd
		},
		ylabel style={font=\scriptsize, at={(-0.15,0.5)}},
		extra y ticks=1.0,
		extra y tick labels={},
		extra y tick style={
			ymajorgrids=true,
			ytick style={
				/pgfplots/major tick length=0pt,
			},
			grid style={
				gray,
				dashed,
				/pgfplots/on layer=axis foreground,
			},
		},
		grid style={dashed,white!90!black},
		xtick style={draw=none},
		xtick=data,
		xticklabels={50,20,10,5,1},
		xlabel style={font=\footnotesize},
		x tick label style={font=\scriptsize},
		]
		
		\node[font=\tiny] at (axis cs: 0.5,0.95) {$Base$};
		
		\addplot [	
		draw=blue1,smooth,mark=*,mark options={blue1,scale=0.7}
		] coordinates {(0,0.327755764914704) (1,0.323871164248097) (2,0.321281218259916) (3,0.304846522153109) (4,0.37936784022744)};
		
		\addplot [	
		draw=blue2,smooth,mark=*,mark options={blue2,scale=0.7}
		] coordinates {(0,0.521453475942921) (1,0.51645858599242) (2,0.517958289169789) (3,0.524652944955192) (4,0.58647781899406)};
		
		\addplot [	
		draw=blue3,smooth,mark=*,mark options={blue3,scale=0.7}
		] coordinates {(0,0.812741809156096) (1,0.82305304675471) (2,0.841888126404002) (3,0.849666089331746) (4,0.987225801900935)};
		
		\addplot [	
		draw=blue4,smooth,mark=*,mark options={blue4,scale=0.7}
		] coordinates {(0,1.01993236008648) (1,1.01457684233261) (2,1.02189483917661) (3,1.01860194702221) (4,1.06693240663554)};
		
		\end{axis}
		
		\end{tikzpicture}%
		\hspace*{0.05cm}
		\begin{tikzpicture}
		
		\begin{axis}
		[
		title={DRIFT},
		title style={yshift=-1.5ex},
		width=1.05\textwidth,
		height=.75\textwidth,
		ymin=0.25,
		ymax=1.35,
		ytick={0.4,0.6,0.8,1.0,1.2},
		ymajorgrids=true,
		ylabel near ticks,
		y tick label style={
			font=\scriptsize,
			/pgf/number format/.cd,
			fixed,
			fixed zerofill,
			precision=2,
			/tikz/.cd
		},
		ylabel style={font=\footnotesize, at={(-0.15,0.5)}},
		extra y ticks=1.0,
		extra y tick labels={},
		extra y tick style={
			ymajorgrids=true,
			ytick style={
				/pgfplots/major tick length=0pt,
			},
			grid style={
				gray,
				dashed,
				/pgfplots/on layer=axis foreground,
			},
		},
		grid style={dashed,white!90!black},
		xtick style={draw=none},
		xtick=data,
		xticklabels={50,20,10,5,1},
		xlabel style={font=\footnotesize},
		x tick label style={font=\scriptsize}
		]
		
		\node[font=\tiny] at (axis cs: 0.5,0.93) {$Base$};
		
		\addplot [	
		draw=blue1,smooth,mark=*,mark options={blue1,scale=0.7}
		] coordinates {(0,0.328498587034075) (1,0.349633234642019) (2,0.374702624628602) (3,0.352140813576666) (4,0.435265618614236)};
		
		\addplot [	
		draw=blue2,smooth,mark=*,mark options={blue2,scale=0.7}
		] coordinates {(0,0.522750882266987) (1,0.544919848781614) (2,0.563950543327518) (3,0.645510063480793) (4,0.751383894786007)};
		
		\addplot [	
		draw=blue3,smooth,mark=*,mark options={blue3,scale=0.7}
		] coordinates {(0,0.863736213854579) (1,0.888547153527287) (2,0.975482975242411) (3,1.1129838277361) (4,1.15802157409264)};
		
		\addplot [	
		draw=blue4,smooth,mark=*,mark options={blue4,scale=0.7}
		] coordinates {(0,1.02075672660287) (1,1.03626817243876) (2,1.09968777058396) (3,1.17869577586247) (4,1.02397951672521)};
		
		\end{axis}
		
		\end{tikzpicture}%
		\hspace*{0.05cm}
		\begin{tikzpicture}
		
		\begin{axis}
		[
		title={AVG},
		title style={yshift=-1.5ex},
		width=1.05\textwidth,
		height=.75\textwidth,
		ymajorgrids=true,
		ylabel near ticks,
		y tick label style={
			font=\scriptsize,
			/pgf/number format/.cd,
			fixed,
			fixed zerofill,
			precision=2,
			/tikz/.cd
		},
		ylabel style={font=\footnotesize, at={(-0.15,0.5)}},
		extra y ticks=1.0,
		extra y tick labels={},
		extra y tick style={
			ymajorgrids=true,
			ytick style={
				/pgfplots/major tick length=0pt,
			},
			grid style={
				gray,
				dashed,
				/pgfplots/on layer=axis foreground,
			},
		},
		grid style={dashed,white!90!black},
		xtick style={draw=none},
		xtick=data,
		xticklabels={50,20,10,5,1},
		xlabel style={font=\footnotesize},
		x tick label style={font=\scriptsize},
		]
		
		\node[font=\tiny] at (axis cs: 0.5,0.94) {$Base$};
		
		\addplot [	
		draw=blue1,smooth,mark=*,mark options={blue1,scale=0.7}
		] coordinates {(0,0.328091834273443) (1,0.33528841886855) (2,0.343834490563765) (3,0.323293775326298) (4,0.399454268439875)};
		
		\addplot [	
		draw=blue2,smooth,mark=*,mark options={blue2,scale=0.7}
		] coordinates {(0,0.522040451613115) (1,0.529072070821672) (2,0.537375146613196) (3,0.571793557407912) (4,0.645735523840015)};
		
		\addplot [	
		draw=blue3,smooth,mark=*,mark options={blue3,scale=0.7}
		] coordinates {(0,0.835812819616603) (1,0.852078776508872) (2,0.898288756957417) (3,0.952373812380902) (4,1.04859992295056)};
		
		\addplot [	
		draw=blue4,smooth,mark=*,mark options={blue4,scale=0.7}
		] coordinates {(0,1.02030532195623) (1,1.02419002307047) (2,1.05473720115481) (3,1.0810469335774) (4,1.05149762256004)};
		
		\end{axis}
		
		\end{tikzpicture}
	\end{subfigure}%
	\caption{Improvement over Base given a budget for AHT using fixed intensities: \mycircle{blue1} 1k, \mycircle{blue2} 100, \mycircle{blue3} 10, \mycircle{blue4} 1, \mytriangle{red2} best EW.}
	\label{fig:fix_gen_aht}
\end{figure}
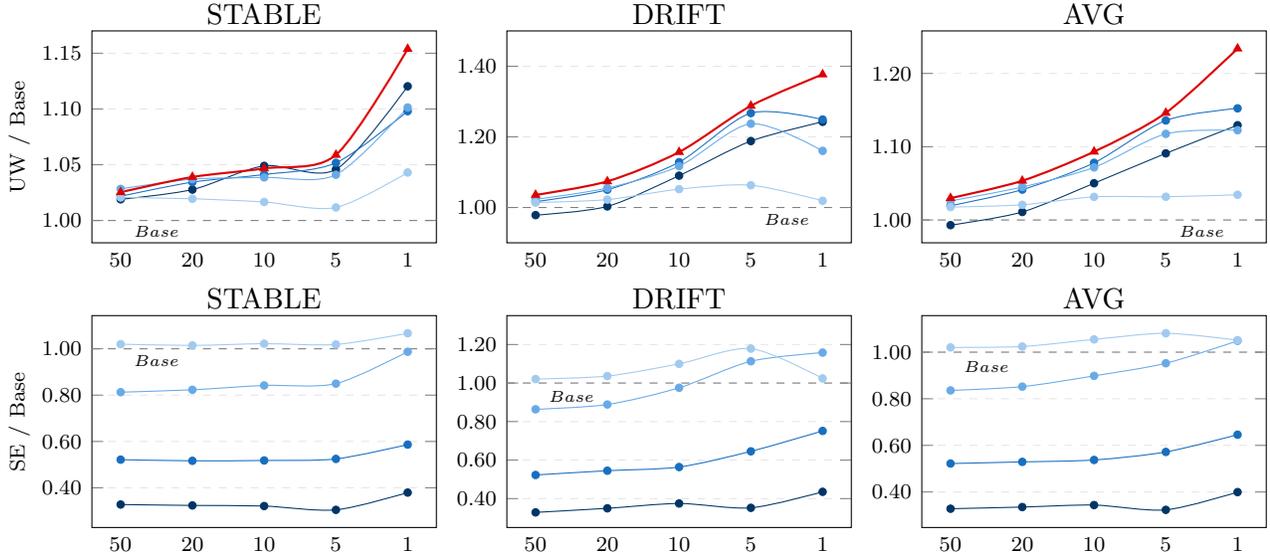

\begin{figure}[h]
	\centering
	\leftskip0pt\begin{subfigure}{0.33\textwidth}
		\begin{tikzpicture}
		
		\begin{axis}
		[
		title={STABLE},
		title style={yshift=-1.5ex},
		width=1.05\textwidth,
		height=.75\textwidth,
		ymin=0.75,
		ymax=3.45,
		ymajorgrids=true,
		ylabel={UW / Base},
		ylabel near ticks,
		ylabel style={font=\scriptsize, at={(-0.15,0.5)}},
		y tick label style={
			font=\scriptsize,
			/pgf/number format/.cd,
			fixed,
			fixed zerofill,
			precision=2,
			/tikz/.cd
		},
		extra y ticks=1.0,
		extra y tick labels={},
		extra y tick style={
			ymajorgrids=true,
			ytick style={
				/pgfplots/major tick length=0pt,
			},
			grid style={
				gray,
				dashed,
				/pgfplots/on layer=axis foreground,
			},
		},
		grid style={dashed,white!90!black},
		xtick style={draw=none},
		xtick=data,
		xticklabels={50,20,10,5,1},
		xlabel style={font=\footnotesize},
		x tick label style={font=\scriptsize},
		]
		
		\node[font=\tiny] at (axis cs: 0.5,0.87) {$Base$};
		
		\addplot [	
		draw=blue1,smooth,mark=*,mark options={blue1,scale=0.7}
		] coordinates {(0,1.71483286001613) (1,2.25109597856433) (2,2.73427727332352) (3,3.12629150621111) (4,3.08962567597951)};
		
		\addplot [	
		draw=blue2,smooth,mark=*,mark options={blue2,scale=0.7}
		] coordinates {(0,1.66933306741231) (1,2.12961255940472) (2,2.52679883811415) (3,2.7259556254465) (4,2.37625174787219)};
		
		\addplot [	
		draw=blue3,smooth,mark=*,mark options={blue3,scale=0.7}
		] coordinates {(0,1.48974393175142) (1,1.69489562255845) (2,1.88865562162743) (3,1.90298147577507) (4,1.36145227325155)};
		
		\addplot [	
		draw=blue4,smooth,mark=*,mark options={blue4,scale=0.7}
		] coordinates {(0,1.15254873811833) (1,1.23349973930064) (2,1.23683436091615) (3,1.16981410241866) (4,0.988921899817938)};
		
		\addplot [	
		draw=red2,smooth,mark=triangle*,mark options={red2,scale=0.7}, line width=0.8pt
		] coordinates {(0,1.72513581452955) (1,2.2546500326659) (2,2.76312189418025) (3,3.16977372320708) (4,3.25020809726809)};
		
		\end{axis}
		
		\end{tikzpicture}%
		\hspace*{0.05cm}
		\begin{tikzpicture}
		
		\begin{axis}
		[
		title={DRIFT},
		title style={yshift=-1.5ex},
		width=1.05\textwidth,
		height=.75\textwidth,
		ymin=0.6,
		ymax=3.7,
		ymajorgrids=true,
		ylabel near ticks,
		y tick label style={
			font=\scriptsize,
			/pgf/number format/.cd,
			fixed,
			fixed zerofill,
			precision=2,
			/tikz/.cd
		},
		extra y ticks=1.0,
		extra y tick labels={},
		extra y tick style={
			ymajorgrids=true,
			ytick style={
				/pgfplots/major tick length=0pt,
			},
			grid style={
				gray,
				dashed,
				/pgfplots/on layer=axis foreground,
			},
		},
		grid style={dashed,white!90!black},
		xtick style={draw=none},
		xtick=data,
		xticklabels={50,20,10,5,1},
		xlabel style={font=\footnotesize},
		x tick label style={font=\scriptsize}
		]
		
		\node[font=\tiny] at (axis cs: 0.5,0.83) {$Base$};
		
		\addplot [	
		draw=blue1,smooth,mark=*,mark options={blue1,scale=0.7}
		] coordinates {(0,2.1815535284795) (1,3.02584870565871) (2,3.37879106631933) (3,3.28582639017075) (4,2.55621156703242)};
		
		\addplot [	
		draw=blue2,smooth,mark=*,mark options={blue2,scale=0.7}
		] coordinates {(0,2.00147522951245) (1,2.62419793088778) (2,2.77145826139974) (3,2.51042110835616) (4,2.00615431425707)};
		
		\addplot [	
		draw=blue3,smooth,mark=*,mark options={blue3,scale=0.7}
		] coordinates {(0,1.58456805176115) (1,1.89562389332268) (2,1.96179031992252) (3,1.69408768991993) (4,1.20511698172335)};
		
		\addplot [	
		draw=blue4,smooth,mark=*,mark options={blue4,scale=0.7}
		] coordinates {(0,1.20681994969672) (1,1.30572932900629) (2,1.16028936391312) (3,1.03514733166859) (4,0.973570700491524)};
		
		\addplot [	
		draw=red2,smooth,mark=triangle*,mark options={red2,scale=0.7}, line width=0.8pt
		] coordinates {(0,2.19999853018682) (1,3.0807736963246) (2,3.46827328305367) (3,3.43592692127765) (4,2.89614349653329)};
		
		\end{axis}
		
		\end{tikzpicture}%
		\hspace*{0.05cm}
		\begin{tikzpicture}
		
		\begin{axis}
		[
		title={AVG},
		title style={yshift=-1.5ex},
		width=1.05\textwidth,
		height=.75\textwidth,
		ymajorgrids=true,
		ylabel near ticks,
		y tick label style={
			font=\scriptsize,
			/pgf/number format/.cd,
			fixed,
			fixed zerofill,
			precision=2,
			/tikz/.cd
		},
		ylabel style={font=\scriptsize, at={(-0.24,0.5)}},
		extra y ticks=1.0,
		extra y tick labels={},
		extra y tick style={
			ymajorgrids=true,
			ytick style={
				/pgfplots/major tick length=0pt,
			},
			grid style={
				gray,
				dashed,
				/pgfplots/on layer=axis foreground,
			},
		},
		grid style={dashed,white!90!black},
		xtick style={draw=none},
		xtick=data,
		xticklabels={50,20,10,5,1},
		xlabel style={font=\footnotesize},
		x tick label style={font=\scriptsize},
		]
		
		\node[font=\tiny] at (axis cs: 0.5,0.87) {$Base$};
		
		\addplot [	
		draw=blue1,smooth,mark=*,mark options={blue1,scale=0.7}
		] coordinates {(0,1.90106973434089) (1,2.54378717688883) (2,2.98680095036795) (3,3.19288128472758) (4,2.86641075682458)};
		
		\addplot [	
		draw=blue2,smooth,mark=*,mark options={blue2,scale=0.7}
		] coordinates {(0,1.80186869799549) (1,2.31646028512497) (2,2.62265760461855) (3,2.63599162886171) (4,2.22137907514347)};
		
		\addplot [	
		draw=blue3,smooth,mark=*,mark options={blue3,scale=0.7}
		] coordinates {(0,1.52758186976122) (1,1.77072807342046) (2,1.91731015646239) (3,1.81578931699255) (4,1.29603148877247)};
		
		\addplot [	
		draw=blue4,smooth,mark=*,mark options={blue4,scale=0.7}
		] coordinates {(0,1.17420473386297) (1,1.26078711038835) (2,1.20684365425848) (3,1.11360426156627) (4,0.982497966406539)};
		
		\addplot [	
		draw=red2,smooth,mark=triangle*,mark options={red2,scale=0.7}, line width=0.8pt
		] coordinates {(0,1.91462163307825) (1,2.56674848243494) (2,3.0394036798091) (3,3.28086593119999) (4,3.10204459745995)};
		
		\end{axis}
		
		\end{tikzpicture}%
	\end{subfigure}\\
	\begin{subfigure}{0.33\textwidth}
		\begin{tikzpicture}
		
		\begin{axis}
		[
		title={STABLE},
		title style={yshift=-1.5ex},
		width=1.05\textwidth,
		height=.75\textwidth,
		ymin=0.75,
		ymax=3.45,
		ymajorgrids=true,
		ylabel={SE / Base},
		ylabel near ticks,
		y tick label style={
			font=\scriptsize,
			/pgf/number format/.cd,
			fixed,
			fixed zerofill,
			precision=2,
			/tikz/.cd
		},
		ylabel style={font=\scriptsize, at={(-0.15,0.5)}},
		extra y ticks=1.0,
		extra y tick labels={},
		extra y tick style={
			ymajorgrids=true,
			ytick style={
				/pgfplots/major tick length=0pt,
			},
			grid style={
				gray,
				dashed,
				/pgfplots/on layer=axis foreground,
			},
		},
		grid style={dashed,white!90!black},
		xtick style={draw=none},
		xtick=data,
		xticklabels={50,20,10,5,1},
		xlabel style={font=\footnotesize},
		x tick label style={font=\scriptsize}
		]
		
		\node[font=\tiny] at (axis cs: 0.5,0.87) {$Base$};
		
		\addplot [	
		draw=blue1,smooth,mark=*,mark options={blue1,scale=0.7}
		] coordinates {(0,1.62887645595342) (1,2.14173731912172) (2,2.63277356027865) (3,3.02480897073971) (4,3.30221387703675)};
		
		\addplot [	
		draw=blue2,smooth,mark=*,mark options={blue2,scale=0.7}
		] coordinates {(0,1.64194954918135) (1,2.11028746619462) (2,2.5246343144027) (3,2.77942840561535) (4,2.48026537071541)};
		
		\addplot [	
		draw=blue3,smooth,mark=*,mark options={blue3,scale=0.7}
		] coordinates {(0,1.50306339358714) (1,1.76416923359536) (2,1.93455674884306) (3,1.94310213727799) (4,1.38627379210424)};
		
		\addplot [	
		draw=blue4,smooth,mark=*,mark options={blue4,scale=0.7}
		] coordinates {(0,1.16771327454983) (1,1.24097608471383) (2,1.23738001787601) (3,1.16738832497633) (4,1.03644294781513)};
		
		\end{axis}
		
		\end{tikzpicture}%
		\hspace*{0.05cm}
		\begin{tikzpicture}
		
		\begin{axis}
		[
		title={DRIFT},
		title style={yshift=-1.5ex},
		width=1.05\textwidth,
		height=.75\textwidth,
		ymin=0.6,
		ymax=3.7,
		ymajorgrids=true,
		ylabel near ticks,
		y tick label style={
			font=\scriptsize,
			/pgf/number format/.cd,
			fixed,
			fixed zerofill,
			precision=2,
			/tikz/.cd
		},
		ylabel style={font=\scriptsize, at={(-0.24,0.5)}},
		extra y ticks=1.0,
		extra y tick labels={},
		extra y tick style={
			ymajorgrids=true,
			ytick style={
				/pgfplots/major tick length=0pt,
			},
			grid style={
				gray,
				dashed,
				/pgfplots/on layer=axis foreground,
			},
		},
		grid style={dashed,white!90!black},
		xtick style={draw=none},
		xtick=data,
		xticklabels={50,20,10,5,1},
		xlabel style={font=\footnotesize},
		x tick label style={font=\scriptsize}
		]
		
		\node[font=\tiny] at (axis cs: 0.5,0.83) {$Base$};
		
		\addplot [	
		draw=blue1,smooth,mark=*,mark options={blue1,scale=0.7}
		] coordinates {(0,2.13259983901149) (1,2.99853283180654) (2,3.35549524148651) (3,3.3361008179447) (4,3.06455513560546)};
		
		\addplot [	
		draw=blue2,smooth,mark=*,mark options={blue2,scale=0.7}
		] coordinates {(0,2.01481118005843) (1,2.63118640386282) (2,2.81978122407037) (3,2.64606094123038) (4,2.24761555066451)};
		
		\addplot [	
		draw=blue3,smooth,mark=*,mark options={blue3,scale=0.7}
		] coordinates {(0,1.5956615580294) (1,1.99867088549332) (2,2.06217960878896) (3,1.77585554096561) (4,1.21118852413674)};
		
		\addplot [	
		draw=blue4,smooth,mark=*,mark options={blue4,scale=0.7}
		] coordinates {(0,1.21969489344903) (1,1.3171383692083) (2,1.16030051083126) (3,1.04857835436021) (4,0.963507539831374)};
		
		\end{axis}
		
		\end{tikzpicture}%
		\hspace*{0.05cm}
		\begin{tikzpicture}
		
		\begin{axis}
		[
		title={AVG},
		title style={yshift=-1.5ex},
		width=1.05\textwidth,
		height=.75\textwidth,
		ymajorgrids=true,
		ylabel near ticks,
		y tick label style={
			font=\scriptsize,
			/pgf/number format/.cd,
			fixed,
			fixed zerofill,
			precision=2,
			/tikz/.cd
		},
		ylabel style={font=\scriptsize, at={(-0.24,0.5)}},
		extra y ticks=1.0,
		extra y tick labels={},
		extra y tick style={
			ymajorgrids=true,
			ytick style={
				/pgfplots/major tick length=0pt,
			},
			grid style={
				gray,
				dashed,
				/pgfplots/on layer=axis foreground,
			},
		},
		grid style={dashed,white!90!black},
		xtick style={draw=none},
		xtick=data,
		xticklabels={50,20,10,5,1},
		xlabel style={font=\footnotesize},
		x tick label style={font=\scriptsize},
		]
		
		\node[font=\tiny] at (axis cs: 0.5,0.89) {$Base$};
		
		\addplot [	
		draw=blue1,smooth,mark=*,mark options={blue1,scale=0.7}
		] coordinates {(0,1.82987862861034) (1,2.46542318243876) (2,2.91593947294288) (3,3.15474202766667) (4,3.20276210801669)};
		
		\addplot [	
		draw=blue2,smooth,mark=*,mark options={blue2,scale=0.7}
		] coordinates {(0,1.79073358507761) (1,2.30707610460814) (2,2.64027432662206) (3,2.72376089457153) (4,2.382909657953)};
		
		\addplot [	
		draw=blue3,smooth,mark=*,mark options={blue3,scale=0.7}
		] coordinates {(0,1.54001310222274) (1,1.85276081527395) (2,1.98456001365864) (3,1.8732934939916) (4,1.31300680632516)};
		
		\addplot [	
		draw=blue4,smooth,mark=*,mark options={blue4,scale=0.7}
		] coordinates {(0,1.18845564762279) (1,1.26974917519319) (2,1.20717988755259) (3,1.11779710434475) (4,1.00592206184014)};
		
		\end{axis}
		
		\end{tikzpicture}%
	\end{subfigure}
	\caption{Improvement over Base given a budget for SGD using fixed intensities: \mycircle{blue1} 1k, \mycircle{blue2} 100, \mycircle{blue3} 10, \mycircle{blue4} 1, \mytriangle{red2} best EW.}
	\label{fig:fix_gen_sgd}
\end{figure}
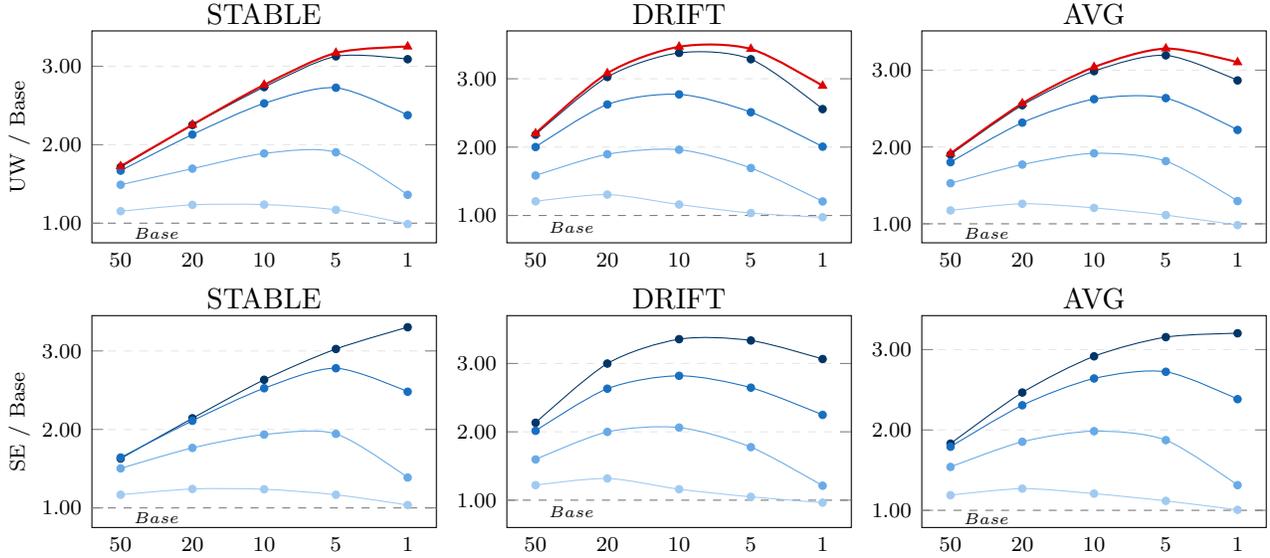

Interestingly, in the case of AHT, the SE strategy was able to provide some small improvements (1-1.1 on average) only for lower budgets and only with $\lambda=1$, so when using the newest instance twice. It shows that this strategy may tend to overfit more than UW or EW as it focuses entirely on one instance without replaying a wider context of instances. Nevertheless, even with $\lambda=1$ the SE strategy provides competitive results for budgets higher than $B=5\%$, when exploitation becomes less important.

Trends presented in Fig. \ref{fig:fix_gen_sgd} confirm that SGD requires much more attention that AHT, especially for lower budgets. One can easily see that the baseline using this classifier benefits enormously from employing instance exploitation. Regardless of a strategy used, our wrapper was able to improve the learning process 2-3 times over the baseline. The characteristic curvature of the trends comes from the fact that there is a sweet spot between possible improvements and available supervision. Analogously to the results for AHT, the EW strategy seems to be the best choice. 

\medskip
\noindent\textbf{Dynamic intensity}. In our next experiment, we investigated if the proposed heuristic for controlling intensity in a dynamic way -- increasing it during drifts and decreasing for stable concepts, based on the current error -- may provide any improvements. Tab. \ref{tab:dyn_gen} presents the average performance of strategies using the dynamic control. We can see that general trends and relations remained the same for both classifiers.

\begin{table}[h]
	\caption{Average kappa for AHT and SGD using dynamic intensities.}
	\vspace*{-0.1cm}
	\centering
	\scalebox{0.8}{
		\begin{subtable}{\textwidth}\leftskip-15pt
			\begin{tabular}[H]{l >{\centering\arraybackslash} m{1.1cm} >{\centering\arraybackslash} m{1.1cm} >{\centering\arraybackslash} m{1.1cm} >{\centering\arraybackslash} m{1.1cm} >{\centering\arraybackslash} m{1.1cm}}
				\textbf{AHT}& $50\%$ & $20\%$ & $10\%$ & $5\%$ & $1\%$\\
				\midrule
				Base & 0.8556 & 0.8311 & 0.8103 & 0.7850 & 0.6916\\
				\hdashline
				UW-1 & \textbf{0.8677} & 0.8453 & 0.8203 & 0.7966 & 0.7092\\
				UW-10 & 0.8763 & \textbf{0.8566} & \textbf{0.8391} & \textbf{0.8245} & \textbf{0.7500}\\
				UW-100 & 0.8612 & 0.8431 & 0.8312 & 0.8060 & 0.7382\\
				UW-1k & 0.8362 & 0.8240 & 0.8067 & 0.7831 & 0.7098\\
				\hdashline
				EW-1 & 0.8652 & 0.8467 & 0.8230 & 0.7994 & 0.7187\\
				EW-10 & \textbf{0.8774} & \textbf{0.8577} & \textbf{0.8431} & \textbf{0.8225} & 0.7596\\
				EW-100 & 0.8621 & 0.8461 & 0.8358 & 0.8207 & \textbf{0.7674}\\
				EW-1k & 0.8472 & 0.8320 & 0.8232 & 0.8018 & 0.7467\\
				\hdashline
				SE-1 & \textbf{0.8668} & \textbf{0.8460} & \textbf{0.8263} & \textbf{0.8004} & \textbf{0.7201}\\
				SE-10 & 0.7058 & 0.6969 & 0.6950 & 0.6852 & 0.6825\\
				SE-100 & 0.4587 & 0.4418 & 0.4380 & 0.4328 & 0.4347\\
				SE-1k & 0.3005 & 0.2872 & 0.2777 & 0.2773 & 0.2864\\
				\bottomrule
			\end{tabular}
			\hspace*{0.25cm}
			\begin{tabular}[H]{l >{\centering\arraybackslash} m{1.1cm} >{\centering\arraybackslash} m{1.1cm} >{\centering\arraybackslash} m{1.1cm} >{\centering\arraybackslash} m{1.1cm} >{\centering\arraybackslash} m{1.1cm}}
				\textbf{SGD}& $50\%$ & $20\%$ & $10\%$ & $5\%$ & $1\%$\\
				\midrule
				Base & 0.4089 & 0.2892 & 0.2229 & 0.2018 & 0.1750\\
				\hdashline
				UW-1 & 0.4731 & 0.3810 & 0.2941 & 0.2339 & 0.1766\\
				UW-10 & 0.5756 & 0.4924 & 0.4186 & 0.3337 & 0.1987\\
				UW-100 & 0.6618 & 0.6292 & 0.5865 & 0.5382 & 0.3592\\
				UW-1k & \textbf{0.7039} & \textbf{0.6888} & \textbf{0.6650} & \textbf{0.6394} & \textbf{0.5385}\\
				\hdashline
				EW-1 & 0.4758 & 0.3819 & 0.2924 & 0.2292 & 0.1795\\
				EW-10 & 0.5821 & 0.5020 & 0.4256 & 0.3393 & 0.1985\\
				EW-100 & 0.6748 & 0.6361 & 0.5983 & 0.5547 & 0.3756\\
				EW-1k & \textbf{0.7101} & \textbf{0.6950} & \textbf{0.6813} & \textbf{0.6601} & \textbf{0.5651}\\
				\hdashline
				SE-1 & 0.4799 & 0.3816 & 0.2987 & 0.2276 & 0.1779\\
				SE-10 & 0.5845 & 0.5059 & 0.4292 & 0.3423 & 0.1980\\
				SE-100 & 0.6632 & 0.6309 & 0.5989 & 0.5547 & 0.3830\\
				SE-1k & \textbf{0.6745} & \textbf{0.6655} & \textbf{0.6544} & \textbf{0.6361} & \textbf{0.5679}\\
				\bottomrule
			\end{tabular}
	\end{subtable}}
	\label{tab:dyn_gen}
\end{table} 

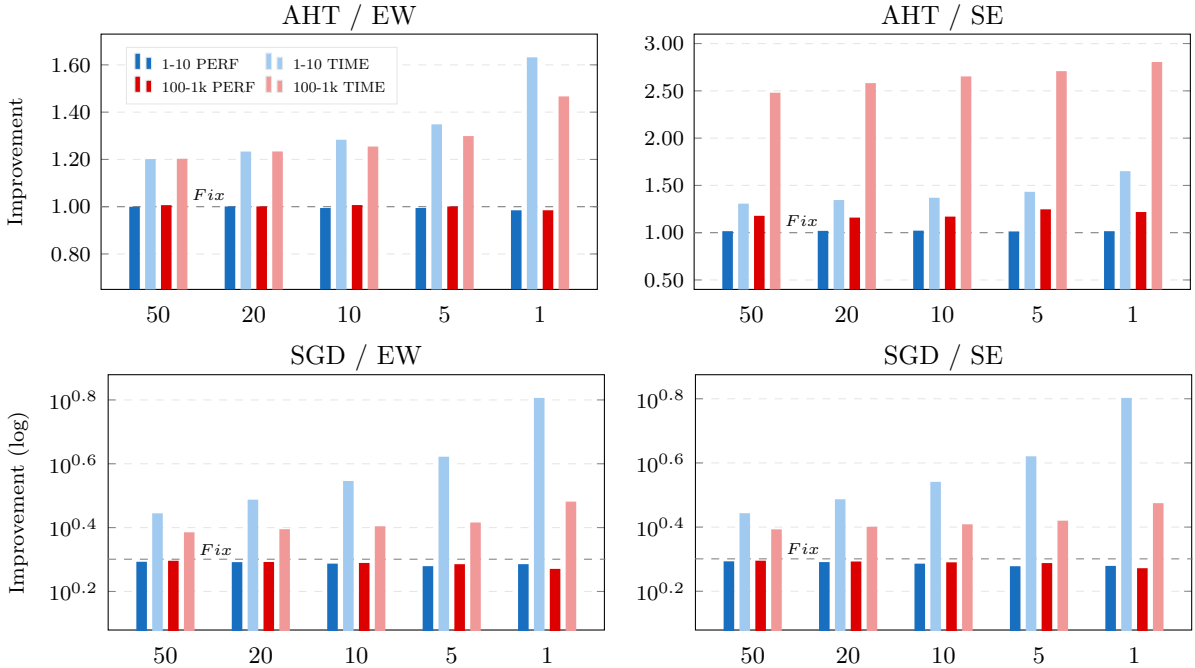
\begin{figure}[h]
	\centering
	\leftskip20pt\begin{subfigure}{0.4\textwidth}
		\begin{tikzpicture}
		\begin{axis}[
		ybar,
		title style={yshift=-0.5ex},
		title={AHT / EW},
		footnotesize,
		width=1.15\textwidth,
		height=0.7\textwidth,
		ymin=0.65,
		ymajorgrids=true,
		ylabel={Improvement},
		ylabel near ticks,
		ylabel style={font=\scriptsize, at={(-0.13,0.5)}},
		y tick label style={
			font=\scriptsize,
			/pgf/number format/.cd,
			fixed,
			fixed zerofill,
			precision=2,
			/tikz/.cd
		},
		extra y ticks=1.0,
		extra y tick labels={},
		extra y tick style={
			ymajorgrids=true,
			ytick style={
				/pgfplots/major tick length=0pt,
			},
			grid style={
				gray,
				dashed,
				/pgfplots/on layer=axis foreground,
			},
		},
		grid style={dashed,white!90!black},
		xtick style={draw=none},
		symbolic x coords={50,20,10,5,1},
		xtick=data,
		enlarge x limits={0.15},
		legend cell align={left},
		legend image post style={scale=0.7},
		legend style={font=\scriptsize,column sep=0.5ex, legend columns=2, at={(0.05,0.95)}, anchor=north west, nodes={scale=0.7}, draw opacity=0.1},
		legend entries={1-10 PERF, 1-10 TIME, 100-1k PERF, 100-1k TIME},
		clip mode=individual
		]
		
		\node[font=\tiny, shift={(0.67cm,0.0cm)}] at (axis cs: 50,1.05) {$Fix$};
		
		\addplot [bar width=4pt, fill=blue2, draw=none] 
		coordinates {(50,0.999596150905184) (20,1.00088121022185) (10,0.994358313119455) (5,0.994389325302087) (1,0.984708400138725)};
		
		\addplot [bar width=4pt, fill=blue4, draw=none] 
		coordinates {(50,1.20129727930037) (20,1.23332886074766) (10,1.28303703346197) (5,1.34828777384647) (1,1.63196664953622)};
		
		\addplot [bar width=4pt, fill=red2, draw=none] 
		coordinates {(50,1.00652733314413) (20,1.0010613577619) (10,1.00689097095556) (5,1.00081679623178) (1,0.984805379179393)};
		
		\addplot [bar width=4pt, fill=red4, draw=none] 
		coordinates {(50,1.20265586100045) (20,1.23363700181112) (10,1.25437435537883) (5,1.2987810600039) (1,1.46653637259363)};
		
		\end{axis}
		\end{tikzpicture}%
		\hspace*{0.4cm}
		\begin{tikzpicture}
		\begin{axis}[
		ybar,
		title style={yshift=-0.5ex},
		title={AHT / SE},
		footnotesize,
		width=1.15\textwidth,
		height=0.7\textwidth,
		ymin=0.4,
		ymax=3.1,
		ymajorgrids=true,
		ylabel near ticks,
		ylabel style={font=\scriptsize},
		y tick label style={
			font=\scriptsize,
			/pgf/number format/.cd,
			fixed,
			fixed zerofill,
			precision=2,
			/tikz/.cd
		},
		extra y ticks=1.0,
		extra y tick labels={},
		extra y tick style={
			ymajorgrids=true,
			ytick style={
				/pgfplots/major tick length=0pt,
			},
			grid style={
				gray,
				dashed,
				/pgfplots/on layer=axis foreground,
			},
		},
		grid style={dashed,white!90!black},
		xtick style={draw=none},
		symbolic x coords={50,20,10,5,1},
		xtick=data,
		enlarge x limits={0.15},
		]
		
		\node[font=\tiny, shift={(0.67cm,0.0cm)}] at (axis cs: 50,1.13) {$Fix$};
		
		\addplot [bar width=4pt, fill=blue2, draw=none] 
		coordinates {(50,1.01588242499491) (20,1.01924789603554) (10,1.02158367420729) (5,1.01311600991915) (1,1.01575106206426)};
		
		\addplot [bar width=4pt, fill=blue4, draw=none] 
		coordinates {(50,1.30718237731729) (20,1.34509738591202) (10,1.36945667909816) (5,1.43225374011569) (1,1.65050798201843)};
		
		\addplot [bar width=4pt, fill=red2, draw=none] 
		coordinates {(50,1.17853271262723) (20,1.15958262193215) (10,1.17004529944882) (5,1.2469291497342) (1,1.21849249055887)};
		
		\addplot [bar width=4pt, fill=red4, draw=none] 
		coordinates {(50,2.48014227082559) (20,2.5817523844745) (10,2.65270439024902) (5,2.70829654729361) (1,2.80526909294148)};
		
		\end{axis}
		
		\end{tikzpicture}
		\begin{tikzpicture}
		\begin{axis}[
		ybar,
		title style={yshift=-0.5ex},
		title={SGD / EW},
		footnotesize,
		width=1.15\textwidth,
		height=0.7\textwidth,
		ymin=1.2,
		ymajorgrids=true,
		ylabel={Improvement (log)},
		ylabel near ticks,
		ylabel style={font=\scriptsize},
		y tick label style={
			font=\scriptsize,
			/pgf/number format/.cd,
			/tikz/.cd
		},
		extra y ticks=2.0,
		extra y tick labels={},
		extra y tick style={
			ymajorgrids=true,
			ytick style={
				/pgfplots/major tick length=0pt,
			},
			grid style={
				gray,
				dashed,
				/pgfplots/on layer=axis foreground,
			},
		},
		ymode=log,
		log basis y={10},
		grid style={dashed,white!90!black},
		xtick style={draw=none},
		symbolic x coords={50,20,10,5,1},
		xtick=data,
		enlarge x limits={0.15},
		clip mode=individual
		]
		
		\node[font=\tiny, shift={(0.67cm,0.0cm)}] at (axis cs: 50,2.15) {$Fix$};
		
		\addplot [bar width=4pt, fill=blue2, draw=none] 
		coordinates {(50,1.963702172650874) (20,1.958470345220187) (10,1.937267724746455) (5,1.902063845546036) (1,1.930053674519836)};
		
		\addplot [bar width=4pt, fill=blue4, draw=none] 
		coordinates {(50,2.78458666227384) (20,3.07329909731747) (10,3.51599105855485) (5,4.1887364018789) (1,6.40270966669437)};
		
		\addplot [bar width=4pt, fill=red2, draw=none] 
		coordinates {(50,1.977765857691693) (20,1.960372925560755) (10,1.947050077695957) (5,1.929460433137187) (1,1.865401084535496)};	
		
		\addplot [bar width=4pt, fill=red4, draw=none] 
		coordinates {(50,2.42844585091421) (20,2.48352714284084) (10,2.53663394424383) (5,2.60614783973116) (1,3.02966432474629)};
		
		\end{axis}
		\end{tikzpicture}%
		\hspace*{0.2cm}
		\begin{tikzpicture}
		\begin{axis}[
		ybar,
		title style={yshift=-0.5ex},
		title={SGD / SE},
		footnotesize,
		width=1.15\textwidth,
		height=0.7\textwidth,
		ymin=1.2,
		ymajorgrids=true,
		ylabel near ticks,
		ylabel style={font=\scriptsize},
		y tick label style={
			font=\scriptsize,
			/pgf/number format/.cd,
			/tikz/.cd
		},
		extra y ticks=2.0,
		extra y tick labels={},
		extra y tick style={
			ymajorgrids=true,
			ytick style={
				/pgfplots/major tick length=0pt,
			},
			grid style={
				gray,
				dashed,
				/pgfplots/on layer=axis foreground,
			},
		},
		ymode=log,
		log basis y={10},
		grid style={dashed,white!90!black},
		xtick style={draw=none},
		symbolic x coords={50,20,10,5,1},
		xtick=data,
		enlarge x limits={0.15},
		clip mode=individual,
		]
		
		\node[font=\tiny, shift={(0.67cm,0.0cm)}] at (axis cs: 50,2.15) {$Fix$};
		
		\addplot [bar width=4pt, fill=blue2, draw=none] 
		coordinates {(50,1.965107622358298) (20,1.954351762952433) (10,1.932440727696415) (5,1.89736840930171) (1,1.90152346591575)};
		
		\addplot [bar width=4pt, fill=blue4, draw=none] 
		coordinates {(50,2.77548330704252) (20,3.06726564363521) (10,3.47574248469325) (5,4.17572766913803) (1,6.34836360644272)};
		
		\addplot [bar width=4pt, fill=red2, draw=none] 
		coordinates {(50,1.973913173401006) (20,1.962651118427877) (10,1.950719396351632) (5,1.940530987457339) (1,1.870394201349555)};
		
		\addplot [bar width=4pt, fill=red4, draw=none] 
		coordinates {(50,2.47159537919051) (20,2.51793793969116) (10,2.56267843083044) (5,2.63027802510628) (1,2.98273495409206)};
		
		\end{axis}
		\end{tikzpicture}
	\end{subfigure}
	\caption{Trade-off between performance and time-consumption for AHT and SGD using dynamic intensity (vs. fixed).}
	\label{fig:gen_dyn_trade}
\end{figure}

More useful information gives us Fig. {\ref{fig:gen_dyn_trade}} presenting the trade-off between obtained change in performance and computational time for less ($\lambda_{max}=\{1, 10\}$) and more risky ($\lambda_{max}=\{100, 1000\}$) exploitation, compared with its fixed-intensity counterparts. It is clear that those adjustments provide speed-up approximately proportional to the error used as the control signal (Eq. \ref{eq:lambda}). In the case of AHT, it is the most significant for the worst-performing risky SE (more than 2.5 times faster) and lower for more reliable strategies and configurations (1.2-1.6). For SGD the speed-up is extremely high (1.5-5.5) due to generally worse performance of the classifier. We can see that the speed-up for both base learners increases as budget gets smaller, since the performance gets worse on average when we limit supervision.

On the other hand, one should notice that the mentioned improvement of running time comes at some cost of predictive performance. It is significant for SGD, especially on very low budgets (about 0.9 of the kappa obtained for fixed intensity), but barely noticeable for AHT with $B\leq10\%$ (0.98 at most) and practically negligible for higher budgets. In fact, it even provided some improvements for SE, which exhibited the worst performance so far (which we can most likely attribute to overfitting). The explanation of these observations is simple -- since the dynamic control may only lower intensity and since SGD gained a substantial amount of enhancement by enabling very intensive instance exploitation, any reduction of it will cause drops in obtained improvements. Due to the fact that AHT is less reliant on our strategies, its adaptation will be impaired to a lesser extent.

\subsubsection{Instance window}

\smallskip
\noindent\textbf{Fixed window size}. In this subsection, we investigate how the size of a sliding window of labeled instances should be configured in order to provide necessary reactivity to new concepts and better performance as a consequence. The average results presented in Tab. \ref{tab:fix_win} definitely outline an easily visible trend for both classifiers -- the less labeled data we have, the smaller sliding window should be used.

\begin{table}[h]
	\caption{Average kappa for AHT and SGD using fixed sliding windows.}
	\vspace*{-0.1cm}
	\centering
	\scalebox{0.8}{
		\begin{subtable}{\textwidth}\leftskip-15pt
			\begin{tabular}[H]{l >{\centering\arraybackslash} m{1.1cm} >{\centering\arraybackslash} m{1.1cm} >{\centering\arraybackslash} m{1.1cm} >{\centering\arraybackslash} m{1.1cm} >{\centering\arraybackslash} m{1.1cm}}
			\textbf{AHT}& $50\%$ & $20\%$ & $10\%$ & $5\%$ & $1\%$\\
			\midrule
			UW-10 & 0.4966 & 0.4918 & 0.4887 & 0.4867 & 0.4864\\
			UW-100 & 0.7340 & 0.7319 & 0.7286 & 0.7195 & \textbf{0.7054}\\
			UW-1k & 0.8169 & \textbf{0.8093} & \textbf{0.7991} & \textbf{0.7713} & 0.6563\\
			UW-10k & \textbf{0.8323} & 0.7921 & 0.7220 & 0.6170 & 0.5323\\
			\hdashline
			EW-10 & 0.4752 & 0.4695 & 0.4641 & 0.4629 & 0.4534\\
			EW-100 & 0.7060 & 0.7053 & 0.7040 & 0.7027 & 0.6860\\
			EW-1k & 0.8142 & 0.8062 & \textbf{0.7981} & \textbf{0.7818} & \textbf{0.7065}\\
			EW-10k & \textbf{0.8446} & \textbf{0.8230} & 0.7786 & 0.7174 & 0.5618\\
			\bottomrule
		\end{tabular}
		\hspace*{0.25cm}
		\begin{tabular}[H]{l >{\centering\arraybackslash} m{1.1cm} >{\centering\arraybackslash} m{1.1cm} >{\centering\arraybackslash} m{1.1cm} >{\centering\arraybackslash} m{1.1cm} >{\centering\arraybackslash} m{1.1cm}}
			\textbf{SGD}& $50\%$ & $20\%$ & $10\%$ & $5\%$ & $1\%$\\
			\midrule	
			UW-10 & 0.6871 & 0.6650 & 0.6386 & 0.6018 & 0.4691\\
			UW-100 & 0.6929 & \textbf{0.6706} & \textbf{0.6443} & \textbf{0.6084} & \textbf{0.4697}\\
			UW-1k & \textbf{0.6953} & 0.6703 & 0.6400 & 0.5941 & 0.4117\\
			UW-10k & 0.6780 & 0.6275 & 0.5567 & 0.4614 & 0.2631\\
			\hdashline
			EW-10 & 0.6857 & 0.6634 & 0.6376 & 0.6022 & 0.4694\\
			EW-100 & 0.6918 & 0.6697 & 0.6432 & \textbf{0.6078} & \textbf{0.4731}\\
			EW-1k & \textbf{0.6952} & \textbf{0.6718} & \textbf{0.6435} & 0.6032 & 0.4505\\
			EW-10k & 0.6878 & 0.6512 & 0.6035 & 0.5349 & 0.3188\\
			\bottomrule
		\end{tabular}
	\end{subtable}}
	\label{tab:fix_win}
\end{table} 

The differences between considered sizes for a given budget are more significant and dynamic for AHT, which is visualized in Fig. \ref{fig:fix_win_aht} showing obtained kappa for both UW and EW. It is clear and intuitive that larger windows ($\omega_{max}=10 000$ and $\omega_{max}=1000$) are more reliable when we have more labeled data and when concepts are stable, since more labeled instances will better generalize a given problem and help prevent overfitting. Once we limit the available supervision, the rate of changes between subsequent instances is more likely to increase (Sec. \ref{sec:risky}). Therefore, in order to keep the representation of current concepts up-to-date we need to replace older instances with newer ones faster, so the window size should be smaller. For example, we can see that while $\omega_{max}=10 000$ is the optimal size if we can label half of a whole stream, it becomes completely useless once we limit the budget to $B=1\%$, when even $\omega_{max}=1000$ becomes less adequate than $\omega_{max}=100$. An observation that the size has to be even smaller for drifting concepts is not surprising -- the problem of developing reactive windows for drifting data has been already covered in many related publications \cite{Bifet:2007}.

\begin{figure}[h]
	\centering
	\leftskip0pt\begin{subfigure}{0.33\textwidth}
		\begin{tikzpicture}
		
		\begin{axis}
		[
		title={STABLE},
		title style={yshift=-1.5ex},
		width=1.05\textwidth,
		height=.75\textwidth,
		ymin=0.55,
		ymax=0.95,
		ytick={0.6, 0.7, 0.8, 0.9},
		ymajorgrids=true,
		ylabel={UW},
		ylabel near ticks,
		y tick label style={
			font=\scriptsize,
			/pgf/number format/.cd,
			fixed,
			fixed zerofill,
			precision=2,
			/tikz/.cd
		},
		ylabel style={font=\scriptsize, at={(-0.15,0.5)}},
		extra y ticks=1.0,
		extra y tick labels={},
		grid style={dashed,white!90!black},
		xtick style={draw=none},
		xtick=data,
		xticklabels={50,20,10,5,1},
		xlabel style={font=\footnotesize},
		x tick label style={font=\scriptsize},
		]
		
		\addplot [	
		draw=red1,smooth,mark=*,mark options={red1,scale=0.7}
		] coordinates {(0,0.8864) (1,0.8443) (2,0.7756) (3,0.6611) (4,0.5739)};
		
		\addplot [	
		draw=red2,smooth,mark=*,mark options={red2,scale=0.7}
		] coordinates {(0,0.8663) (1,0.8596) (2,0.8524) (3,0.8244) (4,0.7063)};
		
		\addplot [	
		draw=red3,smooth,mark=*,mark options={red3,scale=0.7}
		] coordinates {(0,0.7883) (1,0.7877) (2,0.7839) (3,0.7744) (4,0.7677)};
		
		\addplot [	
		draw=red4,smooth,mark=*,mark options={red4,scale=0.7}
		] coordinates {(0,0.5470) (1,0.5406) (2,0.5369) (3,0.5343) (4,0.5382)};
		
		\addplot [	
		draw=blue2,smooth,mark=triangle*,mark options={blue2,scale=0.8},line width=0.8pt
		] coordinates {(0,0.8927) (1,0.8743) (2,0.8481) (3,0.8329) (4,0.7578)};
		
		\end{axis}
		
		\end{tikzpicture}%
		\hspace*{0.05cm}
		\begin{tikzpicture}
		
		\begin{axis}
		[
		title={DRIFT},
		title style={yshift=-1.5ex},
		width=1.05\textwidth,
		height=.75\textwidth,
		ymajorgrids=true,
		ylabel near ticks,
		y tick label style={
			font=\scriptsize,
			/pgf/number format/.cd,
			fixed,
			fixed zerofill,
			precision=2,
			/tikz/.cd
		},
		ylabel style={font=\scriptsize, at={(-0.24,0.5)}},
		extra y ticks=1.0,
		extra y tick labels={},
		grid style={dashed,white!90!black},
		xtick style={draw=none},
		xtick=data,
		xticklabels={50,20,10,5,1},
		xlabel style={font=\footnotesize},
		x tick label style={font=\scriptsize}
		]
		
		\addplot [	
		draw=red1,smooth,mark=*,mark options={red1,scale=0.7}
		] coordinates {(0,0.5648) (1,0.5395) (2,0.5130) (3,0.4971) (4,0.4386)};
		
		\addplot [	
		draw=red2,smooth,mark=*,mark options={red2,scale=0.7}
		] coordinates {(0,0.6891) (1,0.6437) (2,0.5719) (3,0.5129) (4,0.4424)};
		
		\addplot [	
		draw=red3,smooth,mark=*,mark options={red3,scale=0.7}
		] coordinates {(0,0.6732) (1,0.6562) (2,0.6381) (3,0.6057) (4,0.4578)};
		
		\addplot [	
		draw=red4,smooth,mark=*,mark options={red4,scale=0.7}
		] coordinates {(0,0.4646) (1,0.4624) (2,0.4629) (3,0.4479) (4,0.4012)};
		
		\addplot [	
		draw=blue2,smooth,mark=triangle*,mark options={blue2,scale=0.7},line width=0.8pt
		] coordinates {(0,0.7126) (1,0.6779) (2,0.6367) (3,0.6034) (4,0.4870)};
		
		\end{axis}
		
		\end{tikzpicture}%
		\hspace*{0.05cm}
		\begin{tikzpicture}
		
		\begin{axis}
		[
		title={AVG},
		title style={yshift=-1.5ex},
		width=1.05\textwidth,
		height=.75\textwidth,
		ymax=0.85,
		ymajorgrids=true,
		ylabel near ticks,
		y tick label style={
			font=\scriptsize,
			/pgf/number format/.cd,
			fixed,
			fixed zerofill,
			precision=2,
			/tikz/.cd
		},
		ylabel style={font=\scriptsize, at={(-0.24,0.5)}},
		extra y ticks=1.0,
		extra y tick labels={},
		grid style={dashed,white!90!black},
		xtick style={draw=none},
		xtick=data,
		xticklabels={50,20,10,5,1},
		xlabel style={font=\footnotesize},
		x tick label style={font=\scriptsize},
		]
		
		\addplot [	
		draw=red1,smooth,mark=*,mark options={red1,scale=0.7}
		] coordinates {(0,0.7256) (1,0.6919) (2,0.6443) (3,0.5791) (4,0.5063)};
		
		\addplot [	
		draw=red2,smooth,mark=*,mark options={red2,scale=0.7}
		] coordinates {(0,0.7777) (1,0.7516) (2,0.7121) (3,0.6687) (4,0.5744)};
		
		\addplot [	
		draw=red3,smooth,mark=*,mark options={red3,scale=0.7}
		] coordinates {(0,0.7307) (1,0.7220) (2,0.7110) (3,0.6901) (4,0.6128)};
		
		\addplot [	
		draw=red4,smooth,mark=*,mark options={red4,scale=0.7}
		] coordinates {(0,0.5058) (1,0.5015) (2,0.4999) (3,0.4911) (4,0.4697)};
		
		\addplot [	
		draw=blue2,smooth,mark=triangle*,mark options={blue2,scale=0.7},line width=0.8pt
		] coordinates {(0,0.7877) (1,0.7663) (2,0.7424) (3,0.7091) (4,0.6156)};
		
		\end{axis}
		
		\end{tikzpicture}%
	\end{subfigure}\\
	\begin{subfigure}{0.33\textwidth}
		\begin{tikzpicture}
		
		\begin{axis}
		[
		title={STABLE},
		title style={yshift=-1.5ex},
		width=1.05\textwidth,
		height=.75\textwidth,
		ymin=0.55,
		ymax=0.95,
		ytick={0.6, 0.7, 0.8, 0.9},
		ymajorgrids=true,
		ylabel={EW},
		ylabel near ticks,
		y tick label style={
			font=\scriptsize,
			/pgf/number format/.cd,
			fixed,
			fixed zerofill,
			precision=2,
			/tikz/.cd
		},
		ylabel style={font=\scriptsize, at={(-0.15,0.5)}},
		extra y ticks=1.0,
		extra y tick labels={},
		grid style={dashed,white!90!black},
		xtick style={draw=none},
		xtick=data,
		xticklabels={50,20,10,5,1},
		xlabel style={font=\footnotesize},
		x tick label style={font=\scriptsize},
		]
		
		\addplot [	
		draw=red1,smooth,mark=*,mark options={red1,scale=0.7}
		] coordinates {(0,0.8927) (1,0.8743) (2,0.8344) (3,0.7686) (4,0.5951)};
		
		\addplot [	
		draw=red2,smooth,mark=*,mark options={red2,scale=0.7}
		] coordinates {(0,0.8629) (1,0.8548) (2,0.8481) (3,0.8329) (4,0.7578)};
		
		\addplot [	
		draw=red3,smooth,mark=*,mark options={red3,scale=0.7}
		] coordinates {(0,0.7594) (1,0.7588) (2,0.7578) (3,0.7572) (4,0.7443)};
		
		\addplot [	
		draw=red4,smooth,mark=*,mark options={red4,scale=0.7}
		] coordinates {(0,0.5263) (1,0.5194) (2,0.5130) (3,0.5130) (4,0.5046)};
		
		
		\end{axis}
		
		\end{tikzpicture}%
		\hspace*{0.05cm}
		\begin{tikzpicture}
		
		\begin{axis}
		[
		title={DRIFT},
		title style={yshift=-1.5ex},
		width=1.05\textwidth,
		height=.75\textwidth,
		ymajorgrids=true,
		ylabel near ticks,
		y tick label style={
			font=\scriptsize,
			/pgf/number format/.cd,
			fixed,
			fixed zerofill,
			precision=2,
			/tikz/.cd
		},
		ylabel style={font=\scriptsize, at={(-0.24,0.5)}},
		extra y ticks=1.0,
		extra y tick labels={},
		grid style={dashed,white!90!black},
		xtick style={draw=none},
		xtick=data,
		xticklabels={50,20,10,5,1},
		xlabel style={font=\footnotesize},
		x tick label style={font=\scriptsize}
		]
		
		\addplot [	
		draw=red1,smooth,mark=*,mark options={red1,scale=0.7}
		] coordinates {(0,0.6259) (1,0.5656) (2,0.5335) (3,0.5047) (4,0.4486)};
		
		\addplot [	
		draw=red2,smooth,mark=*,mark options={red2,scale=0.7}
		] coordinates {(0,0.7126) (1,0.6779) (2,0.6367) (3,0.5854) (4,0.4613)};
		
		\addplot [	
		draw=red3,smooth,mark=*,mark options={red3,scale=0.7}
		] coordinates {(0,0.6477) (1,0.6433) (2,0.6310) (3,0.6034) (4,0.4870)};
		
		\addplot [	
		draw=red4,smooth,mark=*,mark options={red4,scale=0.7}
		] coordinates {(0,0.4480) (1,0.4400) (2,0.4286) (3,0.4203) (4,0.3780)};
		
		
		\end{axis}
		
		\end{tikzpicture}%
		\hspace*{0.05cm}
		\begin{tikzpicture}
		
		\begin{axis}
		[
		title={AVG},
		title style={yshift=-1.5ex},
		width=1.05\textwidth,
		height=.75\textwidth,
		ymin=0.35,
		ymax=0.85,
		ytick={0.4,0.5,0.6,0.7,0.8},
		ymajorgrids=true,
		ylabel near ticks,
		y tick label style={
			font=\scriptsize,
			/pgf/number format/.cd,
			fixed,
			fixed zerofill,
			precision=2,
			/tikz/.cd
		},
		ylabel style={font=\scriptsize, at={(-0.24,0.5)}},
		extra y ticks=1.0,
		extra y tick labels={},
		grid style={dashed,white!90!black},
		xtick style={draw=none},
		xtick=data,
		xticklabels={50,20,10,5,1},
		xlabel style={font=\footnotesize},
		x tick label style={font=\scriptsize},
		]
		
		\addplot [	
		draw=red1,smooth,mark=*,mark options={red1,scale=0.7}
		] coordinates {(0,0.7593) (1,0.7200) (2,0.6840) (3,0.6367) (4,0.5219)};
		
		\addplot [	
		draw=red2,smooth,mark=*,mark options={red2,scale=0.7}
		] coordinates {(0,0.7877) (1,0.7663) (2,0.7424) (3,0.7091) (4,0.6095)};
		
		\addplot [	
		draw=red3,smooth,mark=*,mark options={red3,scale=0.7}
		] coordinates {(0,0.7035) (1,0.7010) (2,0.6944) (3,0.6803) (4,0.6156)};
		
		\addplot [	
		draw=red4,smooth,mark=*,mark options={red4,scale=0.7}
		] coordinates {(0,0.4871) (1,0.4797) (2,0.4708) (3,0.4667) (4,0.4413)};
		
		
		\end{axis}
		
		\end{tikzpicture}%
	\end{subfigure}
	\caption{Average kappa given a budget for AHT using fixed sliding windows: \mycircle{red1} 10k, \mycircle{red2} 1k, \mycircle{red3} 100, \mycircle{red4} 10, \mytriangle{blue2} best EW.}
	\label{fig:fix_win_aht}
\end{figure}
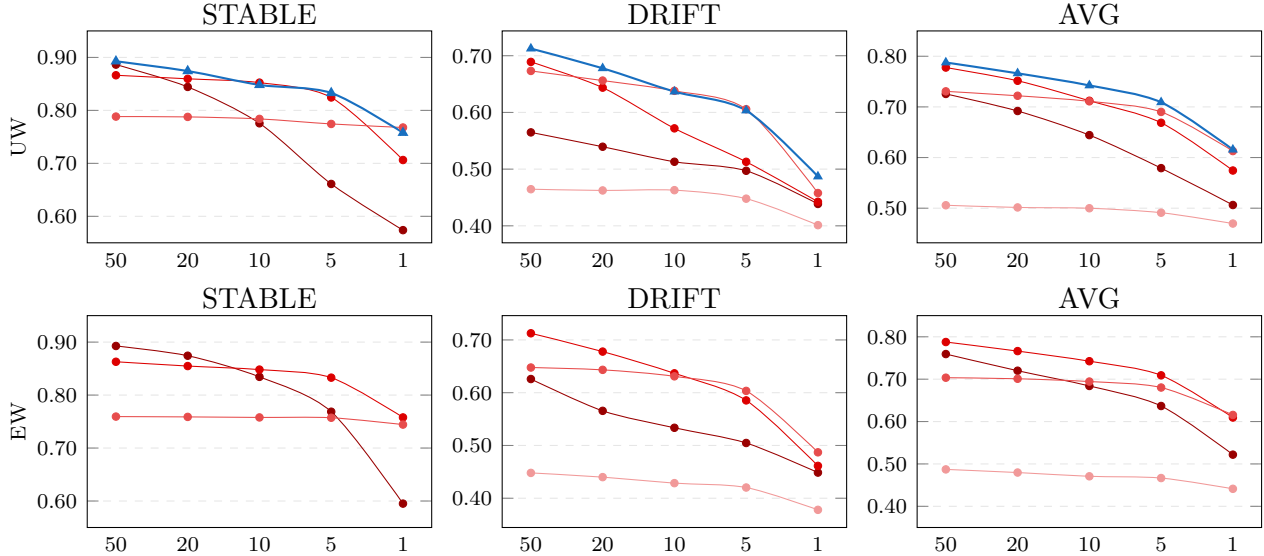

\begin{figure}[h]
	\centering
	\leftskip0pt\begin{subfigure}{0.33\textwidth}
		\begin{tikzpicture}
		
		\begin{axis}
		[
		title={STABLE},
		title style={yshift=-1.5ex},
		width=1.05\textwidth,
		height=.75\textwidth,
		ymin=0.15,
		ytick={0.2, 0.35, 0.5, 0.65},
		ymajorgrids=true,
		ylabel={UW},
		ylabel near ticks,
		y tick label style={
			font=\scriptsize,
			/pgf/number format/.cd,
			fixed,
			fixed zerofill,
			precision=2,
			/tikz/.cd
		},
		ylabel style={font=\scriptsize, at={(-0.15,0.5)}},
		extra y ticks=1.0,
		extra y tick labels={},
		grid style={dashed,white!90!black},
		xtick style={draw=none},
		xtick=data,
		xticklabels={50,20,10,5,1},
		xlabel style={font=\footnotesize},
		x tick label style={font=\scriptsize},
		]
		
		\addplot [	
		draw=red1,smooth,mark=*,mark options={red1,scale=0.7}
		] coordinates {(0,0.6832) (1,0.6384) (2,0.5682) (3,0.4632) (4,0.2613)};
		
		\addplot [	
		draw=red2,smooth,mark=*,mark options={red2,scale=0.7}
		] coordinates {(0,0.6889) (1,0.6700) (2,0.6463) (3,0.6023) (4,0.4125)};
		
		\addplot [	
		draw=red3,smooth,mark=*,mark options={red3,scale=0.7}
		] coordinates {(0,0.6846) (1,0.6669) (2,0.6453) (3,0.6124) (4,0.4741)};
		
		\addplot [	
		draw=red4,smooth,mark=*,mark options={red4,scale=0.7}
		] coordinates {(0,0.6778) (1,0.6603) (2,0.6390) (3,0.6053) (4,0.4710)};
		
		\addplot [	
		draw=blue2,smooth,mark=triangle*,mark options={blue2,scale=0.7},line width=0.8pt
		] coordinates {(0,0.6876) (1,0.6695) (2,0.6461) (3,0.6114) (4,0.4756)};
		
		\end{axis}
		
		\end{tikzpicture}%
		\hspace*{0.05cm}
		\begin{tikzpicture}
		
		\begin{axis}
		[
		title={DRIFT},
		title style={yshift=-1.5ex},
		width=1.05\textwidth,
		height=.75\textwidth,
		ymin=0.13,
		ymax=0.58,
		ytick={0.2,0.3,0.4,0.5},
		ymajorgrids=true,
		ylabel near ticks,
		y tick label style={
			font=\scriptsize,
			/pgf/number format/.cd,
			fixed,
			fixed zerofill,
			precision=2,
			/tikz/.cd
		},
		ylabel style={font=\scriptsize, at={(-0.24,0.5)}},
		extra y ticks=1.0,
		extra y tick labels={},
		grid style={dashed,white!90!black},
		xtick style={draw=none},
		xtick=data,
		xticklabels={50,20,10,5,1},
		xlabel style={font=\footnotesize},
		x tick label style={font=\scriptsize}
		]
		
		\addplot [	
		draw=red1,smooth,mark=*,mark options={red1,scale=0.7}
		] coordinates {(0,0.4055) (1,0.3554) (2,0.3301) (3,0.3082) (4,0.2324)};
		
		\addplot [	
		draw=red2,smooth,mark=*,mark options={red2,scale=0.7}
		] coordinates {(0,0.5357) (1,0.4623) (2,0.3884) (3,0.3465) (4,0.2627)};
		
		\addplot [	
		draw=red3,smooth,mark=*,mark options={red3,scale=0.7}
		] coordinates {(0,0.5536) (1,0.5041) (2,0.4590) (3,0.4072) (4,0.3005)};
		
		\addplot [	
		draw=red4,smooth,mark=*,mark options={red4,scale=0.7}
		] coordinates {(0,0.5503) (1,0.5021) (2,0.4566) (3,0.4151) (4,0.3120)};
		
		\addplot [	
		draw=blue2,smooth,mark=triangle*,mark options={blue2,scale=0.7},line width=0.8pt
		] coordinates {(0,0.5558) (1,0.5066) (2,0.4612) (3,0.4132) (4,0.3100)};
		
		\end{axis}
		
		\end{tikzpicture}%
		\hspace*{0.05cm}
		\begin{tikzpicture}
		
		\begin{axis}
		[
		title={AVG},
		title style={yshift=-1.5ex},
		width=1.05\textwidth,
		height=.75\textwidth,
		ymin=0.15,
		ymax=0.7,
		ytick={0.2, 0.35, 0.5, 0.65},
		ymajorgrids=true,
		ylabel near ticks,
		y tick label style={
			font=\scriptsize,
			/pgf/number format/.cd,
			fixed,
			fixed zerofill,
			precision=2,
			/tikz/.cd
		},
		ylabel style={font=\scriptsize, at={(-0.24,0.5)}},
		extra y ticks=1.0,
		extra y tick labels={},
		grid style={dashed,white!90!black},
		xtick style={draw=none},
		xtick=data,
		xticklabels={50,20,10,5,1},
		xlabel style={font=\footnotesize},
		x tick label style={font=\scriptsize}
		]
		
		\addplot [	
		draw=red1,smooth,mark=*,mark options={red1,scale=0.7}
		] coordinates {(0,0.5443) (1,0.4969) (2,0.4492) (3,0.3857) (4,0.2469)};
		
		\addplot [	
		draw=red2,smooth,mark=*,mark options={red2,scale=0.7}
		] coordinates {(0,0.6123) (1,0.5662) (2,0.5173) (3,0.4744) (4,0.3376)};
		
		\addplot [	
		draw=red3,smooth,mark=*,mark options={red3,scale=0.7}
		] coordinates {(0,0.6191) (1,0.5855) (2,0.5521) (3,0.5098) (4,0.3873)};
		
		\addplot [	
		draw=red4,smooth,mark=*,mark options={red4,scale=0.7}
		] coordinates {(0,0.6141) (1,0.5812) (2,0.5478) (3,0.5102) (4,0.3915)};
		
		\addplot [	
		draw=blue2,smooth,mark=triangle*,mark options={blue2,scale=0.7},line width=0.8pt
		] coordinates {(0,0.6194) (1,0.5861) (2,0.5522) (3,0.5123) (4,0.3928)};
		
		\end{axis}
		
		\end{tikzpicture}
	\end{subfigure}\\
	\begin{subfigure}{0.33\textwidth}
	\begin{tikzpicture}
	
	\begin{axis}
	[
	title={STABLE},
	title style={yshift=-1.5ex},
	width=1.05\textwidth,
	height=.75\textwidth,
	ymin=0.15,
	ytick={0.2, 0.35, 0.5, 0.65},
	ymajorgrids=true,
	ylabel={EW},
	ylabel near ticks,
	y tick label style={
		font=\scriptsize,
		/pgf/number format/.cd,
		fixed,
		fixed zerofill,
		precision=2,
		/tikz/.cd
	},
	ylabel style={font=\scriptsize, at={(-0.15,0.5)}},
	extra y ticks=1.0,
	extra y tick labels={},
	grid style={dashed,white!90!black},
	xtick style={draw=none},
	xtick=data,
	xticklabels={50,20,10,5,1},
	xlabel style={font=\footnotesize},
	x tick label style={font=\scriptsize},
	]
	
	\addplot [	
	draw=red1,smooth,mark=*,mark options={red1,scale=0.7}
	] coordinates {(0,0.6868) (1,0.6569) (2,0.6116) (3,0.5407) (4,0.3111)};
	
	\addplot [	
	draw=red2,smooth,mark=*,mark options={red2,scale=0.7}
	] coordinates {(0,0.6876) (1,0.6695) (2,0.6461) (3,0.6090) (4,0.4539)};
	
	\addplot [	
	draw=red3,smooth,mark=*,mark options={red3,scale=0.7}
	] coordinates {(0,0.6831) (1,0.6656) (2,0.6431) (3,0.6114) (4,0.4756)};
	
	\addplot [	
	draw=red4,smooth,mark=*,mark options={red4,scale=0.7}
	] coordinates {(0,0.6761) (1,0.6590) (2,0.6374) (3,0.6049) (4,0.4689)};
	
	
	\end{axis}
	
	\end{tikzpicture}%
	\hspace*{0.05cm}
	\begin{tikzpicture}
	
	\begin{axis}
	[
	title={DRIFT},
	title style={yshift=-1.5ex},
	width=1.05\textwidth,
	height=.75\textwidth,
	ymajorgrids=true,
	ylabel near ticks,
	y tick label style={
		font=\scriptsize,
		/pgf/number format/.cd,
		fixed,
		fixed zerofill,
		precision=2,
		/tikz/.cd
	},
	ylabel style={font=\scriptsize, at={(-0.24,0.5)}},
	extra y ticks=1.0,
	extra y tick labels={},
	grid style={dashed,white!90!black},
	xtick style={draw=none},
	xtick=data,
	xticklabels={50,20,10,5,1},
	xlabel style={font=\footnotesize},
	x tick label style={font=\scriptsize}
	]
	
	\addplot [	
	draw=red1,smooth,mark=*,mark options={red1,scale=0.7}
	] coordinates {(0,0.4700) (1,0.3888) (2,0.3544) (3,0.3230) (4,0.2499)};
	
	\addplot [	
	draw=red2,smooth,mark=*,mark options={red2,scale=0.7}
	] coordinates {(0,0.5479) (1,0.4873) (2,0.4310) (3,0.3725) (4,0.2707)};
	
	\addplot [	
	draw=red3,smooth,mark=*,mark options={red3,scale=0.7}
	] coordinates {(0,0.5558) (1,0.5066) (2,0.4612) (3,0.4132) (4,0.3100)};
	
	\addplot [	
	draw=red4,smooth,mark=*,mark options={red4,scale=0.7}
	] coordinates {(0,0.5498) (1,0.5016) (2,0.4549) (3,0.4124) (4,0.3090)};
	
	
	\end{axis}
	
	\end{tikzpicture}%
	\hspace*{0.05cm}
	\begin{tikzpicture}
	
	\begin{axis}
	[
	title={AVG},
	title style={yshift=-1.5ex},
	width=1.05\textwidth,
	height=.75\textwidth,
	ymajorgrids=true,
	ylabel near ticks,
	y tick label style={
		font=\scriptsize,
		/pgf/number format/.cd,
		fixed,
		fixed zerofill,
		precision=2,
		/tikz/.cd
	},
	ylabel style={font=\scriptsize, at={(-0.24,0.5)}},
	extra y ticks=1.0,
	extra y tick labels={},
	grid style={dashed,white!90!black},
	xtick style={draw=none},
	xtick=data,
	xticklabels={50,20,10,5,1},
	xlabel style={font=\footnotesize},
	x tick label style={font=\scriptsize},
	]
	
	\addplot [	
	draw=red1,smooth,mark=*,mark options={red1,scale=0.7}
	] coordinates {(0,0.5784) (1,0.5228) (2,0.4830) (3,0.4319) (4,0.2805)};
	
	\addplot [	
	draw=red2,smooth,mark=*,mark options={red2,scale=0.7}
	] coordinates {(0,0.6178) (1,0.5784) (2,0.5386) (3,0.4908) (4,0.3623)};
	
	\addplot [	
	draw=red3,smooth,mark=*,mark options={red3,scale=0.7}
	] coordinates {(0,0.6194) (1,0.5861) (2,0.5522) (3,0.5123) (4,0.3928)};
	
	\addplot [	
	draw=red4,smooth,mark=*,mark options={red4,scale=0.7}
	] coordinates {(0,0.6130) (1,0.5803) (2,0.5462) (3,0.5086) (4,0.3890)};
	
	
	\end{axis}
	
	\end{tikzpicture}%
	\end{subfigure}
	\caption{Average kappa given a budget for SGD using fixed sliding windows: \mycircle{red1} 10k, \mycircle{red2} 1k, \mycircle{red3} 100, \mycircle{red4} 10, \mytriangle{blue2} best EW.}
	\label{fig:fix_win_sgd}
\end{figure}
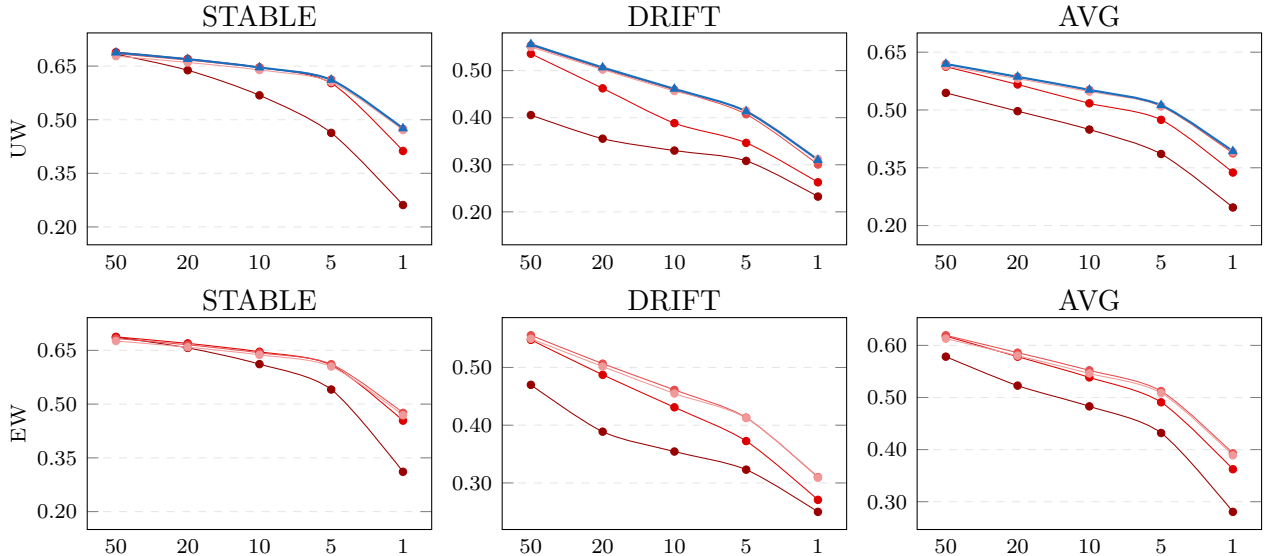

In the case of SGD, the differences between window sizes are slightly smaller (Fig. \ref{fig:fix_win_sgd}), since the overall performance of this classifier is predominantly reliant on a proper choice of the level of intensity. Also, it seems that SGD works best with $\omega_{max}\leq100$, which suggests that it prefers intensive updates based on the most recent data.

Finally, for both classifiers, the EW strategy was able to reduce the negative impact of too large window sizes for smaller budgets -- one can notice that the curves for $\omega_{max}=10000$ and $\omega_{max}=1000$ are slightly elevated compared with UW. The reason for that is the inherent focus of EW on the most recent instances, which alleviates the problem of storing too old instances, which we assumed in Sec. \ref{sec:strategies}. Regardless of that, we did not find any significant difference between EW and UW in this experiment, which suggests that intensity is a more impactful factor.

\medskip
\noindent\textbf{Dynamic window size}. Analogously to the experiments focused on intensity, we studied the impact of controlling the window size in a dynamic way. In Tab. \ref{tab:dyn_win} one can see that while the general trends and relations remain, again, unchanged, using the size returned by ADWIN is definitely the most reliable solution for AHT, providing from 0.02 to more than 0.15 higher kappa values than any other window. For SGD differences between smaller sizes and ADWIN are hardly significant for $B \geq 20\%$ and in favor of the former for $B \leq 10\%$.

\begin{table}[b]
	\caption{Average kappa for AHT and SGD using dynamic sliding windows.}
	\vspace*{-0.1cm}
	\centering
	\scalebox{0.8}{
		\begin{subtable}{\textwidth}\leftskip-30pt
			\begin{tabular}[H]{l >{\centering\arraybackslash} m{1.1cm} >{\centering\arraybackslash} m{1.1cm} >{\centering\arraybackslash} m{1.1cm} >{\centering\arraybackslash} m{1.1cm} >{\centering\arraybackslash} m{1.1cm}}
				\textbf{AHT}& $50\%$ & $20\%$ & $10\%$ & $5\%$ & $1\%$\\
				\midrule
				UW-10 & 0.4439 & 0.4295 & 0.4350 & 0.4281 & 0.4263\\
				UW-100 & 0.7161 & 0.7128 & 0.7102 & 0.7068 & 0.6896\\
				UW-1k & 0.8161 & 0.8104 & 0.7988 & 0.7845 & 0.6976\\
				UW-10k & 0.8416 & 0.8078 & 0.7651 & 0.6896 & 0.5585\\
				UW-ADW & \textbf{0.8603} & \textbf{0.8452} & \textbf{0.8243} & \textbf{0.8098} & \textbf{0.7418}\\
				\hdashline
				EW-10 & 0.4424 & 0.4252 & 0.4235 & 0.4259 & 0.4250\\
				EW-100 & 0.6798 & 0.6779 & 0.6778 & 0.6748 & 0.6730\\
				EW-1k & 0.8081 & 0.8062 & 0.7985 & 0.7919 & 0.7356\\
				EW-10k & 0.8502 & 0.8306 & 0.8068 & 0.7557 & 0.6209\\
				EW-ADW & \textbf{0.8598} & \textbf{0.8483} & \textbf{0.8312} & \textbf{0.8202} & \textbf{0.7717}\\
				\bottomrule
			\end{tabular}
			\hspace*{0.25cm}
			\begin{tabular}[H]{l >{\centering\arraybackslash} m{1.1cm} >{\centering\arraybackslash} m{1.1cm} >{\centering\arraybackslash} m{1.1cm} >{\centering\arraybackslash} m{1.1cm} >{\centering\arraybackslash} m{1.1cm}}
				\textbf{SGD}& $50\%$ & $20\%$ & $10\%$ & $5\%$ & $1\%$\\
				\midrule
				UW-10 & 0.6864 & 0.6642 & 0.6395 & 0.6023 & 0.4664\\
				UW-100 & 0.6940 & 0.6717 & \textbf{0.6459} & \textbf{0.6088} & \textbf{0.4712}\\
				UW-1k & \textbf{0.6966} & \textbf{0.6730} & 0.6451 & 0.6041 & 0.4467\\
				UW-10k & 0.6831 & 0.6418 & 0.5887 & 0.5154 & 0.3180\\
				UW-ADW & 0.6892 & 0.6568 & 0.6275 & 0.5799 & 0.4379\\
				\hdashline
				EW-10 & 0.6851 & 0.6631 & 0.6373 & 0.6003 & 0.4679\\
				EW-100 & 0.6918 & 0.6694 & 0.6445 & 0.6065 & \textbf{0.4731}\\
				EW-1k & \textbf{0.6966} & \textbf{0.6731} & \textbf{0.6474} & \textbf{0.6072} & 0.4643\\
				EW-10k & 0.6908 & 0.6594 & 0.6208 & 0.5631 & 0.3823\\
				EW-ADW & 0.6941 & 0.6682 & 0.6380 & 0.5968 & 0.4572\\
				\bottomrule
			\end{tabular}
	\end{subtable}}
	\label{tab:dyn_win}
\end{table} 

\begin{figure}[h]
	\centering
	\leftskip15pt\begin{subfigure}{0.4\textwidth}
		\begin{tikzpicture}
		\begin{axis}[
		ybar,
		title style={yshift=-0.5ex},
		title={AHT / EW},
		footnotesize,
		width=1.15\textwidth,
		height=0.7\textwidth,
		ymin=0.7,
		ymax=1.9,
		ymajorgrids=true,
		ylabel={Improvement},
		ylabel near ticks,
		ylabel style={font=\scriptsize},
		y tick label style={
			font=\scriptsize,
			/pgf/number format/.cd,
			fixed,
			fixed zerofill,
			precision=2,
			/tikz/.cd
		},
		extra y ticks=1.0,
		extra y tick labels={},
		extra y tick style={
			ymajorgrids=true,
			ytick style={
				/pgfplots/major tick length=0pt,
			},
			grid style={
				gray,
				dashed,
				/pgfplots/on layer=axis foreground,
			},
		},
		grid style={dashed,white!90!black},
		xtick style={draw=none},
		symbolic x coords={50,20,10,5,1},
		xtick=data,
		clip mode=individual,
		enlarge x limits={0.15},
		legend cell align={left},
		legend image post style={scale=0.7},
		legend style={font=\scriptsize,column sep=0.5ex, legend columns=2, at={(0.05,0.95)}, anchor=north west, nodes={scale=0.75}, draw opacity=0.1},
		legend entries={10-100 PERF, 10-100 MEM, 1k-10k PERF, 1k-10k MEM}
		]
		
		\node[font=\tiny, shift={(0.7cm,0.0cm)}] at (axis cs: 50, 1.05) {$Fix$};
		
		\addplot [bar width=4pt, fill=blue2, draw=none] 
		coordinates {(50,0.94571396600022) (20,0.930954802315649) (10,0.938624007228922) (5,0.949166352334759) (1,0.960896349218413)};
		
		\addplot [bar width=4pt, fill=blue4, draw=none] 
		coordinates {(50,1.44739098375749) (20,1.44573978586891) (10,1.44879613005496) (5,1.4482458379837) (1,1.45347474700451)};
		
		\addplot [bar width=4pt, fill=red2, draw=none] 
		coordinates {(50,1.01487498199656) (20,1.01908104111032) (10,1.02745152701351) (5,1.03819825650222) (1,1.05729585518725)};
		
		\addplot [bar width=4pt, fill=red4, draw=none] 
		coordinates {(50,1.11208498504937) (20,1.12077869842001) (10,1.13398190323755) (5,1.15780775315885) (1,1.25454630002006)};
		
		\end{axis}
		\end{tikzpicture}%
		\hspace*{0.2cm}
		\begin{tikzpicture}
		\begin{axis}[
		ybar,
		title style={yshift=-0.5ex},
		title={AHT / EW-ADW},
		footnotesize,
		width=1.15\textwidth,
		height=0.7\textwidth,
		ymin=0.2,
		ymax=2.2,
		ytick={0.3, 0.6, 0.9, 1.2, 1.5, 1.8, 2.1},
		ymajorgrids=true,
		ylabel near ticks,
		ylabel style={font=\scriptsize},
		y tick label style={
			font=\scriptsize,
			/pgf/number format/.cd,
			fixed,
			fixed zerofill,
			precision=2,
			/tikz/.cd
		},
		extra y ticks=1.0,
		extra y tick labels={},
		extra y tick style={
			ymajorgrids=true,
			ytick style={
				/pgfplots/major tick length=0pt,
			},
			grid style={
				gray,
				dashed,
				/pgfplots/on layer=axis foreground,
			},
		},
		grid style={dashed,white!90!black},
		xtick style={draw=none},
		symbolic x coords={50,20,10,5,1},
		xtick=data,
		enlarge x limits={0.15},
		clip mode=individual,
		legend cell align={left},
		legend image post style={scale=0.7},
		legend style={font=\scriptsize,column sep=0.5ex, legend columns=2, at={(0.05,0.95)}, anchor=north west, nodes={scale=0.75}, draw opacity=0.1},
		legend entries={ADW PERF, ADW MEM}
		]
		
		\node[font=\tiny, shift={(0.5cm,0.0cm)}] at (axis cs: 50, 1.08) {$Fix$};
		
		\addplot [bar width=4pt, fill=green2, draw=none] 
		coordinates {(50,1.0616998090226) (20,1.0686114212596) (10,1.07266729358845) (5,1.1024933349337) (1,1.11179251177111)};
		
		\addplot [bar width=4pt, fill=green4, draw=none] 
		coordinates {(50,0.373084236837807) (20,0.709938871900499) (10,1.14203157930311) (5,1.70143194992726) (1,1.90308166404234)};
		
		\end{axis}
		\end{tikzpicture}
		\begin{tikzpicture}
		\begin{axis}[
		ybar,
		title style={yshift=-0.5ex},
		title={SGD / EW},
		footnotesize,
		width=1.15\textwidth,
		height=0.7\textwidth,
		ymin=0.7,
		ymajorgrids=true,
		ylabel={Improvement},
		ylabel near ticks,
		ylabel style={font=\scriptsize},
		y tick label style={
			font=\scriptsize,
			/pgf/number format/.cd,
			fixed,
			fixed zerofill,
			precision=2,
			/tikz/.cd
		},
		extra y ticks=1.0,
		extra y tick labels={},
		extra y tick style={
			ymajorgrids=true,
			ytick style={
				/pgfplots/major tick length=0pt,
			},
			grid style={
				gray,
				dashed,
				/pgfplots/on layer=axis foreground,
			},
		},
		grid style={dashed,white!90!black},
		xtick style={draw=none},
		symbolic x coords={50,20,10,5,1},
		xtick=data,
		enlarge x limits={0.15},
		clip mode=individual
		]
		
		\node[font=\tiny, shift={(0.7cm,0.0cm)}] at (axis cs: 50, 1.05) {$Fix$};
		
		\addplot [bar width=4pt, fill=blue2, draw=none] 
		coordinates {(50,0.998446830641616) (20,0.999428009274032) (10,1.002778874198) (5,1.00041484860534) (1,0.996949006582305)};
		
		\addplot [bar width=4pt, fill=blue4, draw=none] 
		coordinates {(50,1.40930039426965) (20,1.43593450187205) (10,1.46903125571606) (5,1.50932961505217) (1,1.69441570476055)};
		
		\addplot [bar width=4pt, fill=red2, draw=none] 
		coordinates {(50,1.01673032959242) (20,1.02775517652381) (10,1.02840765873539) (5,1.03763640484149) (1,1.11675106608308)};
		
		\addplot [bar width=4pt, fill=red4, draw=none] 
		coordinates {(50,1.35071291136052) (20,1.37862300681295) (10,1.41188421615588) (5,1.46306687232385) (1,1.69465759277365)};
		
		\end{axis}
		\end{tikzpicture}%
		\hspace*{0.2cm}
		\begin{tikzpicture}
		\begin{axis}[
		ybar,
		title style={yshift=-0.5ex},
		title={SGD / EW-ADW},
		footnotesize,
		width=1.15\textwidth,
		height=0.7\textwidth,
		ymin=0.2,
		ymajorgrids=true,
		ylabel near ticks,
		ylabel style={font=\scriptsize},
		y tick label style={
			font=\scriptsize,
			/pgf/number format/.cd,
			fixed,
			fixed zerofill,
			precision=2,
			/tikz/.cd
		},
		extra y ticks=1.0,
		extra y tick labels={},
		extra y tick style={
			ymajorgrids=true,
			ytick style={
				/pgfplots/major tick length=0pt,
			},
			grid style={
				gray,
				dashed,
				/pgfplots/on layer=axis foreground,
			},
		},
		grid style={dashed,white!90!black},
		xtick style={draw=none},
		symbolic x coords={50,20,10,5,1},
		xtick=data,
		enlarge x limits={0.15},
		clip mode=individual
		]
		
		\node[font=\tiny, shift={(0.5cm,0.0cm)}] at (axis cs: 50, 1.1) {$Fix$};
		
		\addplot [bar width=4pt, fill=green2, draw=none] 
		coordinates {(50,1.04209435782583) (20,1.07331914266605) (10,1.09813542793405) (5,1.11718413071174) (1,1.11981151051368)};
		
		\addplot [bar width=4pt, fill=green4, draw=none] 
		coordinates {(50,0.29109294938429) (20,0.679699113636252) (10,1.08189543561764) (5,1.87623073044891) (1,2.4882381336009)};
		
		\end{axis}
		\end{tikzpicture}
	\end{subfigure}
	\caption{Trade-off between performance and memory given a budget for AHT and SGD using dynamic windows (vs. fixed).}
	\label{fig:win_dyn_trade}
\end{figure}
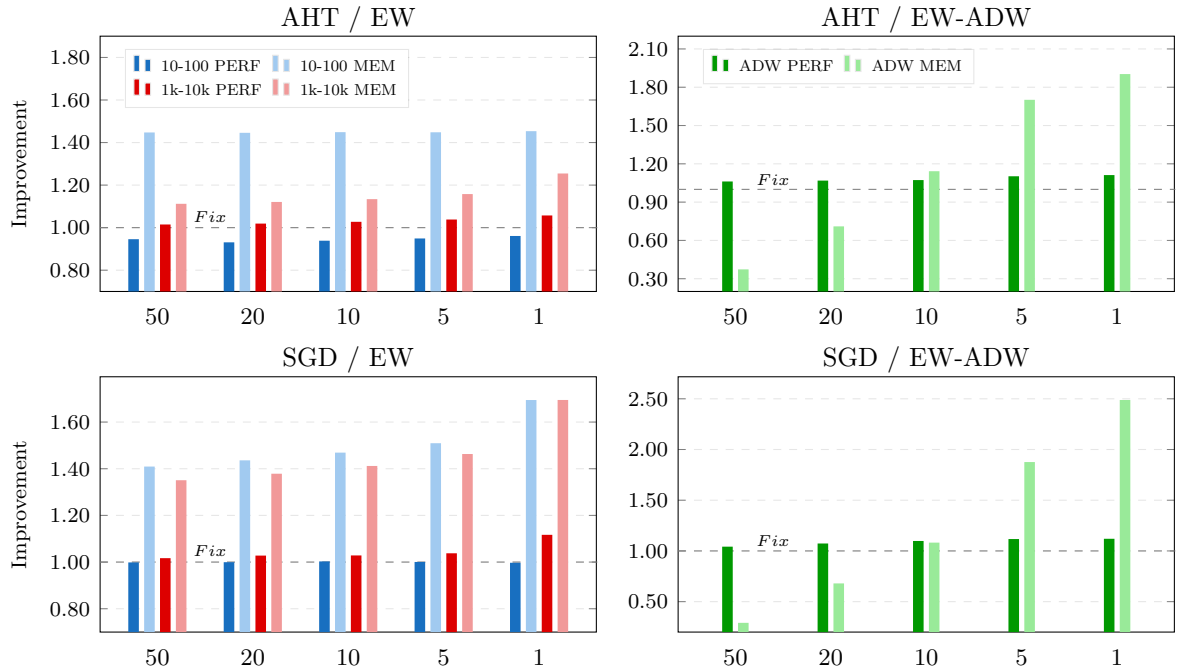

The trade-off between the quality of classification and memory consumption is presented in Fig. \ref{fig:win_dyn_trade}. The results for EW (UW exhibits analogous characteristics) show that while combining the dynamic adjustments with smaller windows we can use less memory (about 1.4-1.7 times less) at a cost of impaired performance (about 0.94), doing the same with larger windows results in improving both (about 1.1-1.7 for memory and 1.01-1.05 for kappa). It is intuitive, since the shrinking heuristic (Eq. \ref{eq:omega}), by making the small windows (containing 10-100 instances) even smaller, will more likely increase a chance of overfitting than efficiently improve reactivity. On the other hand, making too large windows more flexible provides enhanced reactivity to new concepts without too significant loss in generalization, especially for smaller budgets $B\leq10\%$ for which concepts evolve at a higher rate. 

When it comes to using a window size based on ADWIN, beside of the good predictive performance, we also observed that this approach tends to aggregate relatively large amounts of instances, especially for high budgets. Since there is no fixed counterpart for ADWIN, in Fig. \ref{fig:win_dyn_trade} we present a ratio between ADWIN and a standard window storing a comparable number of instances, which was $\omega_{max}=10000$. Based on that, we can see that the ADWIN-driven control provides the presented performance at a quite relevant memory cost, which decreases with the budget. It is also important that even if ADWIN stores plenty of instances, it maintains a good quality of classification regardless of the number of labeled instances, doing it better than $\omega_{max}=10000$ and $\omega_{max}=1000$ for AHT and competitively for SGD.

\subsubsection{Elevating significance}

In the final section of the first part of the experiments, we analyze the significance level $\alpha_e$, which is the only parameter that has to be set for the elevating strategy. Simultaneously, we also use this section to investigate the potential usefulness of the proposed ensemble techniques. All results for the elevating ensemble (EWE, SEE) using different $\alpha_e$ along with results for the switching committee (EWS, SES), baseline and single-classifier exploitation methods (EW, SE) are presented in Fig. \ref{tab:sign_perf}. Since EW performed at least as good as UW in all previous evaluations, we decided to omit it in the rest of the study.

\begin{table}[b]
	\caption{Average kappa for AHT and SGD ensembles compared with Base and single-classifier models.}
	\vspace*{-0.1cm}
	\centering
	\scalebox{0.8}{
		\begin{subtable}{\textwidth}\leftskip-45pt
			\begin{tabular}[H]{l >{\centering\arraybackslash} m{1cm} >{\centering\arraybackslash} m{1cm} >{\centering\arraybackslash} m{1cm} >{\centering\arraybackslash} m{1cm} >{\centering\arraybackslash} m{1cm} >{\centering\arraybackslash} m{1cm}}
				\textbf{AHT}& $100\%$ & $50\%$ & $20\%$ & $10\%$ & $5\%$ & $1\%$\\
				\midrule
				Base & 0.8673 & 0.8556 & 0.8311 & 0.8103 & 0.7850 & 0.6916\\
				\hdashline
				EW & 0.8779 & 0.8636 & 0.8425 & 0.8402 & 0.8226 & 0.7560\\
				EWS & 0.8798 & 0.8691 & 0.8457 & 0.8478 & 0.8318 & 0.7612\\
				EWE-1 & 0.8795 & 0.8704 & 0.8451 & 0.8523 & 0.8301 & 0.7604\\
				EWE-5 & 0.8787 & 0.8677 & 0.8440 & 0.8462 & 0.8272 & 0.7611\\
				EWE-10 & 0.8793 & 0.8706 & 0.8462 & 0.8463 & 0.8323 & 0.7599\\
				EWE-20 & 0.8796 & 0.8697 & 0.8449 & 0.8527 & 0.8286 & 0.7556\\
				\hdashline
				SE & 0.8768 & 0.8695 & 0.8459 & 0.6914 & 0.6856 & 0.6800\\
				SES & 0.8800 & 0.8698 & 0.8470 & 0.8136 & 0.7919 & 0.7394\\
				SEE-1 & 0.8817 & 0.8737 & 0.8457 & 0.8189 & 0.7983 & 0.7472\\
				SEE-5 & 0.8819 & 0.8697 & 0.8424 & 0.8174 & 0.8022 & 0.7485\\
				SEE-10 & 0.8782 & 0.8687 & 0.8441 & 0.8238 & 0.7966 & 0.7573\\
				SEE-20 & 0.8808 & 0.8692 & 0.8450 & 0.8205 & 0.8018 & 0.7480\\
				\bottomrule
			\end{tabular}%
		\hspace*{0.15cm}
			\begin{tabular}[H]{l >{\centering\arraybackslash} m{1cm} >{\centering\arraybackslash} m{1cm} >{\centering\arraybackslash} m{1cm} >{\centering\arraybackslash} m{1cm} >{\centering\arraybackslash} m{1cm} >{\centering\arraybackslash} m{1cm}}
				\textbf{SGD}& $100\%$ & $50\%$ & $20\%$ & $10\%$ & $5\%$ & $1\%$\\
				\midrule
				Base & 0.4772 & 0.4089 & 0.2892 & 0.2229 & 0.2018 & 0.1750\\
				\hdashline
				EW & 0.7026 & 0.6752 & 0.6424 & 0.6020 & 0.5621 & 0.3707\\
				EWS & 0.7027 & 0.6723 & 0.6415 & 0.6005 & 0.5508 & 0.3779\\
				EWE-1 & 0.7030 & 0.6720 & 0.6365 & 0.5980 & 0.5536 & 0.3788\\
				EWE-5 & 0.7028 & 0.6722 & 0.6401 & 0.5980 & 0.5530 & 0.3767\\
				EWE-10 & 0.7025 & 0.6730 & 0.6389 & 0.5985 & 0.5503 & 0.3761\\
				EWE-20 & 0.7023 & 0.6700 & 0.6371 & 0.5926 & 0.5532 & 0.3793\\
				\hdashline
				SE & 0.6913 & 0.6632 & 0.6311 & 0.5987 & 0.5543 & 0.3813\\
				SES & 0.6953 & 0.6643 & 0.6311 & 0.5994 & 0.5551 & 0.3810\\
				SEE-1 & 0.7001 & 0.6687 & 0.6325 & 0.6020 & 0.5576 & 0.3799\\
				SEE-5 & 0.6995 & 0.6697 & 0.6385 & 0.6008 & 0.5564 & 0.3824\\
				SEE-10 & 0.6994 & 0.6692 & 0.6309 & 0.6006 & 0.5566 & 0.3846\\
				SEE-20 & 0.6986 & 0.6695 & 0.6320 & 0.5977 & 0.5564 & 0.3810\\
				\bottomrule
			\end{tabular}
	\end{subtable}}
	\label{tab:sign_perf}
\end{table} 

\begin{figure}[h]
	\centering
	\leftskip0pt\begin{subfigure}{0.33\textwidth}
		\begin{tikzpicture}
		
		\begin{axis}
		[
		title={STABLE},
		title style={yshift=-1.5ex},
		width=1.05\textwidth,
		height=.75\textwidth,
		ymin=0.75,
		ymajorgrids=true,
		ylabel={SE / Base},
		ylabel near ticks,
		y tick label style={
			font=\scriptsize,
			/pgf/number format/.cd,
			fixed,
			fixed zerofill,
			precision=2,
			/tikz/.cd
		},
		ylabel style={font=\scriptsize, at={(-0.15,0.5)}},
		extra y ticks=1.0,
		extra y tick labels={},
		extra y tick style={
			ymajorgrids=true,
			ytick style={
				/pgfplots/major tick length=0pt,
			},
			grid style={
				gray,
				dashed,
				/pgfplots/on layer=axis foreground,
			},
		},
		grid style={dashed,white!90!black},
		xtick style={draw=none},
		xtick=data,
		xticklabels={100,50,20,10,5,1},
		xlabel style={font=\footnotesize},
		x tick label style={font=\scriptsize},
		legend cell align={left},
		legend style={font=\scriptsize,column sep=0.5ex, legend columns=1, at={(0.05,0.4)}, draw opacity=0.1, anchor=north west, nodes={scale=0.75}},
		legend entries={SE,SES,SEE}
		]
		
		\node[font=\tiny, shift={(0.45cm,0.0cm)}] at (axis cs: 0, 0.975) {$Base$};
		
		\addplot [	
		draw=green3,smooth,mark=triangle*,mark options={green3,scale=0.8}
		] coordinates {(0,1.01216038027493) (1,1.01705078039078) (2,1.01663886932884) (3,0.871392848007827) (4,0.88639972272552) (5,0.999999952862728)};
		
		\addplot [	
		draw=red2,smooth,mark=*,mark options={red2,scale=0.7}
		] coordinates {(0,1.01498185602007) (1,1.01840987800054) (2,1.01869976720405) (3,1.0008907586278) (4,1.005966947519) (5,1.07597699077145)};
		
		\addplot [	
		draw=blue2,smooth,mark=*,mark options={blue2,scale=0.7}
		] coordinates {(0,1.01775417252337) (1,1.01915018967809) (2,1.01378704651466) (3,1.0104450698465) (4,1.02201144767867) (5,1.09778248061905)};
		
		\end{axis}
		
		\end{tikzpicture}%
		\hspace*{0.05cm}
		\begin{tikzpicture}
		
		\begin{axis}
		[
		title={DRIFT},
		title style={yshift=-1.5ex},
		width=1.05\textwidth,
		height=.75\textwidth,
		ymajorgrids=true,
		ylabel near ticks,
		y tick label style={
			font=\scriptsize,
			/pgf/number format/.cd,
			fixed,
			fixed zerofill,
			precision=2,
			/tikz/.cd
		},
		ylabel style={font=\footnotesize, at={(-0.25,0.5)}},
		extra y ticks=1.0,
		extra y tick labels={},
		extra y tick style={
			ymajorgrids=true,
			ytick style={
				/pgfplots/major tick length=0pt,
			},
			grid style={
				gray,
				dashed,
				/pgfplots/on layer=axis foreground,
			},
		},
		grid style={dashed,white!90!black},
		xtick style={draw=none},
		xtick=data,
		xticklabels={100,50,20,10,5,1},
		xlabel style={font=\footnotesize},
		x tick label style={font=\scriptsize}
		]
		
		\node[font=\tiny, shift={(0.0cm,0.0cm)}] at (axis cs: 4.5, 1.018) {$Base$};
		
		\addplot [	
		draw=green3,smooth,mark=triangle*,mark options={green3,scale=0.8}
		] coordinates {(0,1.02093313415123) (1,1.02139346487853) (2,1.04589579097367) (3,1.001827145207) (4,1.14123023374736) (5,1.24922976337359)};
		
		\addplot [	
		draw=red2,smooth,mark=*,mark options={red2,scale=0.7}
		] coordinates {(0,1.02409000047476) (1,1.02933897094219) (2,1.04785007033546) (3,1.10281080574929) (4,1.20618707177811) (5,1.28668337858036)};
		
		\addplot [	
		draw=blue2,smooth,mark=*,mark options={blue2,scale=0.7}
		] coordinates {(0,1.02413730423941) (1,1.03193747774953) (2,1.04417985709001) (3,1.11741060461493) (4,1.23524770765055) (5,1.28070676330874)};
		
		\end{axis}
		
		\end{tikzpicture}%
		\hspace*{0.05cm}
		\begin{tikzpicture}
		
		\begin{axis}
		[
		title={AVG},
		title style={yshift=-1.5ex},
		width=1.05\textwidth,
		height=.75\textwidth,
		ymin=0.85,
		ymax=1.25,
		ytick={0.9,1.0,1.1,1.2},
		ymajorgrids=true,
		ylabel near ticks,
		y tick label style={
			font=\scriptsize,
			/pgf/number format/.cd,
			fixed,
			fixed zerofill,
			precision=2,
			/tikz/.cd
		},
		ylabel style={font=\footnotesize, at={(-0.25,0.5)}},
		extra y ticks=1.0,
		extra y tick labels={},
		extra y tick style={
			ymajorgrids=true,
			ytick style={
				/pgfplots/major tick length=0pt,
			},
			grid style={
				gray,
				dashed,
				/pgfplots/on layer=axis foreground,
			},
		},
		grid style={dashed,white!90!black},
		xtick style={draw=none},
		xtick=data,
		xticklabels={100,50,20,10,5,1},
		xlabel style={font=\footnotesize},
		x tick label style={font=\scriptsize},
		]
		
		\node[font=\tiny, shift={(0.45cm,0.0cm)}] at (axis cs: 0, 0.975) {$Base$};
		
		\addplot [	
		draw=green3,smooth,mark=triangle*,mark options={green3,scale=0.8}
		] coordinates {(0,1.01615962552483) (1,1.02086140907367) (2,1.03153171586302) (3,0.926459167344158) (4,0.985796982132004) (5,1.08955873036162)};
		
		\addplot [	
		draw=red2,smooth,mark=*,mark options={red2,scale=0.7}
		] coordinates {(0,1.01913399615297) (1,1.02335444435192) (2,1.03161862147413) (3,1.04391902636167) (4,1.08406329300596) (5,1.15169267849232)};
		
		\addplot [	
		draw=blue2,smooth,mark=*,mark options={blue2,scale=0.7}
		] coordinates {(0,1.02066405801142) (1,1.02493544511154) (2,1.02725655621105) (3,1.05560342483457) (4,1.10518476849612) (5,1.16351488678254)};
		
		\end{axis}
		
		\end{tikzpicture}%
	\end{subfigure}
	\caption{Improvement over Base given a budget for AHT using SE and the SE-based ensembles.}
	\label{fig:sign_aht}
\end{figure}
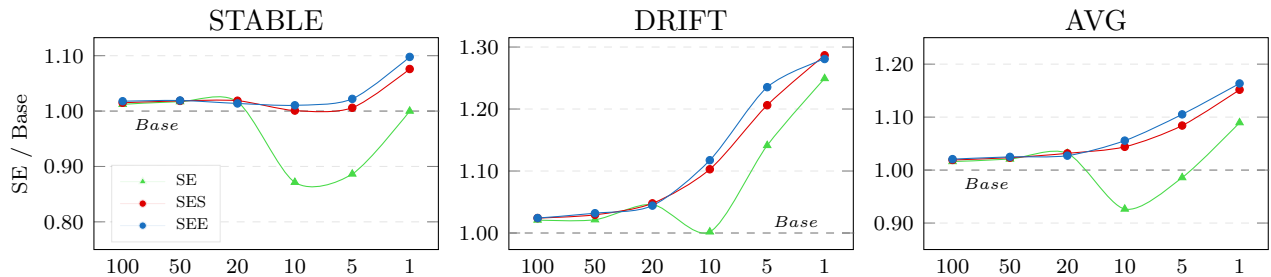

The presented results clearly indicate that there is no significant difference when changing the $\alpha_e$ value. Furthermore, it seems that for synthetic streams the ensemble methods were not able to provide significant improvements over tuned EW or SE for both classifiers. The only exception is for SE for which we purposely set nonoptimal intensity $\lambda_{max}=10$, slightly higher than previously obtained results suggested. This allowed us to check whether the ensembles are able to surpass the ineffective exploitation strategy by using the alternative standard base learner when it is needed. In Fig. \ref{fig:sign_aht} we can see that, indeed, improperly configured SE failed for budgets $B\leq10\%$, performing worse than the baseline for stable concepts by leading to overfitting. However, when combined with switching or elevating, the whole method was able to achieve performance at least as good as the baseline, or even to improve upon it for $B\leq5\%$. In addition, the ensembles boosted adaptation during concept drifts (by about 0.1), which is a very important observation, since dealing with changes under strictly limited supervision is an extremely challenging task.

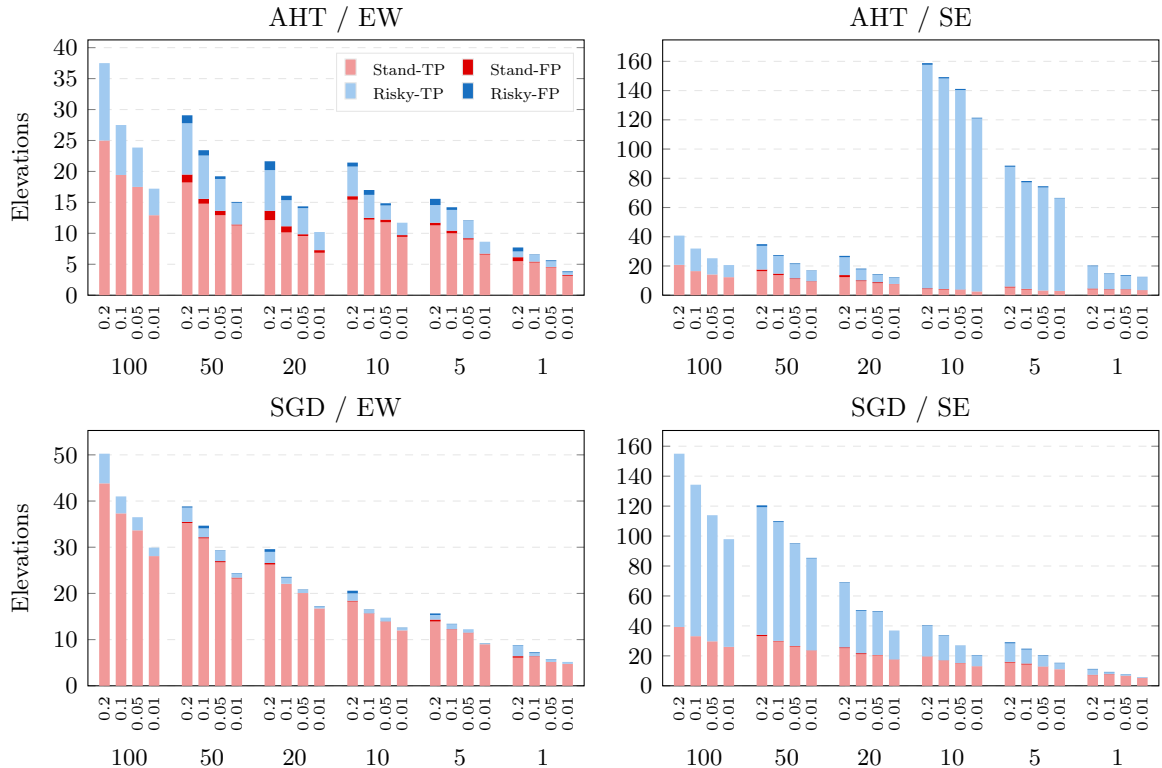
\begin{figure}[t]
	\centering
	\leftskip15pt\begin{subfigure}{0.4\textwidth}
	\begin{tikzpicture}
	\begin{axis}[
	title={AHT / EW},
	title style={yshift=-0.2ex},
	footnotesize,
	width=1.15\textwidth,
	height=0.7\textwidth,
	ymin=0,
	ymajorgrids=true,
	ylabel={Elevations},
	ylabel near ticks,
	ytick style={font=\footnotesize},
	ylabel style={font=\footnotesize, at={(-0.1,0.5)}},
	grid style={dashed,white!90!black},
	xtick style={draw=none},
	xtick=data,
	table/header=false,
	table/row sep=\\,
	xticklabels from table={
		0.2\\0.1\\0.05\\0.01\\ {} \\
		0.2\\0.1\\0.05\\0.01\\ {} \\
		0.2\\0.1\\0.05\\0.01\\ {} \\
		0.2\\0.1\\0.05\\0.01\\ {} \\
		0.2\\0.1\\0.05\\0.01\\ {} \\
		0.2\\0.1\\0.05\\0.01\\ {} \\
	}{[index]0},
	x tick label style={rotate=90, font=\tiny},
	xlabel style={yshift=-4ex,},
	enlarge x limits={abs=1},
	legend image post style={scale=0.7},
	legend cell align={left},
	legend style={font=\scriptsize,column sep=1ex, legend columns=2, at={(0.5,0.95)}, anchor=north west, nodes={scale=0.75}, draw opacity=0.1},
	legend entries={Stand-TP,Stand-FP,Risky-TP,Risky-FP},
	clip mode=individual,
	]
	
	
	\addplot [
	bar width=4pt,
	ybar stacked,
	fill=red4,
	draw=none,
	] table[x expr=\coordindex,y index=0]{
		25\\19.4285714285714\\17.5\\12.9285714285714\\0\\
		18.2142857142857\\14.7857142857143\\12.9285714285714\\11.2857142857143\\0\\
		12.1428571428571\\10.1428571428571\\9.57142857142857\\6.85714285714286\\0\\
		15.4285714285714\\12.2142857142857\\11.7857142857143\\9.42857142857143\\0\\
		11.2857142857143\\10\\9\\6.57142857142857\\0\\
		5.5\\5.28571428571429\\4.5\\3.14285714285714\\};
	
	\addplot [
	bar width=4pt,
	ybar stacked,
	fill=red2,
	draw=none,
	] table[x expr=\coordindex,y index=0]{
		0\\0\\0\\0\\0\\
		1.28571428571429\\0.785714285714286\\0.714285714285714\\0.142857142857143\\0\\
		1.5\\1\\0.285714285714286\\0.428571428571429\\0\\
		0.571428571428571\\0.285714285714286\\0.428571428571429\\0.285714285714286\\0\\
		0.428571428571429\\0.428571428571429\\0.214285714285714\\0.142857142857143\\0\\
		0.642857142857143\\0.142857142857143\\0.071428571428572\\0.142857142857143\\};
	
	\addplot [
	bar width=4pt,	
	ybar stacked,
	fill=blue4,
	draw=none,
	] table[x expr=\coordindex,y index=0]{
		12.5\\8.07142857142857\\6.35714285714286\\4.28571428571429\\0\\
		8.28571428571429\\7\\5.14285714285714\\3.5\\0\\
		6.57142857142857\\4.21428571428571\\4.21428571428571\\2.92857142857143\\0\\
		4.78571428571429\\3.71428571428571\\2.28571428571429\\2\\0\\
		2.85714285714286\\3.35714285714286\\2.78571428571429\\1.92857142857143\\0\\
		0.928571428571429\\1.07142857142857\\0.928571428571429\\0.428571428571429\\};
	
	\addplot [
	bar width=4pt,	
	ybar stacked,
	fill=blue2,
	draw=none,
	] table[x expr=\coordindex,y index=0]{
		0\\0\\0\\0\\0\\
		1.28571428571429\\0.857142857142857\\0.428571428571429\\0.142857142857143\\0\\
		1.42857142857143\\0.714285714285714\\0.285714285714286\\0\\0\\
		0.642857142857143\\0.785714285714286\\0.357142857142857\\0\\0\\
		1\\0.428571428571429\\0.071428571428572\\0\\0\\
		0.642857142857143\\0.071428571428572\\0.142857142857143\\0.142857142857143\\};
	
	\begin{scope}[
	every label/.append style={
		label distance=3.5ex,
		font=\footnotesize
	},
	]
	\node [label=below:{100}]
	at (axis cs:1.5,\pgfkeysvalueof{/pgfplots/ymin}) {};
	\node [label=below:{50}]
	at (axis cs:6.5,\pgfkeysvalueof{/pgfplots/ymin}) {};
	\node [label=below:{20}]
	at (axis cs:11.5,\pgfkeysvalueof{/pgfplots/ymin}) {};
	\node [label=below:{10}]
	at (axis cs:16.5,\pgfkeysvalueof{/pgfplots/ymin}) {};
	\node [label=below:{5}]
	at (axis cs:21.5,\pgfkeysvalueof{/pgfplots/ymin}) {};
	\node [label=below:{1}]
	at (axis cs:26.5,\pgfkeysvalueof{/pgfplots/ymin}) {};
	
	\end{scope}	
	\end{axis}
	\end{tikzpicture}%
	\hspace*{0.2cm}
	\begin{tikzpicture}
	\begin{axis}[
	title={AHT / SE},
	title style={yshift=-0.2ex},
	footnotesize,
	width=1.15\textwidth,
	height=0.7\textwidth,
	ymin=0,
	ymajorgrids=true,
	ylabel near ticks,
	ylabel style={font=\scriptsize},
	ytick style={font=\footnotesize},
	grid style={dashed,white!90!black},
	xtick style={draw=none},
	xtick=data,
	table/header=false,
	table/row sep=\\,
	xticklabels from table={
		0.2\\0.1\\0.05\\0.01\\ {} \\
		0.2\\0.1\\0.05\\0.01\\ {} \\
		0.2\\0.1\\0.05\\0.01\\ {} \\
		0.2\\0.1\\0.05\\0.01\\ {} \\
		0.2\\0.1\\0.05\\0.01\\ {} \\
		0.2\\0.1\\0.05\\0.01\\ {} \\
	}{[index]0},
	x tick label style={rotate=90, font=\tiny},
	xlabel style={yshift=-4ex,},
	enlarge x limits={abs=1},
	clip mode=individual,
	]
	
	\addplot [
	bar width=4pt,
	ybar stacked,
	fill=red4,
	draw=none,
	] table[x expr=\coordindex,y index=0]{
		20.7857142857143\\16.5\\14.2142857142857\\12.3571428571429\\0\\
		16.4285714285714\\13.7857142857143\\11.0714285714286\\9.28571428571429\\0\\
		12.3571428571429\\9.64285714285714\\8.28571428571429\\7.64285714285714\\0\\
		4.21428571428571\\3.92857142857143\\3.64285714285714\\2.5\\0\\
		5.28571428571429\\3.92857142857143\\3.14285714285714\\2.92857142857143\\0\\
		4.07142857142857\\3.71428571428571\\3.78571428571429\\3.42857142857143\\};
	
	\addplot [
	bar width=4pt,
	ybar stacked,
	fill=red2,
	draw=none,
	] table[x expr=\coordindex,y index=0]{
		0\\0\\0\\0\\0\\
		1.14285714285714\\1\\0.5\\0.285714285714286\\0\\
		1.42857142857143\\0.571428571428571\\0.642857142857143\\0.071428571428572\\0\\
		0.5\\0.357142857142857\\0.214285714285714\\0.071428571428572\\0\\
		0.642857142857143\\0.428571428571429\\0.071428571428572\\0\\0\\
		0.5\\0.357142857142857\\0.285714285714286\\0.071428571428572\\};
	
	\addplot [
	bar width=4pt,	
	ybar stacked,
	fill=blue4,
	draw=none,
	] table[x expr=\coordindex,y index=0]{
		20\\15.4285714285714\\11.0714285714286\\8.28571428571429\\0\\
		16.2142857142857\\12.0714285714286\\9.71428571428571\\7\\0\\
		12.2142857142857\\7.42857142857143\\4.92857142857143\\4.14285714285714\\0\\
		152.928571428571\\144\\136.5\\118.357142857143\\0\\
		81.9285714285714\\72.8571428571429\\70.5\\63.3571428571429\\0\\
		15.2857142857143\\10.4285714285714\\9.21428571428571\\8.92857142857143\\};
	
	\addplot [
	bar width=4pt,	
	ybar stacked,
	fill=blue2,
	draw=none,
	] table[x expr=\coordindex,y index=0]{
		0\\0\\0\\0\\0\\
		1.14285714285714\\0.571428571428571\\0.5\\0.285714285714286\\0\\
		0.928571428571429\\0.571428571428571\\0.428571428571429\\0.357142857142857\\0\\
		1.14285714285714\\0.928571428571429\\0.785714285714286\\0.5\\0\\
		0.785714285714286\\0.928571428571429\\0.857142857142857\\0.285714285714286\\0\\
		0.428571428571429\\0.357142857142857\\0.5\\0.142857142857143\\};
	
	\begin{scope}[
	every label/.append style={
		label distance=3.5ex,
		font=\footnotesize
	},
	]
	\node [label=below:{100}]
	at (axis cs:1.5,\pgfkeysvalueof{/pgfplots/ymin}) {};
	\node [label=below:{50}]
	at (axis cs:6.5,\pgfkeysvalueof{/pgfplots/ymin}) {};
	\node [label=below:{20}]
	at (axis cs:11.5,\pgfkeysvalueof{/pgfplots/ymin}) {};
	\node [label=below:{10}]
	at (axis cs:16.5,\pgfkeysvalueof{/pgfplots/ymin}) {};
	\node [label=below:{5}]
	at (axis cs:21.5,\pgfkeysvalueof{/pgfplots/ymin}) {};
	\node [label=below:{1}]
	at (axis cs:26.5,\pgfkeysvalueof{/pgfplots/ymin}) {};
	
	\end{scope}	
	\end{axis}
	\end{tikzpicture}
	\begin{tikzpicture}
	\begin{axis}[
	title={SGD / EW},
	title style={yshift=-0.2ex},
	footnotesize,
	width=1.15\textwidth,
	height=0.7\textwidth,
	ymin=0,
	ymajorgrids=true,
	ylabel={Elevations},
	ylabel near ticks,
	ylabel style={font=\footnotesize, at={(-0.1,0.5)}},
	ytick style={font=\footnotesize},
	grid style={dashed,white!90!black},
	xtick style={draw=none},
	xtick=data,
	table/header=false,
	table/row sep=\\,
	xticklabels from table={
		0.2\\0.1\\0.05\\0.01\\ {} \\
		0.2\\0.1\\0.05\\0.01\\ {} \\
		0.2\\0.1\\0.05\\0.01\\ {} \\
		0.2\\0.1\\0.05\\0.01\\ {} \\
		0.2\\0.1\\0.05\\0.01\\ {} \\
		0.2\\0.1\\0.05\\0.01\\ {} \\
	}{[index]0},
	x tick label style={rotate=90, font=\tiny},
	xlabel style={yshift=-4ex,},
	enlarge x limits={abs=1},
	clip mode=individual,
	]
	
	
	\addplot [
	bar width=4pt,
	ybar stacked,
	fill=red4,
	draw=none,
	] table[x expr=\coordindex,y index=0]{
		43.8333333333333\\37.3333333333333\\33.6666666666667\\28.0833333333333\\0\\
		35.25\\31.9166666666667\\26.75\\23.25\\0\\
		26.25\\22.0833333333333\\20.0833333333333\\16.75\\0\\
		18.25\\15.5833333333333\\13.8333333333333\\12\\0\\
		13.9166666666667\\12.1666666666667\\11.4166666666667\\9\\0\\
		6.08333333333333\\6.33333333333333\\5.16666666666667\\4.75\\};
	
	\addplot [
	bar width=4pt,
	ybar stacked,
	fill=red2,
	draw=none,
	] table[x expr=\coordindex,y index=0]{
		0\\0\\0\\0\\0\\
		0.25\\0.25\\0.25\\0.166666666666667\\0\\
		0.333333333333333\\0\\0\\0\\0\\
		0.166666666666667\\0.083333333333333\\0.083333333333333\\0\\0\\
		0.416666666666667\\0.083333333333333\\0.083333333333333\\0\\0\\
		0.333333333333333\\0.083333333333333\\0\\0\\};
	
	\addplot [
	bar width=4pt,	
	ybar stacked,
	fill=blue4,
	draw=none,
	] table[x expr=\coordindex,y index=0]{
		6.41666666666667\\3.66666666666667\\2.83333333333333\\1.83333333333333\\0\\
		3.08333333333333\\1.91666666666667\\2.25\\0.916666666666667\\0\\
		2.41666666666667\\1.33333333333333\\0.666666666666667\\0.416666666666667\\0\\
		1.58333333333333\\0.75\\0.833333333333333\\0.5\\0\\
		1\\1.08333333333333\\0.75\\0.083333333333333\\0\\
		2.33333333333333\\0.666666666666667\\0.5\\0.416666666666667\\};
	
	\addplot [
	bar width=4pt,	
	ybar stacked,
	fill=blue2,
	draw=none,
	] table[x expr=\coordindex,y index=0]{
		0\\0\\0\\0\\0\\
		0.25\\0.583333333333333\\0.083333333333333\\0.083333333333333\\0\\
		0.583333333333333\\0.166666666666667\\0.083333333333333\\0.083333333333333\\0\\
		0.583333333333333\\0.083333333333333\\0\\0.083333333333333\\0\\
		0.333333333333333\\0.083333333333333\\0\\0.083333333333333\\0\\
		0.083333333333333\\0.166666666666667\\0.083333333333333\\0\\};
	
	\begin{scope}[
	every label/.append style={
		label distance=3.5ex,
		font=\footnotesize
	},
	]
	\node [label=below:{100}]
	at (axis cs:1.5,\pgfkeysvalueof{/pgfplots/ymin}) {};
	\node [label=below:{50}]
	at (axis cs:6.5,\pgfkeysvalueof{/pgfplots/ymin}) {};
	\node [label=below:{20}]
	at (axis cs:11.5,\pgfkeysvalueof{/pgfplots/ymin}) {};
	\node [label=below:{10}]
	at (axis cs:16.5,\pgfkeysvalueof{/pgfplots/ymin}) {};
	\node [label=below:{5}]
	at (axis cs:21.5,\pgfkeysvalueof{/pgfplots/ymin}) {};
	\node [label=below:{1}]
	at (axis cs:26.5,\pgfkeysvalueof{/pgfplots/ymin}) {};
	
	\end{scope}	
	\end{axis}
	\end{tikzpicture}%
	\hspace*{0.2cm}
	\begin{tikzpicture}
	\begin{axis}[
	title={SGD / SE},
	title style={yshift=-0.2ex},
	footnotesize,
	width=1.15\textwidth,
	height=0.7\textwidth,
	ymin=0,
	ymajorgrids=true,
	ylabel near ticks,
	ylabel style={font=\scriptsize},
	ytick style={font=\footnotesize},
	grid style={dashed,white!90!black},
	xtick style={draw=none},
	xtick=data,
	table/header=false,
	table/row sep=\\,
	xticklabels from table={
		0.2\\0.1\\0.05\\0.01\\ {} \\
		0.2\\0.1\\0.05\\0.01\\ {} \\
		0.2\\0.1\\0.05\\0.01\\ {} \\
		0.2\\0.1\\0.05\\0.01\\ {} \\
		0.2\\0.1\\0.05\\0.01\\ {} \\
		0.2\\0.1\\0.05\\0.01\\ {} \\
	}{[index]0},
	x tick label style={rotate=90, font=\tiny},
	xlabel style={yshift=-4ex,},
	enlarge x limits={abs=1},
	clip mode=individual
	]
	
	\addplot [
	bar width=4pt,
	ybar stacked,
	fill=red4,
	draw=none,
	] table[x expr=\coordindex,y index=0]{
		39.2857142857143\\33.1428571428571\\29.7142857142857\\26\\0\\
		33.2142857142857\\29.6428571428571\\26.2142857142857\\23.4285714285714\\0\\
		25.1428571428571\\21.1428571428571\\20.3571428571429\\17.3571428571429\\0\\
		19.3571428571429\\16.7857142857143\\15.0714285714286\\13\\0\\
		15.5\\14.2142857142857\\12.7142857142857\\11.0714285714286\\0\\
		7.14285714285714\\7.57142857142857\\6.57142857142857\\5.21428571428571\\};
	
	\addplot [
	bar width=4pt,
	ybar stacked,
	fill=red2,
	draw=none,
	] table[x expr=\coordindex,y index=0]{
		0\\0\\0\\0\\0\\
		0.857142857142857\\0.357142857142857\\0.357142857142857\\0.214285714285714\\0\\
		0.571428571428571\\0.642857142857143\\0.214285714285714\\0.214285714285714\\0\\
		0.214285714285714\\0.214285714285714\\0.142857142857143\\0.071428571428572\\0\\
		0.5\\0.642857142857143\\0.142857142857143\\0\\0\\
		0.214285714285714\\0.214285714285714\\0.142857142857143\\0.071428571428572\\};
	
	\addplot [
	bar width=4pt,	
	ybar stacked,
	fill=blue4,
	draw=none,
	] table[x expr=\coordindex,y index=0]{
		115.714285714286\\101.142857142857\\84.2142857142857\\71.9285714285714\\0\\
		85.2142857142857\\79.3571428571429\\68.1428571428571\\61.4285714285714\\0\\
		43\\28.2857142857143\\28.7142857142857\\19.3571428571429\\0\\
		20.4285714285714\\16.2857142857143\\11.7142857142857\\7.21428571428571\\0\\
		12.3571428571429\\9.35714285714286\\7.14285714285714\\4\\0\\
		3.57142857142857\\1.21428571428571\\1\\0.571428571428571\\};
	
	\addplot [
	bar width=4pt,	
	ybar stacked,
	fill=blue2,
	draw=none,
	] table[x expr=\coordindex,y index=0]{
		0\\0\\0\\0\\0\\
		1.21428571428571\\0.571428571428571\\0.357142857142857\\0.357142857142857\\0\\
		0.357142857142857\\0.5\\0.428571428571429\\0\\0\\
		0.357142857142857\\0.357142857142857\\0.071428571428572\\0.285714285714286\\0\\
		0.714285714285714\\0.571428571428571\\0.357142857142857\\0.285714285714286\\0\\
		0.357142857142857\\0.285714285714286\\0.142857142857143\\0\\};
	
	\begin{scope}[
	every label/.append style={
		label distance=3.5ex,
		font=\footnotesize
	},
	]
	\node [label=below:{100}]
	at (axis cs:1.5,\pgfkeysvalueof{/pgfplots/ymin}) {};
	\node [label=below:{50}]
	at (axis cs:6.5,\pgfkeysvalueof{/pgfplots/ymin}) {};
	\node [label=below:{20}]
	at (axis cs:11.5,\pgfkeysvalueof{/pgfplots/ymin}) {};
	\node [label=below:{10}]
	at (axis cs:16.5,\pgfkeysvalueof{/pgfplots/ymin}) {};
	\node [label=below:{5}]
	at (axis cs:21.5,\pgfkeysvalueof{/pgfplots/ymin}) {};
	\node [label=below:{1}]
	at (axis cs:26.5,\pgfkeysvalueof{/pgfplots/ymin}) {};
	
	\end{scope}	
	\end{axis}
	\end{tikzpicture}
	\end{subfigure}
	\caption{Number of elevations given a budget and significance level for ATH and SGD using elevating ensembles.}
	\label{fig:elev}
\end{figure}

The elevation of results obtained for SE is a good indicator that our ensembles should be able to increase the lower bound of our approach, by providing that it will never be worse than the baseline. After looking at the results in Fig. \ref{fig:elev} we can understand that this is not a coincidence. The bar plots present how many times on average either the standard base learner (Stand-TP, Stand-FP) or the risky one (Risky-TP, Risky-FP) was correctly (true positive, TP) or incorrectly (false positive, FP) elevated with respect to the available budget and used significance level $\alpha_e$. The correctness depended on the precision of the error estimated based on the partial information from labeled instances. We can easily notice that a prevalent number of elevations was done correctly. The only configuration that stands out with a visible number of false positives is the one using $\alpha_e=0.2$, which is, in fact, an unusual value for the Welch's test. Nevertheless, it still did not affect performance in a meaningful way. The results prove that we can effectively track an error even under strictly limited supervision and utilize it in switching or elevating techniques.

A more careful analysis of the average number of elevations gives us some additional observations. Firstly, the total number of elevations decreases with the budget and significance level $\alpha_e$, which is intuitive. Secondly, in most of the cases, if an exploitation strategy is properly configured, we will be replacing mainly the standard learner with the risky one. This can be mostly seen for EW, however, we would observe the same relation also for SE, if it was properly configured for AHT, or for SGD with the exclusion of the SEA stream. This was the only benchmark for which the SE strategy was constantly failing when combined with the latter classifier, resulting in more than 500 elevations and biasing the average value. We could use median instead, however, we decided to keep the average to show that the SE strategy tends to be more sensitive to improper configuration and overfitting than EW or UW, taking into consideration results for both AHT and SGD. Finally, one may wonder why even if we replace EW or SE with a standard learner multiple times, we do not obtain significant improvements over them (except for the specified cases for AHT with SE)? The reason may be that even if we elevate the exploiting learner it keeps falling into the same pitfalls again and again, as the internal characteristics of the streams do not allow any more improvements than the strategies give themselves in other parts of the streams. For a similar reason, we cannot see additional enhancements when using switching. Fortunately, the differences are more visible for real data streams, in favor of the ensembles, which makes them more than just safe lower bound providers.

\subsubsection{Final comparison}

\smallskip
\noindent\textbf{Default configurations.} Based on the observations made in the previous sections, we determined default settings for our strategies and used them in the final comparison, which was conducted using real streams. Since we found trade-offs obtained for the dynamic heuristics fair, we decided to use dynamic intensity controlled by the ADWIN error ($\epsilon=\epsilon_{ADW}$), as well as dynamic window size determined by the same algorithm ($\omega_{max}=\omega_{ADW}$). We used the same significance levels $\alpha_{\theta}$ as for synthetic streams. For our elevating ensembles, due to the lack of impact, we chose one of the widely used values, $\alpha_e=0.05$. Finally, in order to find a compromise between evidence suggesting lower values of intensity for AHT and making sure that we fully utilize the potential of our strategies, we distinguished two groups of $\lambda_{max}$ values: less risky $\lambda^l_{max}$ and risky $\lambda^r_{max}$. In addition to that, since AHT and SGD exhibited substantially different preferences regarding required intensity, we set: $\lambda^l_{max}=\{1,1,1,1,1,10\}$, $\lambda^r_{max}=\{100, 100, 100, 1000, 1000, 1000\}$ for AHT, where each value corresponds to a budget ranging from $B=100\%$ to $B=1\%$, respectively, and $\lambda^l_{max}=10$, $\lambda^r_{max}=1000$ for SGD.

\medskip
\noindent\textbf{Results.} The average performance of all considered classifiers under varying budget is presented in Tab. \ref{tab:final}. It provides a general overview of the final comparison. The main observation is that the proposed instance exploitation strategies are capable of providing the best predictive quality (in bold) out of all models and that the risky ones exhibited much better quality than the more conservative configurations. Most importantly, our best strategies were able to outperform baseline models using only active learning (ALR, ALRV, ALS) for both base learners (by about 0.1 kappa for AHT and 0.15 for SGD), regardless of the available budget. It shows that the depicted problem of underfitting while learning temporary drifting concepts under limited supervision is critical in real scenarios and active learning fails to solve it efficiently on its own.

\begin{table}[t]
	\caption{Final average kappa for AHT and SGD on real data streams.}
	\vspace*{-0.1cm}
	\centering
	\scalebox{0.8}{
		\begin{subtable}{\textwidth}\leftskip-45pt
			\begin{tabular}[H]{l >{\centering\arraybackslash} m{1cm} >{\centering\arraybackslash} m{1cm} >{\centering\arraybackslash} m{1cm} >{\centering\arraybackslash} m{1cm} >{\centering\arraybackslash} m{1cm} >{\centering\arraybackslash} m{1cm}}
				\textbf{AHT}& $100\%$ & $50\%$ & $20\%$ & $10\%$ & $5\%$ & $1\%$\\
				\midrule
				EW$^{l}$ & 0.5637 & 0.5275 & 0.4611 & 0.4116 & 0.3721 & 0.3040\\
				EWS$^{l}$ & 0.5831 & 0.5387 & 0.4765 & 0.4108 & 0.3880 & 0.3306\\
				EWE$^{l}$ & 0.5638 & 0.5246 & 0.4707 & 0.4177 & 0.3780 & 0.3326\\
				SE$^{l}$ & 0.5752 & 0.5378 & 0.4778 & 0.4362 & 0.3953 & 0.3401\\
				SES$^{l}$ & 0.5914 & 0.5516 & 0.4990 & 0.4463 & 0.4035 & 0.3503\\
				SEE$^{l}$ & 0.5838 & 0.5345 & 0.4859 & 0.4202 & 0.3842 & \textbf{0.3578}\\
				\hdashline
				EW$^{r}$ & 0.6432 & 0.5959 & 0.5520 & 0.5035 & 0.4521 & 0.3315\\
				EWS$^{r}$ & 0.6492 & 0.6128 & \textbf{0.5634} & \textbf{0.5161} & 0.4525 & 0.3287\\
				EWE$^{r}$ & \textbf{0.6569} & \textbf{0.6182} & 0.5620 & 0.5099 & \textbf{0.4690} & 0.3405\\
				SE$^{r}$ & 0.5655 & 0.5404 & 0.4915 & 0.3900 & 0.3530 & 0.2541\\
				SES$^{r}$ & 0.6424 & 0.6052 & 0.5594 & 0.4984 & 0.4684 & 0.3500\\
				SEE$^{r}$ & 0.6416 & 0.6044 & 0.5557 & 0.5025 & 0.4573 & 0.3473\\
				\hdashline
				ALR & 0.5403 & 0.4978 & 0.4475 & 0.3946 & 0.3527 & 0.2997\\
				ALRV & 0.5063 & 0.5036 & 0.4466 & 0.3953 & 0.3537 & 0.3030\\
				ALS & 0.5242 & 0.5082 & 0.4424 & 0.3858 & 0.3496 & 0.2986\\
				AUC & 0.4297 & 0.4218 & 0.3680 & 0.2990 & 0.2656 & 0.2324\\
				DWM & 0.6154 & 0.5897 & 0.5309 & 0.4993 & 0.4501 & 0.3335\\
				LNSE & 0.2734 & 0.2986 & 0.3096 & 0.3000 & 0.2237 & 0.1717\\
				ADOB & 0.2588 & 0.2564 & 0.2608 & 0.2705 & 0.2562 & 0.2464\\
				ABAG & 0.6204 & 0.6003 & 0.5414 & 0.4918 & 0.4455 & 0.3203\\
				OBAG & 0.5752 & 0.5438 & 0.4711 & 0.4129 & 0.3805 & 0.2964\\
				\bottomrule
			\end{tabular}
			\hspace*{0.25cm}
			\begin{tabular}[H]{l >{\centering\arraybackslash} m{1cm} >{\centering\arraybackslash} m{1cm} >{\centering\arraybackslash} m{1cm} >{\centering\arraybackslash} m{1cm} >{\centering\arraybackslash} m{1cm} >{\centering\arraybackslash} m{1cm}}
				\textbf{SGD}& $100\%$ & $50\%$ & $20\%$ & $10\%$ & $5\%$ & $1\%$\\
				\midrule
				EW$^{l}$ & 0.3397 & 0.3171 & 0.2685 & 0.2043 & 0.1568 & 0.0764\\
				EWS$^{l}$ & 0.3751 & 0.3519 & 0.3073 & 0.2706 & 0.2357 & 0.1345\\
				EWE$^{l}$ & 0.3684 & 0.3429 & 0.3021 & 0.2681 & 0.2304 & 0.1164\\
				SE$^{l}$ & 0.3954 & 0.3760 & 0.3363 & 0.3044 & 0.2819 & 0.1743\\
				SES$^{l}$ & 0.3956 & 0.3707 & 0.3355 & 0.3045 & 0.2804 & 0.1679\\
				SEE$^{l}$ & 0.3960 & 0.3710 & 0.3410 & 0.3073 & 0.2812 & 0.1647\\
				\hdashline
				EW$^{r}$ & 0.4124 & 0.3962 & 0.3533 & 0.3027 & 0.2711 & 0.1753\\
				EWS$^{r}$ & 0.4351 & 0.4188 & 0.3879 & 0.3600 & 0.3184 & 0.2344\\
				EWE$^{r}$ & 0.4379 & 0.4208 & 0.3878 & 0.3576 & 0.3176 & 0.2114\\
				SE$^{r}$ & \textbf{0.4714} & \textbf{0.4455} & \textbf{0.4109} & \textbf{0.3813} & \textbf{0.3634} & \textbf{0.2891}\\
				SES$^{r}$ & 0.4698 & 0.4421 & 0.4042 & 0.3730 & 0.3469 & 0.2664\\
				SEE$^{r}$ & 0.4683 & 0.4414 & 0.4044 & 0.3742 & 0.3411 & 0.2679\\
				\hdashline
				ALR & 0.3440 & 0.3098 & 0.2644 & 0.2195 & 0.1865 & 0.1054\\
				ALRV & 0.3151 & 0.3156 & 0.2653 & 0.2215 & 0.1863 & 0.1228\\
				ALS & 0.3204 & 0.3135 & 0.2669 & 0.2199 & 0.1800 & 0.1158\\
				AWE & 0.1289 & 0.1062 & 0.0918 & 0.0559 & 0.0427 & 0.0298\\
				DWM & 0.3689 & 0.3375 & 0.2861 & 0.2593 & 0.2369 & 0.1404\\
				LNSE & 0.1242 & 0.1325 & 0.1163 & 0.0843 & 0.0456 & 0.0210\\
				ADOB & 0.4553 & 0.4347 & 0.4050 & 0.3632 & 0.3271 & 0.2299\\
				ABAG & 0.3673 & 0.3444 & 0.2873 & 0.2427 & 0.2169 & 0.1381\\
				OBAG & 0.3343 & 0.3056 & 0.2553 & 0.2147 & 0.1845 & 0.1239\\
				\bottomrule
			\end{tabular}
		\end{subtable}}
	\label{tab:final}
\end{table}

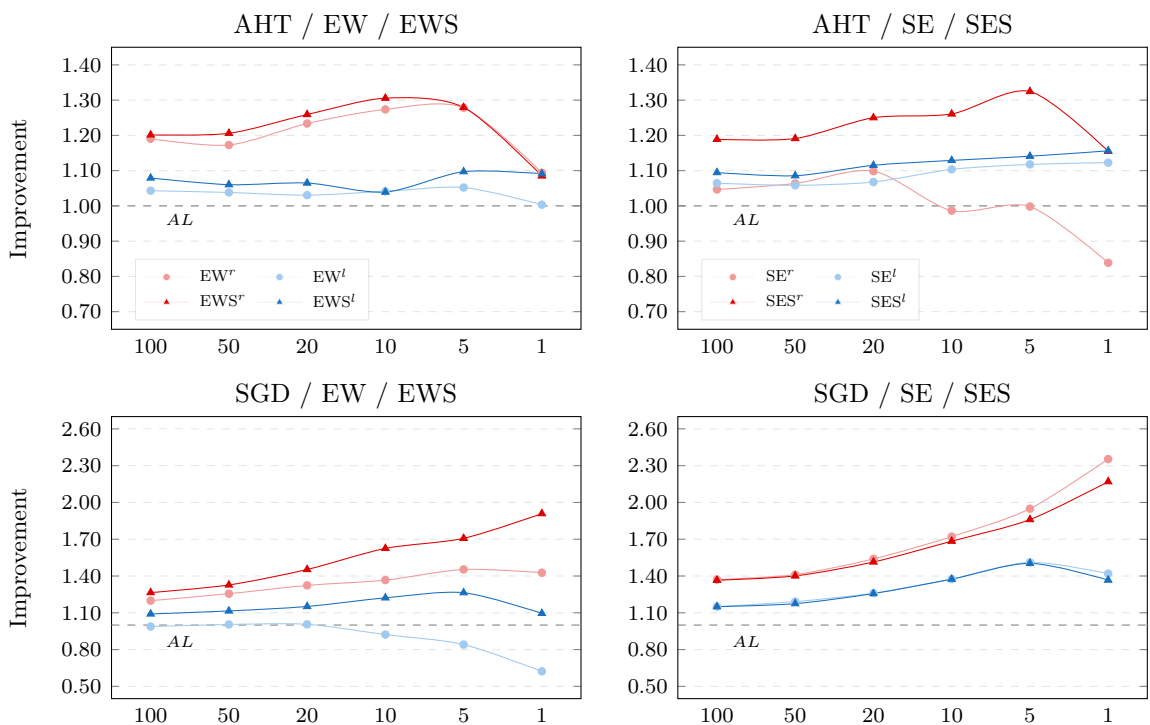
\begin{figure}[h]
	\centering
	\leftskip25pt\begin{subfigure}{0.4\textwidth}
		\begin{tikzpicture}
		
		\begin{axis}
		[
		title={AHT / EW / EWS},
		title style={yshift=-0.2ex},
		footnotesize,
		width=1.1\textwidth,
		height=.75\textwidth,
		ymin=0.65,
		ymax=1.45,
		ymajorgrids=true,
		ylabel={Improvement},
		ylabel near ticks,
		ylabel style={font=\footnotesize, at={(-0.15,0.5)}},
		y tick label style={
			font=\scriptsize,
			/pgf/number format/.cd,
			fixed,
			fixed zerofill,
			precision=2,
			/tikz/.cd
		},
		extra y ticks=1.0,
		extra y tick labels={},
		extra y tick style={
			ymajorgrids=true,
			ytick style={
				/pgfplots/major tick length=0pt,
			},
			grid style={
				gray,
				dashed,
				/pgfplots/on layer=axis foreground,
			},
		},
		grid style={dashed,white!90!black},
		xtick style={draw=none},
		xtick=data,
		xticklabels={100,50,20,10,5,1},
		xlabel style={font=\footnotesize},
		x tick label style={font=\scriptsize},
		legend cell align={left},
		legend style={font=\scriptsize,column sep=0.5ex, legend columns=2, at={(0.05,0.25)}, anchor=north west, draw opacity=0.1, nodes={scale=0.75}},
		legend entries={EW$^{r}$,EW$^{l}$,EWS$^{r}$,EWS$^{l}$}
		]
		
		\node[font=\tiny, shift={(-0.65cm,0.0cm)}] at (axis cs: 1, 0.96) {$AL$};
		
		\addplot [	
		draw=red4,smooth,mark=*,mark options={red4,scale=0.7}
		] coordinates {(0,1.19039628880225) (1,1.17266441887619) (2,1.23367103694875) (3,1.27367480140995) (4,1.27823333144828) (5,1.09404568058795)};
		
		\addplot [	
		draw=blue4,smooth,mark=*,mark options={blue4,scale=0.7}
		] coordinates {(0,1.04315747914919) (1,1.03811185600147) (2,1.03048271752086) (3,1.04135395409239) (4,1.05214047390149) (5,1.00347665361088)};
		
		\addplot [	
		draw=red2,smooth,mark=triangle*,mark options={red2,scale=0.8}
		] coordinates {(0,1.20140156936288) (1,1.20586961940622) (2,1.25916269368296) (3,1.30561786382878) (4,1.27942091274105) (5,1.08500198037231)};
		
		\addplot [	
		draw=blue2,smooth,mark=triangle*,mark options={blue2,scale=0.8}
		] coordinates {(0,1.07913438286532) (1,1.06017868622332) (2,1.06483909415971) (3,1.0392626448316) (4,1.09713660954966) (5,1.09116313867007)};
		
		\end{axis}
		
		\end{tikzpicture}%
		\hspace*{0.4cm}
		\begin{tikzpicture}
		
		\begin{axis}
		[
		title={AHT / SE / SES},
		title style={yshift=-0.2ex},
		footnotesize,
		width=1.1\textwidth,
		height=.75\textwidth,
		ymin=0.65,
		ymax=1.45,
		ymajorgrids=true,
		ylabel near ticks,
		y tick label style={
			font=\scriptsize,
			/pgf/number format/.cd,
			fixed,
			fixed zerofill,
			precision=2,
			/tikz/.cd
		},
		ylabel style={font=\scriptsize},
		extra y ticks=1.0,
		extra y tick labels={},
		extra y tick style={
			ymajorgrids=true,
			ytick style={
				/pgfplots/major tick length=0pt,
			},
			grid style={
				gray,
				dashed,
				/pgfplots/on layer=axis foreground,
			},
		},
		grid style={dashed,white!90!black},
		xtick style={draw=none},
		xtick=data,
		xticklabels={100,50,20,10,5,1},
		xlabel style={font=\footnotesize},
		x tick label style={font=\scriptsize},
		legend cell align={left},
		legend style={font=\scriptsize,column sep=0.5ex, legend columns=2, at={(0.05,0.25)}, anchor=north west, draw opacity=0.1, nodes={scale=0.75}},
		legend entries={SE$^{r}$,SE$^{l}$,SES$^{r}$,SES$^{l}$}
		]
		
		\node[font=\tiny, shift={(-0.65cm,0.0cm)}] at (axis cs: 1, 0.96) {$AL$};
		
		\addplot [	
		draw=red4,smooth,mark=*,mark options={red4,scale=0.7}
		] coordinates {(0,1.04647633617924) (1,1.06355037193498) (2,1.09851013110846) (3,0.986507682188454) (4,0.998058398838809) (5,0.838599656735466)};
		
		\addplot [	
		draw=blue4,smooth,mark=*,mark options={blue4,scale=0.7}
		] coordinates {(0,1.06446478803731) (1,1.05827637327316) (2,1.06775923718713) (3,1.10358726999814) (4,1.11764594997078) (5,1.12267306253576)};
		
		\addplot [	
		draw=red2,smooth,mark=triangle*,mark options={red2,scale=0.8}
		] coordinates {(0,1.1888787445097) (1,1.19105781719429) (2,1.25014898688915) (3,1.26082336869445) (4,1.32447359987936) (5,1.15517317255644)};
		
		\addplot [	
		draw=blue2,smooth,mark=triangle*,mark options={blue2,scale=0.8}
		] coordinates {(0,1.09455658096037) (1,1.08555160515855) (2,1.11510727056019) (3,1.12905401986744) (4,1.14083206092481) (5,1.15622937112177)};
		
		\end{axis}
		
		\end{tikzpicture}
	\end{subfigure}\\
	\vspace*{0.1cm}
	\begin{subfigure}{0.4\textwidth}
		\begin{tikzpicture}
		
		\begin{axis}
		[
		title={SGD / EW / EWS},
		title style={yshift=-0.2ex},
		footnotesize,
		width=1.1\textwidth,
		height=.75\textwidth,
		ymin=0.4,
		ymax=2.7,
		ytick={0.5, 0.8, 1.1, 1.4, 1.7, 2.0, 2.3, 2.6},
		ymajorgrids=true,
		ylabel={Improvement},
		ylabel near ticks,
		ylabel style={font=\footnotesize, at={(-0.15,0.5)}},
		y tick label style={
			font=\scriptsize,
			/pgf/number format/.cd,
			fixed,
			fixed zerofill,
			precision=2,
			/tikz/.cd
		},
		extra y ticks=1.0,
		extra y tick labels={},
		extra y tick style={
			ymajorgrids=true,
			ytick style={
				/pgfplots/major tick length=0pt,
			},
			grid style={
				gray,
				dashed,
				/pgfplots/on layer=axis foreground,
			},
		},
		grid style={dashed,white!90!black},
		xtick style={draw=none},
		xtick=data,
		xticklabels={100,50,20,10,5,1},
		xlabel style={font=\footnotesize},
		x tick label style={font=\scriptsize}
		]
		
		\node[font=\tiny, shift={(-0.65cm,0.0cm)}] at (axis cs: 1, 0.865) {$AL$};
		
		\addplot [	
		draw=red4,smooth,mark=*,mark options={red4,scale=0.7}
		] coordinates {(0,1.19883324288705) (1,1.2555879494655) (2,1.32347578489897) (3,1.36661349065406) (4,1.45347299181353) (5,1.42723769201542)};
		
		\addplot [	
		draw=blue4,smooth,mark=*,mark options={blue4,scale=0.7}
		] coordinates {(0,0.987692844406543) (1,1.00473232771369) (2,1.00591952443989) (3,0.922433254070975) (4,0.841061023129446) (5,0.622157086250882)};
		
		\addplot [	
		draw=red2,smooth,mark=triangle*,mark options={red2,scale=0.8}
		] coordinates {(0,1.26484611210171) (1,1.32705877382009) (2,1.45323076154557) (3,1.62516931041748) (4,1.70746791548994) (5,1.90880964012376)};
		
		\addplot [	
		draw=blue2,smooth,mark=triangle*,mark options={blue2,scale=0.8}
		] coordinates {(0,1.09054965501202) (1,1.11505471753919) (2,1.15148487648925) (3,1.22195466995756) (4,1.2638973295678) (5,1.09531563806112)};
		
		\end{axis}
		
		\end{tikzpicture}%
		\hspace*{0.4cm}
		\begin{tikzpicture}
		
		\begin{axis}
		[
		title={SGD / SE / SES},
		title style={yshift=-0.2ex},
		footnotesize,
		width=1.1\textwidth,
		height=.75\textwidth,
		ymin=0.4,
		ymax=2.7,
		ytick={0.5, 0.8, 1.1, 1.4, 1.7, 2.0, 2.3, 2.6},
		ymajorgrids=true,
		ylabel near ticks,
		y tick label style={
			font=\scriptsize,
			/pgf/number format/.cd,
			fixed,
			fixed zerofill,
			precision=2,
			/tikz/.cd
		},
		ylabel style={font=\scriptsize},
		extra y ticks=1.0,
		extra y tick labels={},
		extra y tick style={
			ymajorgrids=true,
			ytick style={
				/pgfplots/major tick length=0pt,
			},
			grid style={
				gray,
				dashed,
				/pgfplots/on layer=axis foreground,
			},
		},
		grid style={dashed,white!90!black},
		xtick style={draw=none},
		xtick=data,
		xticklabels={100,50,20,10,5,1},
		xlabel style={font=\footnotesize},
		x tick label style={font=\scriptsize},
		]
		
		\node[font=\tiny, shift={(-0.65cm,0.0cm)}] at (axis cs: 1, 0.865) {$AL$};
		
		\addplot [	
		draw=red4,smooth,mark=*,mark options={red4,scale=0.7}
		] coordinates {(0,1.37045507403675) (1,1.41164913170237) (2,1.53960086919599) (3,1.72151822532583) (4,1.94891502520287) (5,2.35363404440102)};
		
		\addplot [	
		draw=blue4,smooth,mark=*,mark options={blue4,scale=0.7}
		] coordinates {(0,1.149643383208) (1,1.19138462838552) (2,1.25980967604965) (3,1.37434909550613) (4,1.51145747685268) (5,1.41942137545459)};
		
		\addplot [	
		draw=red2,smooth,mark=triangle*,mark options={red2,scale=0.8}
		] coordinates {(0,1.36578416931545) (1,1.40089576203152) (2,1.51442415765417) (3,1.68413448514583) (4,1.86032960354628) (5,2.16929924550833)};
		
		\addplot [	
		draw=blue2,smooth,mark=triangle*,mark options={blue2,scale=0.8}
		] coordinates {(0,1.15005039150322) (1,1.17465246968353) (2,1.25698728675975) (3,1.37480058995274) (4,1.50337825760555) (5,1.36693263854964)};
		
		\end{axis}
		
		\end{tikzpicture}
	\end{subfigure}
	\caption{Improvement over the bast AL given a budget for AHT and SGD using different strategies on real data streams.}
	\label{fig:final_impr}
\end{figure}

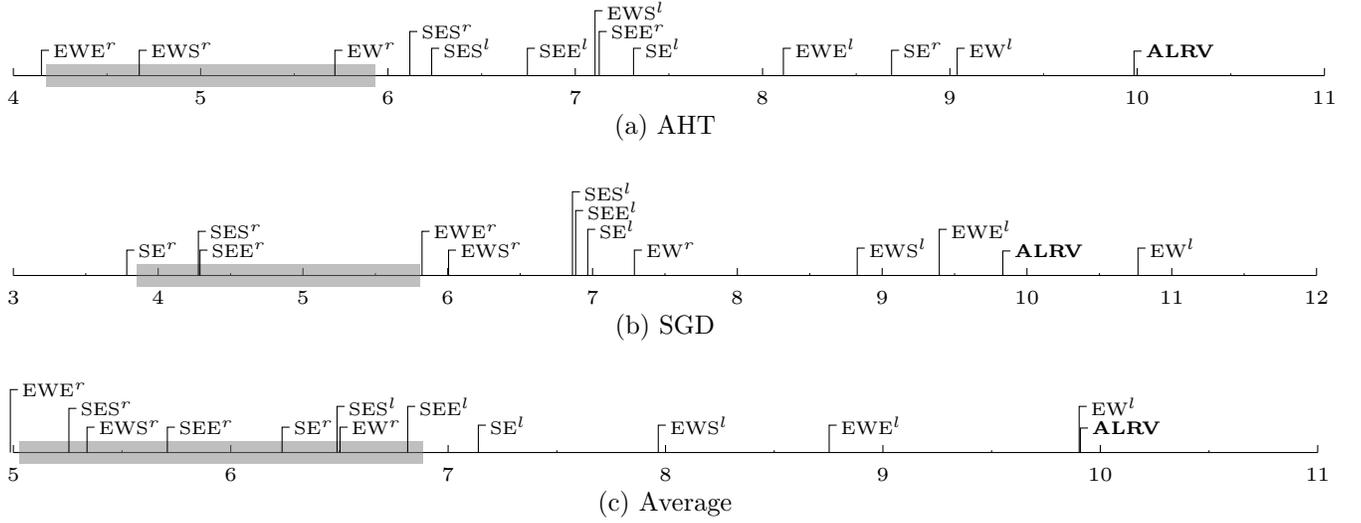
\begin{figure}[h]
	\centering
	\small
	{\begin{subfigure}{\textwidth}
			\begin{tikzpicture}
			\pgfmathsetmacro{\sc}{2.49}
			\pgfmathsetmacro{\b}{4}
			
			\draw (0,0) -- ({\sc*(11-\b)},0);
			\foreach \x in {4,5,6,7,8,9,10,11} {
				\draw ({\sc*(\x-\b)}, 0) -- ++(0,.1) node [below=0.15cm,scale=1.25] {\tiny \x};
				\ifthenelse{\x < 11}{\draw ({\sc*(\x-\b) +.5*\sc}, 0) -- ++(0,.03);}{}
			}
			\coordinate (c0) at ({\sc*(4.15-\b)},0);
			\coordinate (c1) at ({\sc*(4.672222-\b)},0);
			\coordinate (c2) at ({\sc*(5.716667-\b)},0);
			\coordinate (c3) at ({\sc*(6.116667-\b)},0);
			\coordinate (c4) at ({\sc*(6.233333-\b)},0);
			\coordinate (c5) at ({\sc*(6.744444-\b)},0);
			\coordinate (c6) at ({\sc*(7.105556-\b)},0);
			\coordinate (c7) at ({\sc*(7.127778-\b)},0);
			\coordinate (c8) at ({\sc*(7.311111-\b)},0);
			\coordinate (c9) at ({\sc*(8.111111-\b)},0);
			\coordinate (c10) at ({\sc*(8.688889-\b)},0);
			\coordinate (c11) at ({\sc*(9.038889-\b)},0);
			\coordinate (c12) at ({\sc*(9.983333-\b)},0);
			
			\node (l0) at (c0) [above right=.1cm and .1cm, align=left, scale=1.15] {\hspace*{-.075cm}\tiny EWE$^{r}$};
			\node (l1) at (c1) [above right=.1cm and .1cm, align=center, scale=1.15] {\hspace*{-.075cm}\tiny EWS$^{r}$};
			\node (l2) at (c2) [above right=.1cm and .1cm, align=left, scale=1.15] {\hspace*{-.075cm}\tiny EW$^{r}$};
			\node (l3) at (c3) [above right=.35cm and .1cm, align=center, scale=1.15] {\hspace*{-.075cm}\tiny SES$^{r}$};
			\node (l4) at (c4) [above right=.1cm and .1cm, align=left, scale=1.15] {\hspace*{-.075cm}\tiny SES$^{l}$};
			\node (l5) at (c5) [above right=0.1cm and .1cm, align=left, scale=1.15] {\hspace*{-.075cm}\tiny SEE$^{l}$};
			\node (l6) at (c6) [above right=.6cm and .1cm, align=center, scale=1.15] {\hspace*{-.075cm}\tiny EWS$^{l}$};
			\node (l7) at (c7) [above right=.35cm and .1cm, align=center, scale=1.15] {\hspace*{-.075cm}\tiny SEE$^{r}$};
			\node (l8) at (c8) [above right=.1cm and .1cm, align=left, scale=1.15] {\hspace*{-.075cm}\tiny SE$^{l}$};
			\node (l9) at (c9) [above right=.1cm and .1cm, align=left, scale=1.15] {\hspace*{-.075cm}\tiny EWE$^{l}$};
			\node (l10) at (c10) [above right=.1cm and .1cm, align=left, scale=1.15] {\hspace*{-.075cm}\tiny SE$^{r}$};
			\node (l11) at (c11) [above right=.1cm and .1cm, align=left, scale=1.15] {\hspace*{-.075cm}\tiny EW$^{l}$};
			\node (l12) at (c12) [above right=.1cm and .1cm, align=left, scale=1.15] {\hspace*{-.075cm}\tiny \textbf{ALRV}};
			
			\fill[fill=gray,fill opacity=0.5] ({\sc*(4.175-\b)},-0.15) rectangle ({\sc*(4.175 + 1.7539-\b)},0.15);
			
			\foreach \x in {0,...,12} {
				\draw (l\x) -| (c\x);
			};
			\end{tikzpicture}
		\end{subfigure}\vspace*{-0.25cm}\subcaption{AHT}}
	\vspace*{0.35cm}
	{\begin{subfigure}{\textwidth}
			\begin{tikzpicture}
			\pgfmathsetmacro{\sc}{1.925}
			\pgfmathsetmacro{\b}{3}
			
			\draw (0,0) -- ({\sc*(12-\b)},0);
			\foreach \x in {3,4,5,6,7,8,9,10,11,12} {
				\draw ({\sc*(\x-\b)}, 0) -- ++(0,.1) node [below=0.15cm,scale=1.25] {\tiny \x};
				\ifthenelse{\x < 12}{\draw ({\sc*(\x-\b) +.5*\sc}, 0) -- ++(0,.03);}{}
			}
			\coordinate (c0) at ({\sc*(3.783333-\b)},0);
			\coordinate (c1) at ({\sc*(4.277778-\b)},0);
			\coordinate (c2) at ({\sc*(4.288889-\b)},0);
			\coordinate (c3) at ({\sc*(5.822222-\b)},0);
			\coordinate (c4) at ({\sc*(6.005556-\b)},0);
			\coordinate (c5) at ({\sc*(6.861111-\b)},0);
			\coordinate (c6) at ({\sc*(6.883333-\b)},0);
			\coordinate (c7) at ({\sc*(6.966667-\b)},0);
			\coordinate (c8) at ({\sc*(7.288889-\b)},0);
			\coordinate (c9) at ({\sc*(8.827778-\b)},0);
			\coordinate (c10) at ({\sc*(9.394444-\b)},0);
			\coordinate (c11) at ({\sc*(9.833333-\b)},0);
			\coordinate (c12) at ({\sc*(10.766667-\b)},0);
			
			\node (l0) at (c0) [above right=.1cm and .1cm, align=left, scale=1.15] {\hspace*{-.075cm}\tiny SE$^{r}$};
			\node (l1) at (c1) [above right=.35cm and .1cm, align=center, scale=1.15] {\hspace*{-.075cm}\tiny SES$^{r}$};
			\node (l2) at (c2) [above right=.1cm and .1cm, align=left, scale=1.15] {\hspace*{-.075cm}\tiny SEE$^{r}$};
			\node (l3) at (c3) [above right=.35cm and .1cm, align=center, scale=1.15] {\hspace*{-.075cm}\tiny EWE$^{r}$};
			\node (l4) at (c4) [above right=.1cm and .1cm, align=left, scale=1.15] {\hspace*{-.075cm}\tiny EWS$^{r}$};
			\node (l5) at (c5) [above right=.85cm and .1cm, align=left, scale=1.15] {\hspace*{-.075cm}\tiny SES$^{l}$};
			\node (l6) at (c6) [above right=.6cm and .1cm, align=center, scale=1.15] {\hspace*{-.075cm}\tiny SEE$^{l}$};
			\node (l7) at (c7) [above right=.35cm and .1cm, align=center, scale=1.15] {\hspace*{-.075cm}\tiny SE$^{l}$};
			\node (l8) at (c8) [above right=.1cm and .1cm, align=left, scale=1.15] {\hspace*{-.075cm}\tiny EW$^{r}$};
			\node (l9) at (c9) [above right=.1cm and .1cm, align=left, scale=1.15] {\hspace*{-.075cm}\tiny EWS$^{l}$};
			\node (l10) at (c10) [above right=.35cm and .1cm, align=left, scale=1.15] {\hspace*{-.075cm}\tiny EWE$^{l}$};
			\node (l11) at (c11) [above right=.1cm and .1cm, align=left, scale=1.15] {\hspace*{-.075cm}\tiny \textbf{ALRV}};
			\node (l12) at (c12) [above right=.1cm and .1cm, align=left, scale=1.15] {\hspace*{-.075cm}\tiny EW$^{l}$};
			
			\fill[fill=gray,fill opacity=0.5] ({\sc*(3.85-\b)},-0.15) rectangle ({\sc*(3.85 + 1.9581-\b},0.15);
			
			\foreach \x in {0,...,12} {
				\draw (l\x) -| (c\x);
			};
			\end{tikzpicture}
		\end{subfigure}\vspace*{-0.25cm}\subcaption{SGD}}
	\vspace*{0.35cm}
	{\begin{subfigure}{\textwidth}
			\begin{tikzpicture}
			\pgfmathsetmacro{\sc}{2.89}
			\pgfmathsetmacro{\b}{5}
			
			\draw (0,0) -- ({\sc*(11-\b)},0);
			\foreach \x in {5,6,7,8,9,10,11} {
				\ifthenelse{\x > 4}{\draw ({\sc*(\x-\b)}, 0) -- ++(0,.1) node [below=0.15cm,scale=1.25] {\tiny \x};}{}
				\ifthenelse{\x < 11}{\draw ({\sc*(\x-\b) +.5*\sc}, 0) -- ++(0,.03);}{}
			}
			\coordinate (c0) at ({\sc*(4.986111-\b)},0);
			\coordinate (c1) at ({\sc*(5.2555555-\b)},0);
			\coordinate (c2) at ({\sc*(5.338889-\b)},0);
			\coordinate (c3) at ({\sc*(5.7083335-\b)},0);
			\coordinate (c4) at ({\sc*(6.236111-\b)},0);
			\coordinate (c5) at ({\sc*(6.488889-\b)},0);
			\coordinate (c6) at ({\sc*(6.502778-\b)},0);
			\coordinate (c7) at ({\sc*(6.8138885-\b)},0);
			\coordinate (c8) at ({\sc*(7.138889-\b)},0);
			\coordinate (c9) at ({\sc*(7.966667-\b)},0);
			\coordinate (c10) at ({\sc*(8.7527775-\b)},0);
			\coordinate (c11) at ({\sc*(9.902778-\b)},0);
			\coordinate (c12) at ({\sc*(9.908333-\b)},0);
			
			\node (l0) at (c0) [above right=.6cm and .1cm, align=left, scale=1.15] {\hspace*{-.075cm}\tiny EWE$^{r}$};
			\node (l1) at (c1) [above right=.35cm and .1cm, align=center, scale=1.15] {\hspace*{-.075cm}\tiny SES$^{r}$};
			\node (l2) at (c2) [above right=.1cm and .1cm, align=left, scale=1.15] {\hspace*{-.075cm}\tiny EWS$^{r}$};
			\node (l3) at (c3) [above right=.1cm and .1cm, align=center, scale=1.15] {\hspace*{-.075cm}\tiny SEE$^{r}$};
			\node (l4) at (c4) [above right=.1cm and .1cm, align=left, scale=1.15] {\hspace*{-.075cm}\tiny SE$^{r}$};
			\node (l5) at (c5) [above right=.35cm and .1cm, align=left, scale=1.15] {\hspace*{-.075cm}\tiny SES$^{l}$};
			\node (l6) at (c6) [above right=.1cm and .1cm, align=center, scale=1.15] {\hspace*{-.075cm}\tiny EW$^{r}$};
			\node (l7) at (c7) [above right=.35cm and .1cm, align=center, scale=1.15] {\hspace*{-.075cm}\tiny SEE$^{l}$};
			\node (l8) at (c8) [above right=.1cm and .1cm, align=left, scale=1.15] {\hspace*{-.075cm}\tiny SE$^{l}$};
			\node (l9) at (c9) [above right=.1cm and .1cm, align=left, scale=1.15] {\hspace*{-.075cm}\tiny EWS$^{l}$};
			\node (l10) at (c10) [above right=.1cm and .1cm, align=left, scale=1.15] {\hspace*{-.075cm}\tiny EWE$^{l}$};
			\node (l11) at (c11) [above right=.35cm and .1cm, align=left, scale=1.15] {\hspace*{-.075cm}\tiny EW$^{l}$};
			\node (l12) at (c12) [above right=.1cm and .1cm, align=left, scale=1.15] {\hspace*{-.075cm}\tiny \textbf{ALRV}};
			
			\fill[fill=gray,fill opacity=0.5] ({\sc*(5.03-\b)},-0.15) rectangle ({\sc*(5.03 + 1.856-\b)},0.15);
			
			\foreach \x in {0,...,12} {
				\draw (l\x) -| (c\x);
			};
			\end{tikzpicture}
		\end{subfigure}\vspace*{-0.25cm}\subcaption{Average}}
	\caption{Bonferroni-Dunn test for different strategies on real data streams.}
	\label{fig:bonferroni_in}
\end{figure}

In Fig. \ref{fig:final_impr} we can see gain obtained from using our exploitation strategies compared with the best baseline model using only active learning (AL) for each budget. It provides more insight into trends and relations present in the aggregated results. Firstly, as we already mentioned, the more risky approaches, applying more intense exploitation of labeled instances, outperforms the safer configurations in almost all cases. We can see that EW$^r$ and EWS$^r$ provided about 0.1-0.2 improvement in gain over EW$^l$ and EWS$^l$ for both AHT and SGD. In the case of SE and SES, the enhancement remained the same for SGD and for AHT using SES. These relations are analogous for EWE and SEE. Interestingly, for AHT with SE$^r$ we noticed that while for lower budgets the base learner did not work well with this extreme strategy, the switching technique SES$^r$ empowered the strategy to become more efficient even than SE$^l$, which was initially better than SE$^r$. This shows that taking even a very risky adaptation while having a good lower-bound backup may still be beneficial. On the other hand, results for AHT using SE$^r$ and SE$^l$ indicate that we may need to be a bit more careful with more reactive classifiers. Statistical rank tests presented in Fig. \ref{fig:bonferroni_in} prove that the advantage of the more risky exploitation is significant, as the best risky strategies for AHT (EWS$^r$, EWE$^r$) and SGD (SE$^r$, SES$^r$, SEE$^r$) are significantly better than any less risky method.

Secondly, Fig. \ref{fig:final_impr} shows that, generally and analogously to the results for the synthetic streams, the improvement over the baseline models increases as we limit available supervision. It is, again, intuitively correct, since if we have less labeled instances, the risk of encountering underfitting becomes higher. The only exception for this trend can be noticed when using the extremely limited budget $B=1\%$ and the reason for that is the same as the one given for the characteristic curve for SGD in Fig. \ref{fig:fix_gen_sgd}. The increase is more clear for SGD, which, in general, adapts less efficiently than AHT, giving more occasions for improvements. Regardless of the budget, almost all strategies turned out to be significantly better than a model without instance exploitation (Fig. \ref{fig:bonferroni_in}, ALRV, in bold), which proves the general usefulness of the proposed approaches. Interestingly, since our methods were able to provide some improvements also for fully labeled streams ($B=100\%$), we can assume that it may enhance adaptation to unstable data in general -- limiting budget only makes the problem harder and more severe.

Finally, based on Fig. \ref{fig:final_impr} and Fig. \ref{fig:bonferroni_in} we can claim that switching and elevating ensembles provide intended improvements. In almost all cases they maintained at least as good performance as the best base learner, either a single-classifier strategy (SE, EW) or a standard model without exploitation (ALRV). As a consequence, the ensembles are always at least as good as the baseline, providing noticeable improvements in almost all cases, as opposed to simpler strategies (SE$^r$ for AHT and EW$^l$ for SGD). Fig. \ref{fig:bonferroni_in} suggests that the improvements over the single-classifier methods are usually on the verge of significance. The only exception is, one more time, SE$^r$ for SGD -- since it worked outstandingly well with this classifier (especially for the lowest budgets), there was a low chance that the baseline could provide better prediction and, as a consequence, any incorrect switch led to some reduction in improvements. This once again shows that even very extreme and risky exploitation makes sense. In addition, compared with results obtained for synthetic streams, the ensembles were also able to improve upon SE and EW in some cases, even if the simpler strategies already provided gain over the baseline. This very likely shows that real streams are more unstable or complex than synthetic ones, thus requiring more intensive adaptation from classifiers. Last but not least, we did not obtain a significant difference between the switching and elevating techniques.

\begin{figure}[H]
	\centering
	\small
	{\begin{subfigure}{\textwidth}
			\begin{tikzpicture}
			\pgfmathsetmacro{\sc}{1.74}
			\pgfmathsetmacro{\b}{4}
			
			\draw (0,0) -- ({\sc*(14-\b)},0);
			\foreach \x in {4,5,6,7,8,9,10,11,12,13,14} {
				\draw ({\sc*(\x-\b)}, 0) -- ++(0,.1) node [below=0.15cm,scale=1.25] {\tiny \x};
				\ifthenelse{\x < 14}{\draw ({\sc*(\x-\b) +.5*\sc}, 0) -- ++(0,.03);}{}
			}
			\coordinate (c0) at ({\sc*(4.427778-\b)},0);
			\coordinate (c1) at ({\sc*(4.855556-\b)},0);
			\coordinate (c2) at ({\sc*(5.344444-\b)},0);
			\coordinate (c3) at ({\sc*(5.85-\b)},0);
			\coordinate (c4) at ({\sc*(6.238889-\b)},0);
			\coordinate (c5) at ({\sc*(6.311111-\b)},0);
			\coordinate (c6) at ({\sc*(7.111111-\b)},0);
			\coordinate (c7) at ({\sc*(7.188889-\b)},0);
			\coordinate (c8) at ({\sc*(8.872222-\b)},0);
			\coordinate (c9) at ({\sc*(9.155556-\b)},0);
			\coordinate (c10) at ({\sc*(9.355556-\b)},0);
			\coordinate (c11) at ({\sc*(9.888889-\b)},0);
			\coordinate (c12) at ({\sc*(10.255556-\b)},0);
			\coordinate (c13) at ({\sc*(11.833333-\b)},0);
			\coordinate (c14) at ({\sc*(13.311111-\b)},0);
			
			\node (l0) at (c0) [above right=1.35cm and .1cm, align=left, scale=1.15] {\hspace*{-.075cm}\tiny \textbf{EWE$^{\mathbf{r}}$}};
			\node (l1) at (c1) [above right=1.1cm and .1cm, align=center, scale=1.15] {\hspace*{-.075cm}\tiny \textbf{EWS$^{\mathbf{r}}$}};
			\node (l2) at (c2) [above right=.85cm and .1cm, align=left, scale=1.15] {\hspace*{-.075cm}\tiny ABAG};
			\node (l3) at (c3) [above right=.6cm and .1cm, align=center, scale=1.15] {\hspace*{-.075cm}\tiny \textbf{EW$^{\mathbf{r}}$}};
			\node (l4) at (c4) [above right=0.35cm and .1cm, align=left, scale=1.15] {\hspace*{-.075cm}\tiny \textbf{SES$^{\mathbf{r}}$}};
			\node (l5) at (c5) [above right=.1cm and .05cm, align=left, scale=1.15] {\hspace*{-.075cm}\tiny DWM};
			\node (l6) at (c6) [above right=.35cm and .1cm, align=center, scale=1.15] {\hspace*{-.075cm}\tiny \textbf{SEE$^{\mathbf{r}}$}};
			\node (l7) at (c7) [above right=.1cm and .1cm, align=center, scale=1.15] {\hspace*{-.075cm}\tiny OBAG$^{\,}$};
			\node (l8) at (c8) [above right=1.1cm and .1cm, align=left, scale=1.15] {\hspace*{-.075cm}\tiny \textbf{SE$^{\mathbf{r}}$}};
			\node (l9) at (c9) [above right=.85cm and .1cm, align=left, scale=1.15] {\hspace*{-.075cm}\tiny ALR};
			\node (l10) at (c10) [above right=.6cm and .1cm, align=left, scale=1.15] {\hspace*{-.075cm}\tiny ALRV};
			\node (l11) at (c11) [above right=.35cm and .1cm, align=left, scale=1.15] {\hspace*{-.075cm}\tiny ALS};
			\node (l12) at (c12) [above right=.1cm and .1cm, align=left, scale=1.15] {\hspace*{-.075cm}\tiny AUC};
			\node (l13) at (c13) [above right=.1cm and .1cm, align=left, scale=1.15] {\hspace*{-.075cm}\tiny ADOB};
			\node (l14) at (c14) [above right=.1cm and .1cm, align=left, scale=1.15] {\hspace*{-.075cm}\tiny LNSE};
			
			\fill[fill=gray,fill opacity=0.5] ({\sc*(4.5-\b)},-0.15) rectangle ({\sc*(4.5+2.0237-\b)},0.15);
			
			\foreach \x in {0,...,14} {
				\draw (l\x) -| (c\x);
			};
			\end{tikzpicture}
		\end{subfigure}\vspace*{-0.25cm}\subcaption{AHT}}
	\vspace*{0.35cm}
	{\begin{subfigure}{\textwidth}
			\begin{tikzpicture}
			\pgfmathsetmacro{\sc}{1.74}
			\pgfmathsetmacro{\b}{4}
			
			\draw (0,0) -- ({\sc*(14-\b)},0);
			\foreach \x in {4,5,6,7,8,9,10,11,12,13,14} {
				\draw ({\sc*(\x-\b)}, 0) -- ++(0,.1) node [below=0.15cm,scale=1.25] {\tiny \x};
				\ifthenelse{\x < 14}{\draw ({\sc*(\x-\b) +.5*\sc}, 0) -- ++(0,.03);}{}
			}
			\coordinate (c0) at ({\sc*(4.222222-\b)},0);
			\coordinate (c1) at ({\sc*(4.572222-\b)},0);
			\coordinate (c2) at ({\sc*(4.716667-\b)},0);
			\coordinate (c3) at ({\sc*(6.238889-\b)},0);
			\coordinate (c4) at ({\sc*(6.366667-\b)},0);
			\coordinate (c5) at ({\sc*(6.55-\b)},0);
			\coordinate (c6) at ({\sc*(7.588889-\b)},0);
			\coordinate (c7) at ({\sc*(7.988889-\b)},0);
			\coordinate (c8) at ({\sc*(8.25-\b)},0);
			\coordinate (c9) at ({\sc*(9.016667-\b)},0);
			\coordinate (c10) at ({\sc*(9.683333-\b)},0);
			\coordinate (c11) at ({\sc*(10.111111-\b)},0);
			\coordinate (c12) at ({\sc*(10.155556-\b)},0);
			\coordinate (c13) at ({\sc*(11.922222-\b)},0);
			\coordinate (c14) at ({\sc*(12.616667-\b)},0);
			
			\node (l0) at (c0) [above right=.6cm and .1cm, align=left, scale=1.15] {\hspace*{-.075cm}\tiny \textbf{SE$^{\mathbf{r}}$}};
			\node (l1) at (c1) [above right=.35cm and .1cm, align=center, scale=1.15] {\hspace*{-.075cm}\tiny \textbf{SES$^{\mathbf{r}}$}};
			\node (l2) at (c2) [above right=.1cm and .1cm, align=left, scale=1.15] {\hspace*{-.075cm}\tiny \textbf{SEE$^{\mathbf{r}}$}};
			\node (l3) at (c3) [above right=.6cm and .1cm, align=center, scale=1.15] {\hspace*{-.075cm}\tiny \textbf{EWE$^{\mathbf{r}}$}};
			\node (l4) at (c4) [above right=.35cm and .1cm, align=left, scale=1.15] {\hspace*{-.075cm}\tiny \textbf{EWS$^{\mathbf{r}}$}};
			\node (l5) at (c5) [above right=.1cm and .1cm, align=left, scale=1.15] {\hspace*{-.075cm}\tiny ADOB};
			\node (l6) at (c6) [above right=.6cm and .1cm, align=center, scale=1.15] {\hspace*{-.075cm}\tiny \textbf{EW$^{\mathbf{r}}$}};
			\node (l7) at (c7) [above right=.35cm and .1cm, align=center, scale=1.15] {\hspace*{-.075cm}\tiny ABAG};
			\node (l8) at (c8) [above right=.1cm and .05cm, align=left, scale=1.15] {\hspace*{-.075cm}\tiny DWM};
			\node (l9) at (c9) [above right=.1cm and .1cm, align=left, scale=1.15] {\hspace*{-.075cm}\tiny OBAG};
			\node (l10) at (c10) [above right=.6cm and .1cm, align=left, scale=1.15] {\hspace*{-.075cm}\tiny ALRV};
			\node (l11) at (c11) [above right=.35cm and .1cm, align=left, scale=1.15] {\hspace*{-.075cm}\tiny ALS};
			\node (l12) at (c12) [above right=.1cm and .1cm, align=left, scale=1.15] {\hspace*{-.075cm}\tiny ALR};
			\node (l13) at (c13) [above right=.1cm and .1cm, align=left, scale=1.15] {\hspace*{-.075cm}\tiny AWE};
			\node (l14) at (c14) [above right=.1cm and .1cm, align=left, scale=1.15] {\hspace*{-.075cm}\tiny LNSE};
			
			\fill[fill=gray,fill opacity=0.5] ({\sc*(4.29-\b)},-0.15) rectangle ({\sc*(4.29+2.1716-\b)},0.15);
			
			\foreach \x in {0,...,14} {
				\draw (l\x) -| (c\x);
			};
			\end{tikzpicture}
		\end{subfigure}\vspace*{-0.25cm}\subcaption{SGD}}
	\vspace*{0.35cm}
	{\begin{subfigure}{\textwidth}
			\begin{tikzpicture}
			\pgfmathsetmacro{\sc}{1.925}
			\pgfmathsetmacro{\b}{5}
			
			\draw (0,0) -- ({\sc*(14-\b)},0);
			\foreach \x in {5,6,7,8,9,10,11,12,13,14} {
				\draw ({\sc*(\x-\b)}, 0) -- ++(0,.1) node [below=0.15cm,scale=1.25] {\tiny \x};
				\ifthenelse{\x < 14}{\draw ({\sc*(\x-\b) +.5*\sc}, 0) -- ++(0,.03);}{}
			}
			\coordinate (c0) at ({\sc*(5.3333335-\b)},0);
			\coordinate (c1) at ({\sc*(5.4055555-\b)},0);
			\coordinate (c2) at ({\sc*(5.6111115-\b)},0);
			\coordinate (c3) at ({\sc*(5.913889-\b)},0);
			\coordinate (c4) at ({\sc*(6.547222-\b)},0);
			\coordinate (c5) at ({\sc*(6.6666665-\b)},0);
			\coordinate (c6) at ({\sc*(6.7194445-\b)},0);
			\coordinate (c7) at ({\sc*(7.2805555-\b)},0);
			\coordinate (c8) at ({\sc*(8.102778-\b)},0);
			\coordinate (c9) at ({\sc*(9.1916665-\b)},0);
			\coordinate (c10) at ({\sc*(9.5194445-\b)},0);
			\coordinate (c11) at ({\sc*(9.655556-\b)},0);
			\coordinate (c12) at ({\sc*(10-\b)},0);
			\coordinate (c13) at ({\sc*(11.088889-\b)},0);
			\coordinate (c14) at ({\sc*(12.963889-\b)},0);
			
			\node (l0) at (c0) [above right=.85cm and .1cm, align=left, scale=1.15] {\hspace*{-.075cm}\tiny \textbf{EWE$^{\mathbf{r}}$}};
			\node (l1) at (c1) [above right=.6cm and .1cm, align=center, scale=1.15] {\hspace*{-.075cm}\tiny \textbf{SES$^{\mathbf{r}}$}};
			\node (l2) at (c2) [above right=.35cm and .1cm, align=left, scale=1.15] {\hspace*{-.075cm}\tiny \textbf{EWS$^{\mathbf{r}}$}};
			\node (l3) at (c3) [above right=.1cm and .02cm, align=center, scale=1.15] {\hspace*{-.075cm}\tiny \textbf{SEE$^{\mathbf{r}}$}};
			\node (l4) at (c4) [above right=0.85cm and .1cm, align=left, scale=1.15] {\hspace*{-.075cm}\tiny \textbf{SE$^{\mathbf{r}}$}};
			\node (l5) at (c5) [above right=.6cm and .1cm, align=left, scale=1.15] {\hspace*{-.075cm}\tiny ABAG};
			\node (l6) at (c6) [above right=.35cm and .1cm, align=center, scale=1.15] {\hspace*{-.075cm}\tiny \textbf{EW$^{\mathbf{r}}$}};
			\node (l7) at (c7) [above right=.1cm and .1cm, align=center, scale=1.15] {\hspace*{-.075cm}\tiny DWM};
			\node (l8) at (c8) [above right=.1cm and .1cm, align=left, scale=1.15] {\hspace*{-.075cm}\tiny OBAG};
			\node (l9) at (c9) [above right=.85cm and .1cm, align=left, scale=1.15] {\hspace*{-.075cm}\tiny ADOB};
			\node (l10) at (c10) [above right=.6cm and .1cm, align=left, scale=1.15] {\hspace*{-.075cm}\tiny ALRV};
			\node (l11) at (c11) [above right=.35cm and .1cm, align=left, scale=1.15] {\hspace*{-.075cm}\tiny ALR};
			\node (l12) at (c12) [above right=.1cm and .1cm, align=left, scale=1.15] {\hspace*{-.075cm}\tiny ALS};
			\node (l13) at (c13) [above right=.1cm and .1cm, align=left, scale=1.15] {\hspace*{-.075cm}\tiny AUC/AWE};
			\node (l14) at (c14) [above right=.1cm and .1cm, align=left, scale=1.15] {\hspace*{-.075cm}\tiny LNSE};
			
			\fill[fill=gray,fill opacity=0.5] ({\sc*(5.37-\b)},-0.15) rectangle ({\sc*(5.37+2.0976-\b)},0.15);
			
			\foreach \x in {0,...,14} {
				\draw (l\x) -| (c\x);
			};
			\end{tikzpicture}
		\end{subfigure}\vspace*{-0.25cm}\subcaption{Average}}
	\caption{Bonferroni-Dunn test for different strategies and other classifiers on real data streams.}
	\label{fig:bonferroni_ext}
\end{figure}
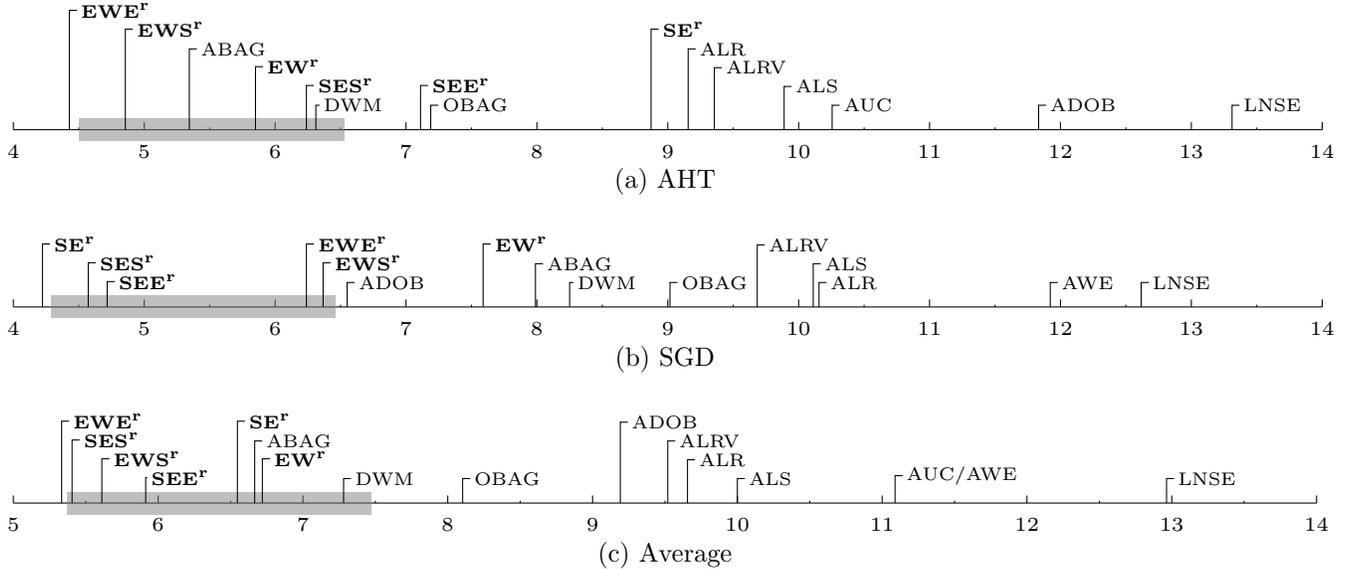

As the final part of our experimental study, we compared our best strategies (the risky ones) with some state-of-the-art classifiers, including models supported by different active learning strategies and popular ensembles. Based on the results in Tab. \ref{tab:final} and the statistical tests in Fig. \ref{fig:bonferroni_ext}, we can claim that our best strategies (in bold in the latter) are capable of significantly improving upon all baseline models using only active learning and upon most of the considered ensembles. We can distinguish EWE$^r$, EWS$^r$ and SES$^r$ methods. They outperformed all other classifiers on average, except for ABAG and DWM, which were significantly competitive, especially for AHT, even under strictly limited supervision. It exhibits their generally good resilience to such limitations. The EW-based ensembles provided the best results with AHT, while the SE-based ones dominated the highest ranks with SGD. This shows that our ensembles are able not only to improve upon single-classifier models without instance exploitation, but also to compete with state-of-the-art ensembles designed for data streams. One should also keep in mind, that while all the other committees use at least 10 base learners, our ensemble utilizes only two of them.

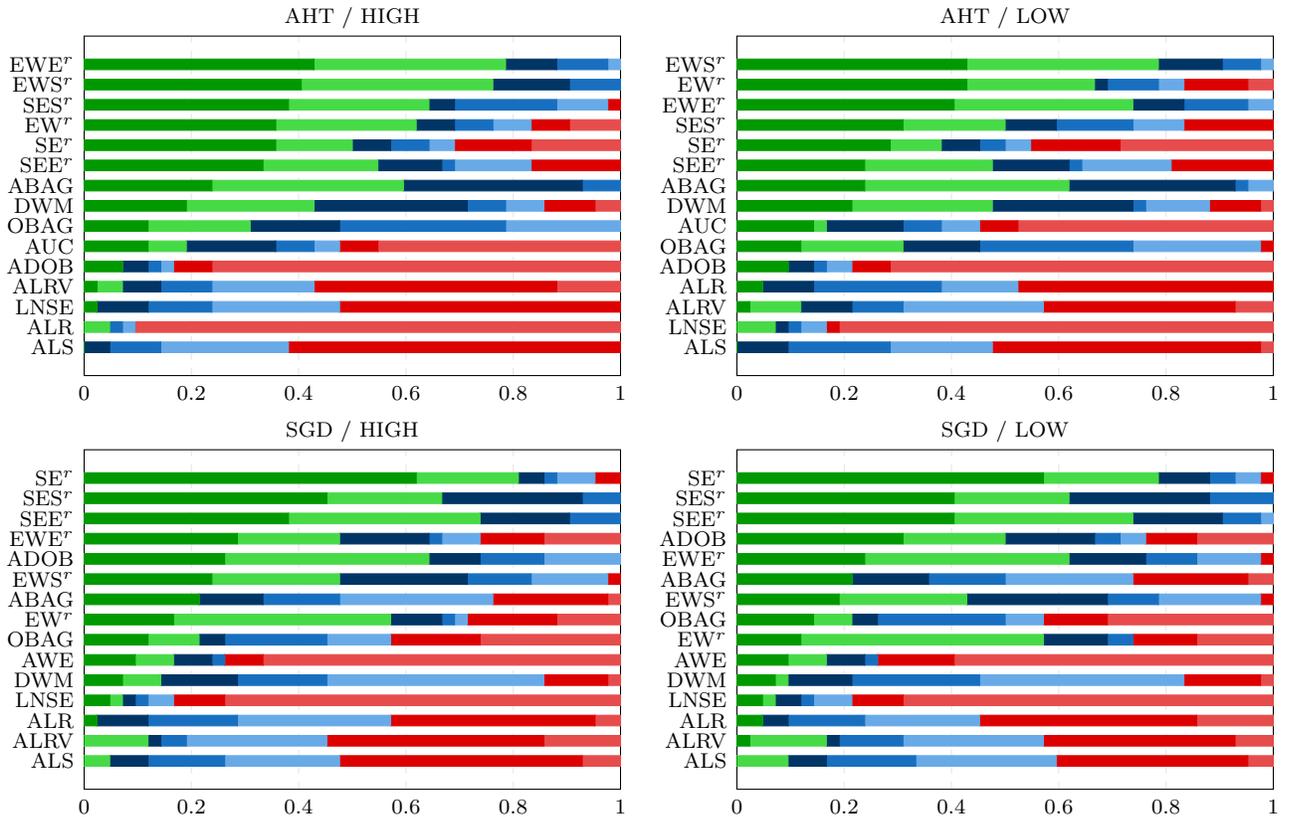
\begin{figure}[h]
	\centering
	\leftskip15pt\begin{subfigure}{.49\textwidth}
		\begin{tikzpicture}
		\begin{axis}[
		title={AHT / HIGH},
		title style={yshift=-1.5ex},
		width=\textwidth,
		height=0.7\textwidth,
		font=\scriptsize,
		xbar stacked,
		xmin=0, xmax=1,
		xmajorgrids=true,
		xlabel style={font=\tiny},
		xtick style={draw=none},
		grid style={dashed,white!90!black},
		table/header=false,
		table/row sep=\\,
		ylabel style={font=\tiny},
		ytick=data,
		yticklabels from table={
			ALS\\ALR\\LNSE\\ALRV\\ADOB\\AUC\\OBAG\\DWM\\ABAG\\SEE$^{r}$\\SE$^{r}$\\EW$^{r}$\\SES$^{r}$\\EWS$^{r}$\\EWE$^{r}$\\
		}{[index]0},
		]
		
		\addplot [
		xbar,
		bar width=4pt,
		fill=green2,
		draw=green2,
		] table[y expr=\coordindex,x index=0]{
			0.000000\\0.000000\\0.023810\\0.023810\\0.071429\\0.119048\\0.119048\\0.190476\\0.238095\\0.333333\\0.357143\\0.357143\\0.380952\\0.404762\\0.428571\\};
		
		\addplot [
		xbar,
		bar width=4pt,
		fill=green3,
		draw=green3,
		] table[y expr=\coordindex,x index=0]{
			0.000000\\0.047619\\0.000000\\0.047619\\0.000000\\0.071429\\0.190476\\0.238095\\0.357143\\0.214286\\0.142857\\0.261905\\0.261905\\0.357143\\0.357143\\};
		
		\addplot [
		xbar,
		bar width=4pt,
		fill=blue1,
		draw=blue1,
		] table[y expr=\coordindex,x index=0]{
			0.047619\\0.000000\\0.095238\\0.071429\\0.047619\\0.166667\\0.166667\\0.285714\\0.333333\\0.119048\\0.071429\\0.071429\\0.047619\\0.142857\\0.095238\\};
		
		\addplot [
		xbar,
		bar width=4pt,
		fill=blue2,
		draw=blue2,
		] table[y expr=\coordindex,x index=0]{
			0.095238\\0.023810\\0.119048\\0.095238\\0.023810\\0.071429\\0.309524\\0.071429\\0.071429\\0.023810\\0.071429\\0.071429\\0.190476\\0.095238\\0.095238\\};
		
		\addplot [
		xbar,
		bar width=4pt,
		fill=blue3,
		draw=blue3,
		] table[y expr=\coordindex,x index=0]{
			0.238095\\0.023810\\0.238095\\0.190476\\0.023810\\0.047619\\0.214286\\0.071429\\0.023810\\0.142857\\0.047619\\0.071429\\0.095238\\0.071429\\0.071429\\};
		
		\addplot [
		xbar,
		bar width=4pt,
		fill=red2,
		draw=red2,
		] table[y expr=\coordindex,x index=0]{
			0.619048\\0.000000\\0.571429\\0.452381\\0.071429\\0.071429\\0.071429\\0.095238\\0.047619\\0.238095\\0.142857\\0.071429\\0.071429\\0.000000\\0.023810\\};
		
		\addplot [
		xbar,
		bar width=4pt,
		fill=red3,
		draw=red3,
		] table[y expr=\coordindex,x index=0]{
			0.071429\\0.976190\\0.023810\\0.190476\\0.833333\\0.523810\\0.000000\\0.119048\\0.000000\\0.000000\\0.238095\\0.166667\\0.023810\\0.000000\\0.000000\\};
		
		\end{axis}
		\end{tikzpicture}%
		\hspace*{0.1cm}
		\begin{tikzpicture}
		\begin{axis}[
		title={AHT / LOW},
		title style={yshift=-1.5ex},
		width=\textwidth,
		height=0.7\textwidth,
		font=\scriptsize,
		xbar stacked,
		xmin=0, xmax=1,
		xmajorgrids=true,
		xlabel style={font=\tiny},
		xtick style={draw=none},
		grid style={dashed,white!90!black},
		table/header=false,
		table/row sep=\\,
		ylabel style={font=\tiny},
		ytick=data,
		yticklabels from table={
			ALS\\LNSE\\ALRV\\ALR\\ADOB\\OBAG\\AUC\\DWM\\ABAG\\SEE$^{r}$\\SE$^{r}$\\SES$^{r}$\\EWE$^{r}$\\EW$^{r}$\\EWS$^{r}$\\
		}{[index]0},
		]
		
		\addplot [
		xbar,
		bar width=4pt,
		fill=green2,
		draw=green2,
		] table[y expr=\coordindex,x index=0]{
			0.000000\\0.000000\\0.023810\\0.047619\\0.095238\\0.119048\\0.142857\\0.214286\\0.238095\\0.238095\\0.285714\\0.309524\\0.404762\\0.428571\\0.428571\\};
		
		\addplot [
		xbar,
		bar width=4pt,
		fill=green3,
		draw=green3,
		] table[y expr=\coordindex,x index=0]{
			0.000000\\0.071429\\0.095238\\0.000000\\0.000000\\0.190476\\0.023810\\0.261905\\0.380952\\0.238095\\0.095238\\0.190476\\0.333333\\0.238095\\0.357143\\};
		
		\addplot [
		xbar,
		bar width=4pt,
		fill=blue1,
		draw=blue1,
		] table[y expr=\coordindex,x index=0]{
			0.095238\\0.023810\\0.095238\\0.095238\\0.047619\\0.142857\\0.142857\\0.261905\\0.309524\\0.142857\\0.071429\\0.095238\\0.095238\\0.023810\\0.119048\\};
		
		\addplot [
		xbar,
		bar width=4pt,
		fill=blue2,
		draw=blue2,
		] table[y expr=\coordindex,x index=0]{
			0.190476\\0.023810\\0.095238\\0.238095\\0.023810\\0.285714\\0.071429\\0.023810\\0.023810\\0.023810\\0.047619\\0.142857\\0.119048\\0.095238\\0.071429\\};
		
		\addplot [
		xbar,
		bar width=4pt,
		fill=blue3,
		draw=blue3,
		] table[y expr=\coordindex,x index=0]{
			0.190476\\0.047619\\0.261905\\0.142857\\0.047619\\0.238095\\0.071429\\0.119048\\0.071429\\0.166667\\0.047619\\0.095238\\0.095238\\0.047619\\0.095238\\};
		
		\addplot [
		xbar,
		bar width=4pt,
		fill=red2,
		draw=red2,
		] table[y expr=\coordindex,x index=0]{
			0.500000\\0.023810\\0.357143\\0.476190\\0.071429\\0.095238\\0.071429\\0.095238\\0.047619\\0.238095\\0.166667\\0.214286\\0.023810\\0.119048\\0.000000\\};
		
		\addplot [
		xbar,
		bar width=4pt,
		fill=red3,
		draw=red3,
		] table[y expr=\coordindex,x index=0]{
			0.095238\\0.880952\\0.142857\\0.071429\\0.785714\\0.000000\\0.547619\\0.095238\\0.000000\\0.023810\\0.357143\\0.023810\\0.000000\\0.119048\\0.000000\\};
		
		\end{axis}
		\end{tikzpicture}
		\vspace*{0.5cm}
		\begin{tikzpicture}
		\begin{axis}[
		title={SGD / HIGH},
		title style={yshift=-1.5ex},
		width=\textwidth,
		height=0.7\textwidth,
		font=\scriptsize,
		xbar stacked,
		xmin=0, xmax=1,
		xmajorgrids=true,
		xlabel style={font=\tiny},
		xtick style={draw=none},
		grid style={dashed,white!90!black},
		table/header=false,
		table/row sep=\\,
		ylabel style={font=\tiny},
		ytick=data,
		yticklabels from table={
			ALS\\ALRV\\ALR\\LNSE\\DWM\\AWE\\OBAG\\EW$^{r}$\\ABAG\\EWS$^{r}$\\ADOB\\EWE$^{r}$\\SEE$^{r}$\\SES$^{r}$\\SE$^{r}$\\
		}{[index]0},
		]
		
		\addplot [
		xbar,
		bar width=4pt,
		fill=green2,
		draw=green2,
		] table[y expr=\coordindex,x index=0]{
			0.000000\\0.000000\\0.023810\\0.047619\\0.071429\\0.095238\\0.119048\\0.166667\\0.214286\\0.238095\\0.261905\\0.285714\\0.380952\\0.452381\\0.619048\\};
		
		\addplot [
		xbar,
		bar width=4pt,
		fill=green3,
		draw=green3,
		] table[y expr=\coordindex,x index=0]{
			0.047619\\0.119048\\0.000000\\0.023810\\0.071429\\0.071429\\0.095238\\0.404762\\0.000000\\0.238095\\0.380952\\0.190476\\0.357143\\0.214286\\0.190476\\};
		
		\addplot [
		xbar,
		bar width=4pt,
		fill=blue1,
		draw=blue1,
		] table[y expr=\coordindex,x index=0]{
			0.071429\\0.023810\\0.095238\\0.023810\\0.142857\\0.071429\\0.047619\\0.095238\\0.119048\\0.238095\\0.095238\\0.166667\\0.166667\\0.261905\\0.047619\\};
		
		\addplot [
		xbar,
		bar width=4pt,
		fill=blue2,
		draw=blue2,
		] table[y expr=\coordindex,x index=0]{
			0.142857\\0.047619\\0.166667\\0.023810\\0.166667\\0.023810\\0.190476\\0.023810\\0.142857\\0.119048\\0.119048\\0.023810\\0.095238\\0.095238\\0.023810\\};
		
		\addplot [
		xbar,
		bar width=4pt,
		fill=blue3,
		draw=blue3,
		] table[y expr=\coordindex,x index=0]{
			0.214286\\0.261905\\0.285714\\0.047619\\0.404762\\0.000000\\0.119048\\0.023810\\0.285714\\0.142857\\0.166667\\0.071429\\0.047619\\0.047619\\0.071429\\};
		
		\addplot [
		xbar,
		bar width=4pt,
		fill=red2,
		draw=red2,
		] table[y expr=\coordindex,x index=0]{
			0.452381\\0.404762\\0.380952\\0.095238\\0.119048\\0.071429\\0.166667\\0.166667\\0.214286\\0.095238\\0.047619\\0.119048\\0.023810\\0.000000\\0.071429\\};
		
		\addplot [
		xbar,
		bar width=4pt,
		fill=red3,
		draw=red3,
		] table[y expr=\coordindex,x index=0]{
			0.142857\\0.214286\\0.119048\\0.809524\\0.095238\\0.738095\\0.333333\\0.190476\\0.095238\\0.000000\\0.000000\\0.214286\\0.000000\\0.000000\\0.047619\\};
		
		\end{axis}
		\end{tikzpicture}%
		\hspace*{0.1cm}
		\begin{tikzpicture}
		\begin{axis}[
		title={SGD / LOW},
		title style={yshift=-1.5ex},
		width=\textwidth,
		height=0.7\textwidth,
		font=\scriptsize,
		xbar stacked,
		xmin=0, xmax=1,
		xmajorgrids=true,
		xlabel style={font=\tiny},
		xtick style={draw=none},
		grid style={dashed,white!90!black},
		table/header=false,
		table/row sep=\\,
		ylabel style={font=\tiny},
		ytick=data,
		yticklabels from table={
			ALS\\ALRV\\ALR\\LNSE\\DWM\\AWE\\EW$^{r}$\\OBAG\\EWS$^{r}$\\ABAG\\EWE$^{r}$\\ADOB\\SEE$^{r}$\\SES$^{r}$\\SE$^{r}$\\
		}{[index]0},
		]
		
		\addplot [
		xbar,
		bar width=4pt,
		fill=green2,
		draw=green2,
		] table[y expr=\coordindex,x index=0]{
			0.000000\\0.023810\\0.047619\\0.047619\\0.071429\\0.095238\\0.119048\\0.142857\\0.190476\\0.214286\\0.238095\\0.309524\\0.404762\\0.404762\\0.571429\\};
		
		\addplot [
		xbar,
		bar width=4pt,
		fill=green3,
		draw=green3,
		] table[y expr=\coordindex,x index=0]{
			0.095238\\0.142857\\0.000000\\0.023810\\0.023810\\0.071429\\0.452381\\0.071429\\0.238095\\0.000000\\0.380952\\0.190476\\0.333333\\0.214286\\0.214286\\};
		
		\addplot [
		xbar,
		bar width=4pt,
		fill=blue1,
		draw=blue1,
		] table[y expr=\coordindex,x index=0]{
			0.071429\\0.023810\\0.047619\\0.047619\\0.119048\\0.071429\\0.119048\\0.047619\\0.261905\\0.142857\\0.142857\\0.166667\\0.166667\\0.261905\\0.095238\\};
		
		\addplot [
		xbar,
		bar width=4pt,
		fill=blue2,
		draw=blue2,
		] table[y expr=\coordindex,x index=0]{
			0.166667\\0.119048\\0.142857\\0.023810\\0.238095\\0.023810\\0.047619\\0.238095\\0.095238\\0.142857\\0.095238\\0.047619\\0.071429\\0.119048\\0.047619\\};
		
		\addplot [
		xbar,
		bar width=4pt,
		fill=blue3,
		draw=blue3,
		] table[y expr=\coordindex,x index=0]{
			0.261905\\0.261905\\0.214286\\0.071429\\0.380952\\0.000000\\0.000000\\0.071429\\0.190476\\0.238095\\0.119048\\0.047619\\0.071429\\0.071429\\0.047619\\};
		
		\addplot [
		xbar,
		bar width=4pt,
		fill=red2,
		draw=red2,
		] table[y expr=\coordindex,x index=0]{
			0.357143\\0.357143\\0.404762\\0.095238\\0.142857\\0.142857\\0.119048\\0.119048\\0.095238\\0.214286\\0.095238\\0.095238\\0.023810\\0.000000\\0.023810\\};
		
		\addplot [
		xbar,
		bar width=4pt,
		fill=red3,
		draw=red3,
		] table[y expr=\coordindex,x index=0]{
			0.119048\\0.142857\\0.214286\\0.761905\\0.095238\\0.666667\\0.214286\\0.380952\\0.000000\\0.119048\\0.000000\\0.214286\\0.000000\\0.000000\\0.071429\\};
		
		\end{axis}
		\end{tikzpicture}
	\end{subfigure}
	\vspace*{-5mm}
	\caption{Ranks for AHT and SGD using different strategies compared with other classifiers on real data streams.}
	\label{fig:ranks}
\end{figure}

The complementary results presented in Fig. \ref{fig:ranks} show the frequencies of occupied ranks for each of the considered methods given high ($B\geq20\%$) and low budgets ($B\leq10\%$). Unsurprisingly, we can see that the best strategies were most frequently the best ones (green bars), while very rarely ending up as the worst ones (red bars), regardless of the available supervision. One interesting observation is that our ensembles were able to maintain safer lower bounds also for the presented ranks -- they very efficiently reduced the number of lowest ranks compared with their simpler counterparts (SE$^r$, EW$^r$), bringing them down to zero for the best strategies (EWS$^r$ and EWE$^r$ for AHT, SES$^r$ and SEE$^r$ for SGD). It is important if someone cares about the worst possible scenario.

\begin{figure}[H]
	\centering
	\hspace*{-8pt}
	\begin{subfigure}{0.32\linewidth}
		\centering
		\includegraphics[width=\linewidth]{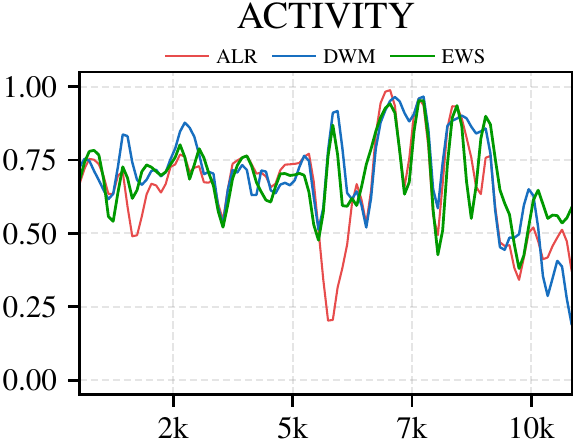}
	\end{subfigure}
	\hspace*{2pt}
	\begin{subfigure}{0.32\linewidth}
		\centering
		\includegraphics[width=\linewidth]{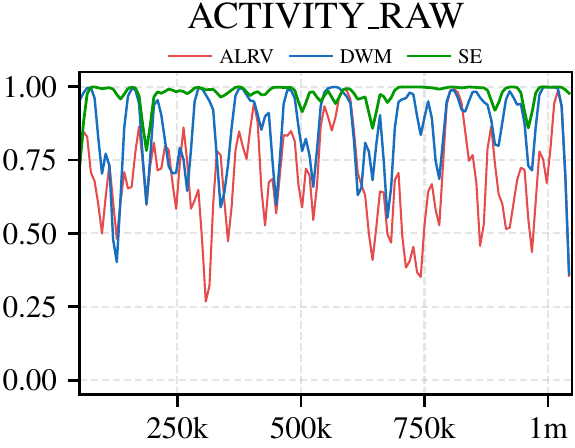}
	\end{subfigure}
	\hspace*{2pt}
	\begin{subfigure}{0.32\linewidth}
		\centering
		\includegraphics[width=\linewidth]{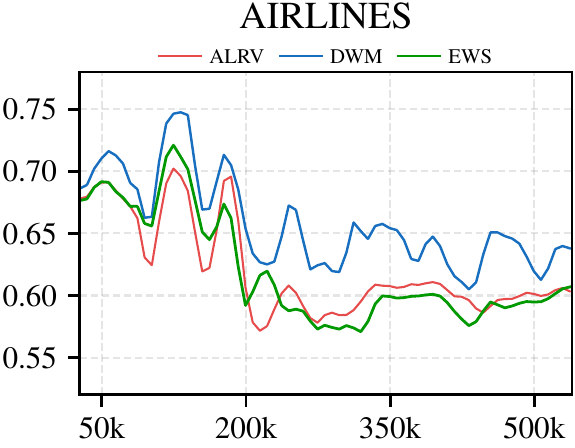}
	\end{subfigure}\\
	\vspace*{8pt}
	\hspace*{-8pt}
	\begin{subfigure}{0.32\linewidth}
		\centering
		\includegraphics[width=\linewidth]{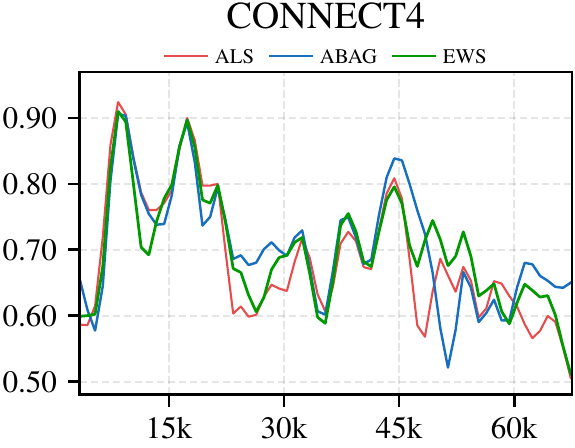}
	\end{subfigure}
	\hspace*{2pt}
	\begin{subfigure}{0.32\linewidth}
		\centering
		\includegraphics[width=\linewidth]{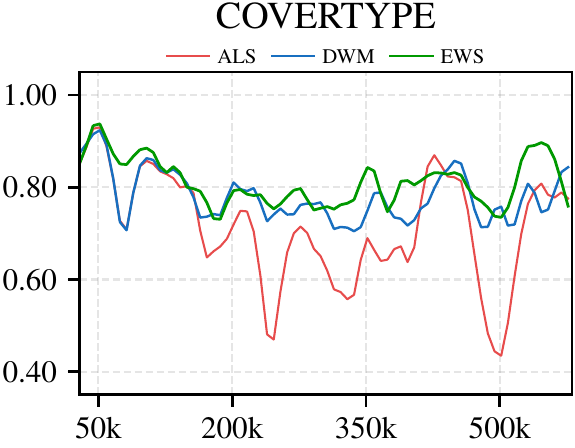}
	\end{subfigure}
	\hspace*{2pt}
	\begin{subfigure}{0.32\linewidth}
		\centering
		\includegraphics[width=\linewidth]{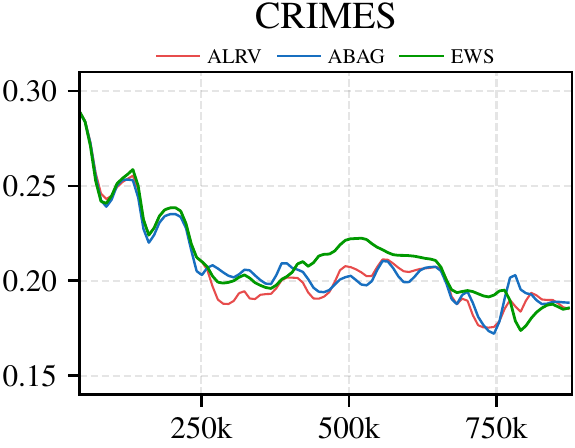}
	\end{subfigure}\\
	\vspace*{8pt}
	\hspace*{-8pt}
	\begin{subfigure}{0.32\linewidth}
		\centering
		\includegraphics[width=\linewidth]{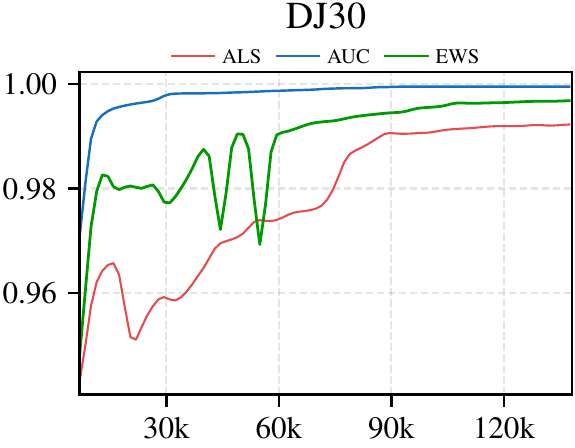}
	\end{subfigure}
	\hspace*{2pt}
	\begin{subfigure}{0.32\linewidth}
		\centering
		\includegraphics[width=\linewidth]{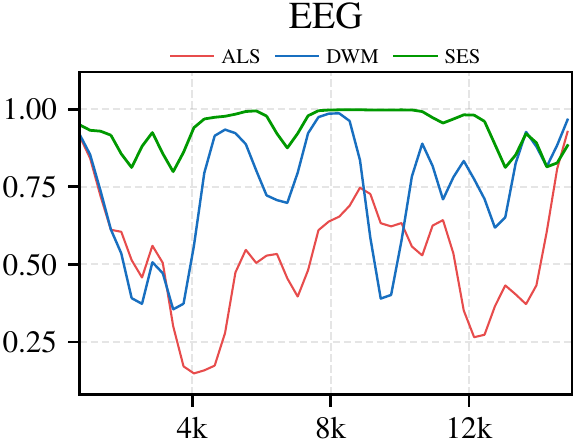}
	\end{subfigure}
	\hspace*{2pt}
	\begin{subfigure}{0.32\linewidth}
		\centering
		\includegraphics[width=\linewidth]{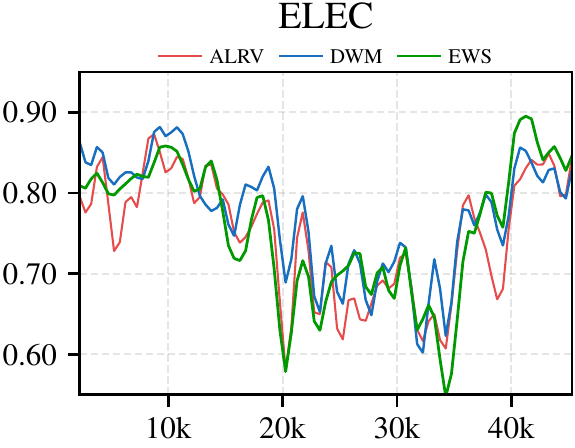}
	\end{subfigure}\\
	\vspace*{8pt}
	\hspace*{-8pt}
	\begin{subfigure}{0.32\linewidth}
		\centering
		\includegraphics[width=\linewidth]{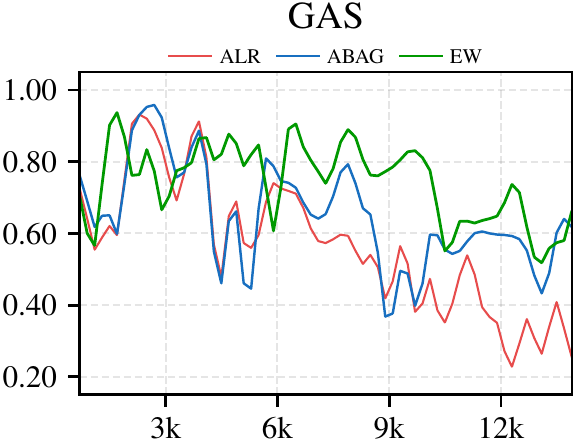}
	\end{subfigure}
	\hspace*{2pt}
	\begin{subfigure}{0.32\linewidth}
		\centering
		\includegraphics[width=\linewidth]{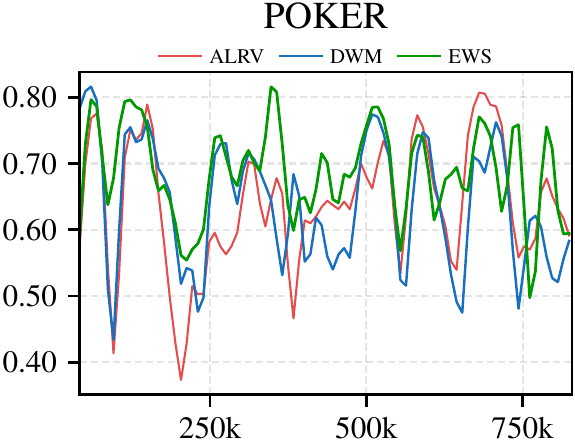}
	\end{subfigure}
	\hspace*{2pt}
	\begin{subfigure}{0.32\linewidth}
		\centering
		\includegraphics[width=\linewidth]{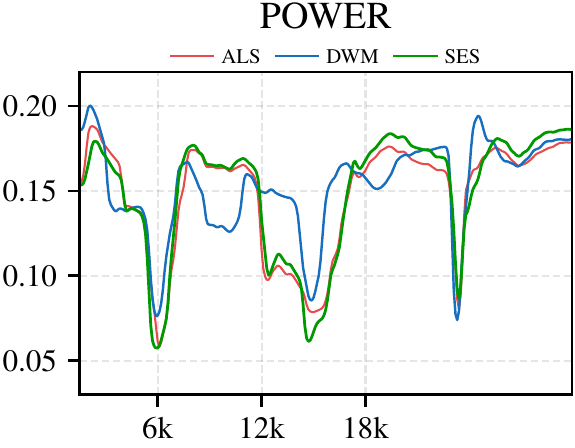}
	\end{subfigure}\\
	\vspace*{8pt}
	\hspace*{-8pt}
	\begin{subfigure}{0.32\linewidth}
		\centering
		\includegraphics[width=\linewidth]{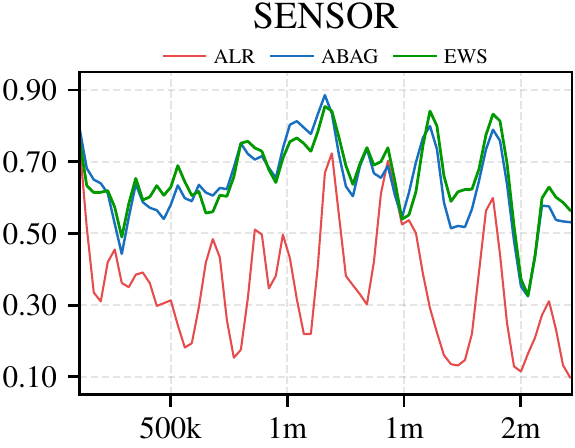}
	\end{subfigure}
	\hspace*{2pt}
	\begin{subfigure}{0.32\linewidth}
		\centering
		\includegraphics[width=\linewidth]{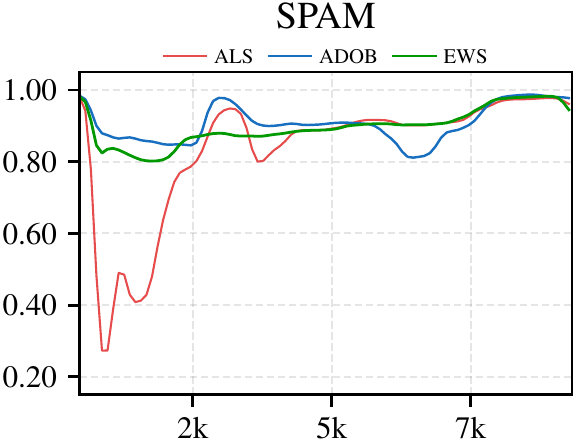}
	\end{subfigure}
	\hspace*{2pt}
	\begin{subfigure}{0.32\linewidth}
		\centering
		\includegraphics[width=\linewidth]{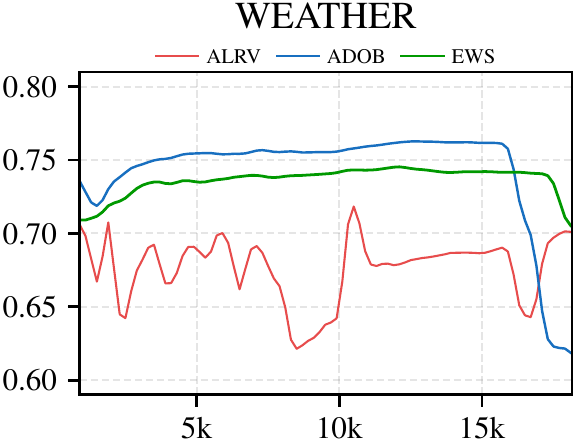}
	\end{subfigure}\\
	\caption{Accuracy series for AHT given $B=10\%$: the best strategy vs. the best AL and the best (not our) ensemble.}
	\label{fig:aht_series}
\end{figure}

\begin{figure}[H]
	\centering
	\hspace*{-8pt}
	\begin{subfigure}{0.32\linewidth}
		\centering
		\includegraphics[width=\linewidth]{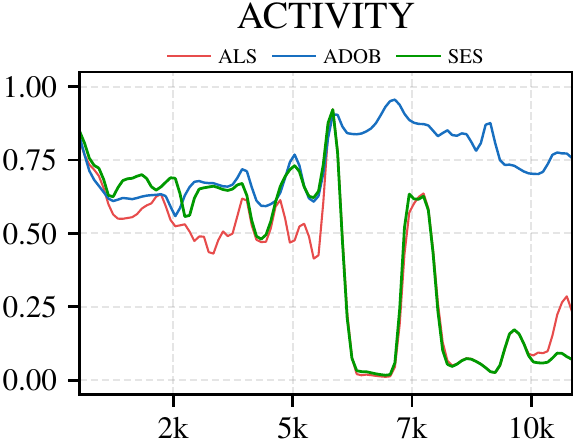}
	\end{subfigure}
	\hspace*{2pt}
	\begin{subfigure}{0.32\linewidth}
		\centering
		\includegraphics[width=\linewidth]{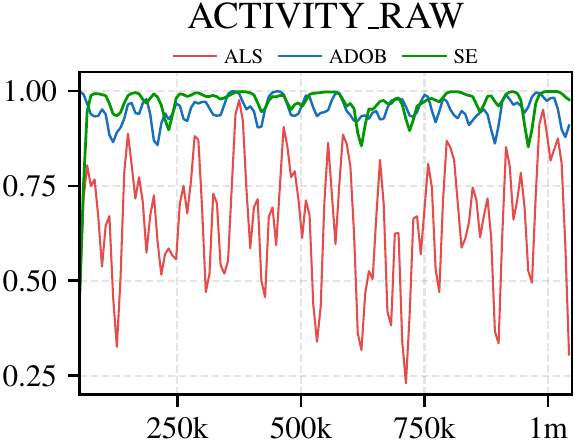}
	\end{subfigure}
	\hspace*{2pt}
	\begin{subfigure}{0.32\linewidth}
		\centering
		\includegraphics[width=\linewidth]{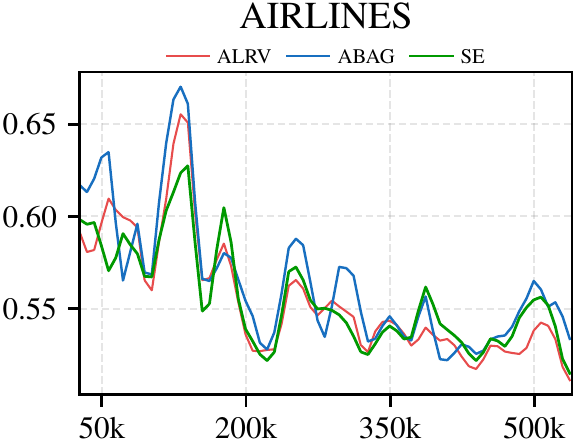}
	\end{subfigure}\\
	\vspace*{8pt}
	\hspace*{-8pt}
	\begin{subfigure}{0.32\linewidth}
		\centering
		\includegraphics[width=\linewidth]{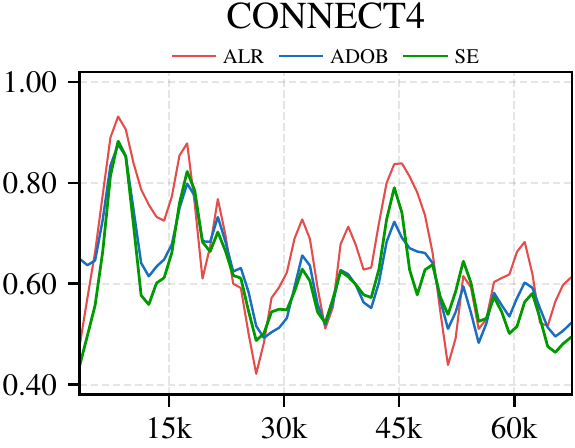}
	\end{subfigure}
	\hspace*{2pt}
	\begin{subfigure}{0.32\linewidth}
		\centering
		\includegraphics[width=\linewidth]{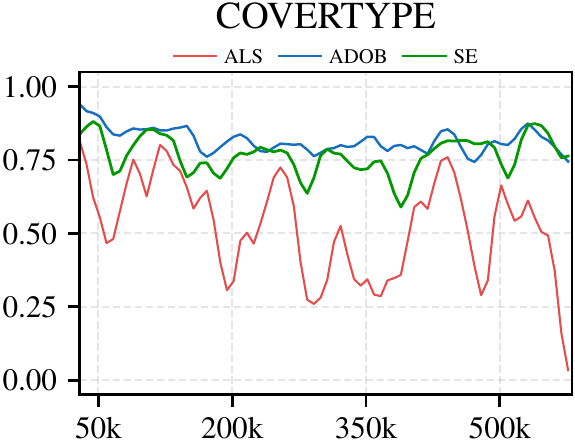}
	\end{subfigure}
	\hspace*{2pt}
	\begin{subfigure}{0.32\linewidth}
		\centering
		\includegraphics[width=\linewidth]{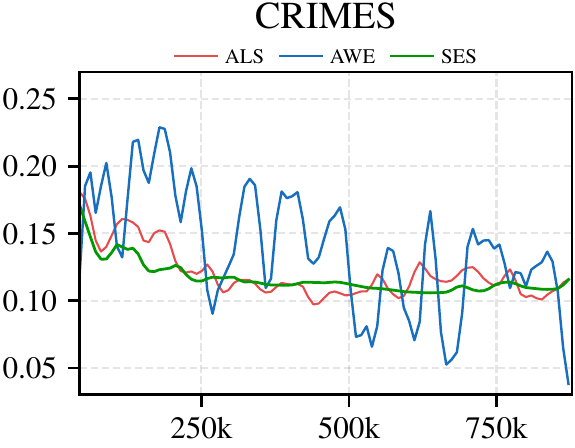}
	\end{subfigure}\\
	\vspace*{8pt}
	\hspace*{-8pt}
	\begin{subfigure}{0.32\linewidth}
		\centering
		\includegraphics[width=\linewidth]{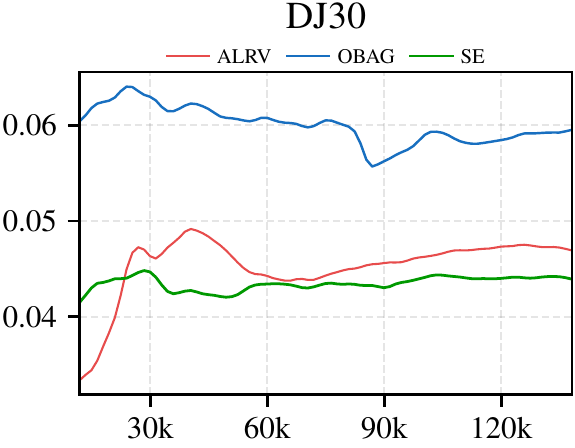}
	\end{subfigure}
	\hspace*{2pt}
	\begin{subfigure}{0.32\linewidth}
		\centering
		\includegraphics[width=\linewidth]{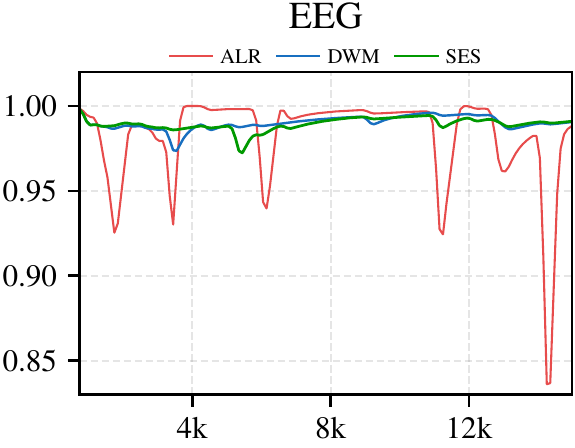}
	\end{subfigure}
	\hspace*{2pt}
	\begin{subfigure}{0.32\linewidth}
		\centering
		\includegraphics[width=\linewidth]{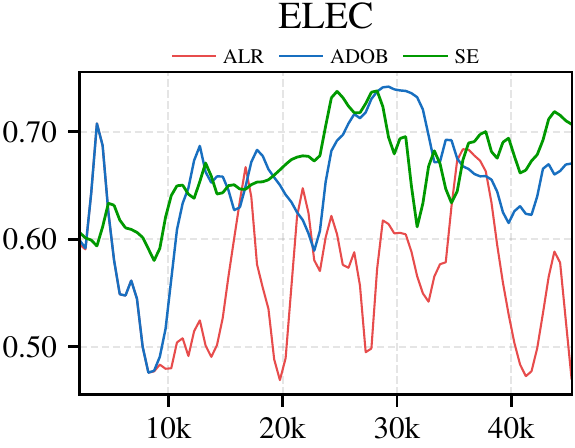}
	\end{subfigure}\\
	\vspace*{8pt}
	\hspace*{-8pt}
	\begin{subfigure}{0.32\linewidth}
		\centering
		\includegraphics[width=\linewidth]{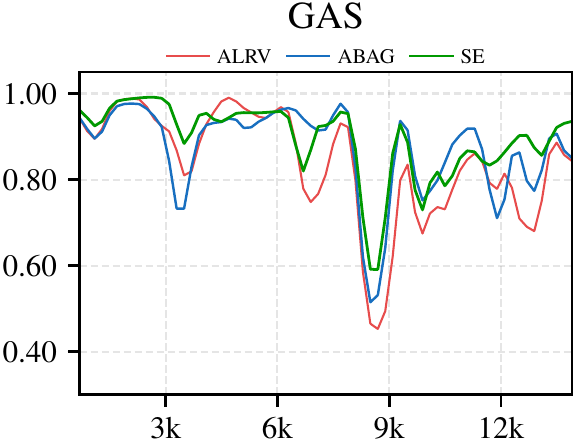}
	\end{subfigure}
	\hspace*{2pt}
	\begin{subfigure}{0.32\linewidth}
		\centering
		\includegraphics[width=\linewidth]{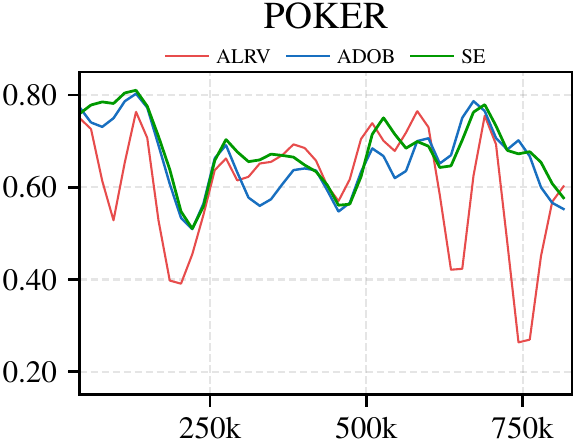}
	\end{subfigure}
	\hspace*{2pt}
	\begin{subfigure}{0.32\linewidth}
		\centering
		\includegraphics[width=\linewidth]{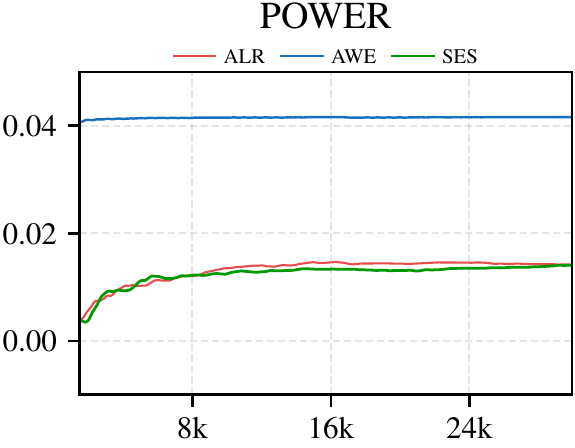}
	\end{subfigure}\\
	\vspace*{8pt}
	\hspace*{-8pt}
	\begin{subfigure}{0.32\linewidth}
		\centering
		\includegraphics[width=\linewidth]{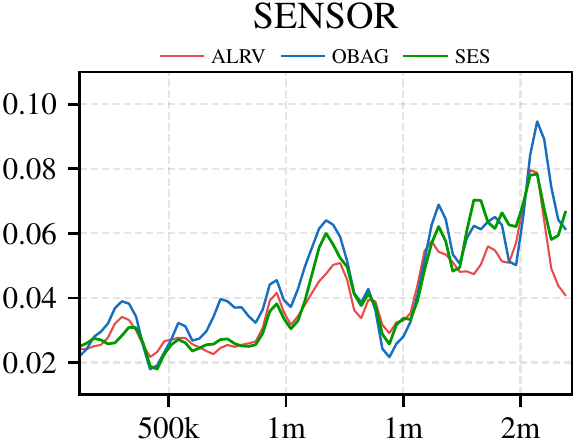}
	\end{subfigure}
	\hspace*{2pt}
	\begin{subfigure}{0.32\linewidth}
		\centering
		\includegraphics[width=\linewidth]{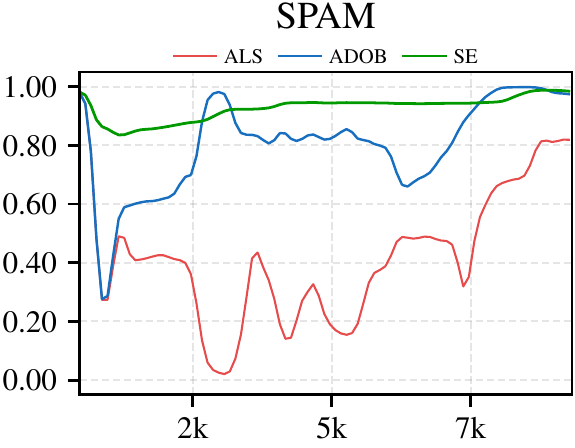}
	\end{subfigure}
	\hspace*{2pt}
	\begin{subfigure}{0.32\linewidth}
		\centering
		\includegraphics[width=\linewidth]{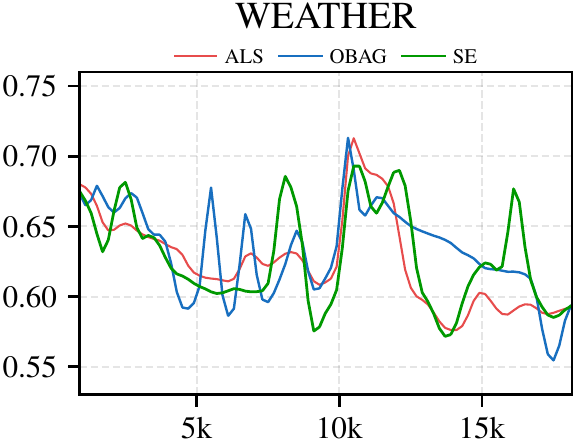}
	\end{subfigure}\\
	\caption{Accuracy series for SGD given $B=10\%$: the best strategy vs. the best AL and the best (not our) ensemble.}
	\label{fig:sgd_series}
\end{figure}

In order to provide the reader with some specific examples of how our methods work in practice, in Fig.\ref{fig:aht_series} and Fig.\ref{fig:sgd_series} we presented accuracy series (instead of kappa for readability) registered for all real streams when the budget was reasonably limited to $B=10\%$. The best exploiting strategy, the best baseline using active learning and the best other ensemble were chosen for each stream separately. For most of the streams, we can easily notice that instance exploitation (green) was able to elevate the temporal performance of standard models that were reliant solely on active learning (red). The improvements may have diverse character. For example, our strategies were able to provide overall a more stable, saturated and higher level of predictive performance for Activity-Raw, Covertype, EGG, Sensor or Weather, among others, for both AHT and SGD. For other streams, the enhanced adaption meant that instance exploitation was able to alleviate severe temporal drops in accuracy, most likely due to concept drifts, for example, for Gas, Poker or Spam. There were very few exceptions when none of our strategies was able to provide improvements, like Airlines and DJ30, or Crimes for SGD. Finally, when compared with the considered state-of-the-art ensembles (blue), we can also see that our strategies were, in most cases, competitive in improving adaptation in the same way as against the baseline models.

\subsubsection{Conclusions}

\smallskip
To summarize all the presented results and observations, we finalize our study with a list of the most important conclusions. Primarily, they address the research questions introduced at the beginning of the experimental study.

\smallskip
\begin{itemize}
	
	\item \textbf{Instance exploitation improves learning from drifting streams on a budget (RQ1).} In most of the considered cases, almost all strategies provided significant improvements over the baseline using only active learning and adapting in a conservative way. Without more intensive learning, the base classifiers were not able to deal with strict supervision limitations and drifts at the same time, even if supported by a more strategic label query. They struggled with providing reactive up-to-date models for dynamic temporary concepts and end up with unresolved and severe underfitting. Our method addressed this problem by forcing the classifiers to take the risk of exploiting only the labeled instances they could use and enhancing the adaptation process as a result. The improvements were more impactful for real streams. With the abundance of positive examples collected from diverse measurements and statistical tests, and supported by the fact that generally higher intensity of exploitation was preferred, we prove that the problem of underfitting is critical for realistic streaming scenarios and that risking overfitting by using instance exploitation is more reasonable in practice.
	
	\item \textbf{No free lunch theorem for strategies and settings (RQ2).} We were not able to explicitly select one of the three strategies (UW/EW, SE) as the best one. While EW$^r$ worked best with AHT, SGD exhibited the best synergy with SE$^r$. Furthermore, although all the strategies generally worked better with the more risky settings using higher values of intensity $\lambda_{max}$, we were forced to distinguish different values of this parameter for different classifiers. Also, AHT preferred larger sliding windows than SGD. The distinction was caused by the fact that some algorithms may be more reactive to changes (AHT) than others (SGD), therefore, they may need different amounts of instance exploitation.
	
	\item \textbf{Dynamic control comes with a reasonable trade-off (RQ2).} The heuristics for controlling intensity and window size in a dynamic way do not provide improvements in predictive performance in most cases, even impairing it to some limited extent. On the other hand, they are able to reduce running time and memory consumption, respectively. For the latter, adjusting the window size based on ADWIN is a contradictory exception from the observations -- this algorithm provides the best quality of classification for AHT and reliable one for SGD, while requiring more memory than most of the other windows. We would recommend using fixed intensity $\lambda_{max}$ and dynamic window size based on ADWIN ($\omega_{max}=\omega_{ADW}$), if someone does not have problems with utilization of resources. Otherwise, we suggest using dynamic intensity (Eq. \ref{eq:lambda}) and dynamic window (Eq. \ref{eq:omega}) with a limited size $\omega_{max}=1000$, while utilizing ADWIN-based error as a control signal $\epsilon=\epsilon_{ADW}$.
	
	\item \textbf{Switching and elevating alleviates the risk of overfitting (RQ3).} The ensemble-based techniques for avoiding turning one problem (underfitting) into another (overfitting) provided the assumed improvements. The methods are able to guarantee a more reliable lower bound on performance. Since we can efficiently track errors even under limited supervision, the ensembles can correctly determine a temporarily better approach. As a consequence in a prevalent number of cases (except for SGD using SE$^r$) they are at least as good as the standard or risky learner, providing additional enhancements in many scenarios. One should, however, keep in mind that the advantage of using the ensembles is usually on the brink of significance. Furthermore, we did not observed a significant difference between switching and elevating, therefore, since the latter requires very time-consuming model replacements, we recommend using switching. Finally, taking into account all considered factors, we can distinguish EWS and SES as the best methods on average.
	
	\item \textbf{Proposed strategies are competitive against other streaming classifiers (RQ4).} The final comparison revealed that our algorithms, mainly committees, not only improve upon the baseline using only active learning, but that they can also be at least competitive against some of the state-of-the-art streaming ensembles, including bagging and boosting, and outperform them in many cases. At the same time, while other ensembles utilize 10 base learners, ours use only two of them. Finally, although we could extend this comparison to more ensembles, one should notice that this study was not entirely focused on the ensemble-based methods. In fact, there is a lot of potential for further improvements in this direction. We conducted only the limited comparison to place our algorithms in some recognizable context.
	
\end{itemize}

\section{Summary}
\label{sec:sum}

In this paper, we have addressed a challenging, yet crucial issue in the data stream mining domain -- how to increase the adaptation capabilities of classifiers under concept drift while dealing with very limited access to class labels. Sparsely labeled data streams are predominant in a plethora of real-world applications and a lack of labeled instances increases the already high difficulty of recovery from changes. We have proposed a novel and flexible framework for instance exploitation in order to boost the learning from streaming data on a budget. By using a sliding window with a probabilistic sampling for selecting instances for additional exposure to the online classifier we were able to force faster adaptation rates, resulting in a significantly improved robustness to concept drift. We developed three instance exploitation strategies to alleviate the problem of underfitting of classifiers to new emerging concepts by various levels of expositions of labeled instanced obtained from active learning. In order to minimize the risk of overfitting, we have proposed two ensemble architectures that dynamically switch between learners based on aggressive and standard instance exposition. All our strategies are incorporated in a flexible wrapper framework capable of working with any active learning strategy and online classifier. 

Based on extensive experimental study, we were able to show the tremendous gains of using our strategies while learning temporary concepts from data streams and having limited access to class labels. The analysis of the results allowed us to reach in-depth and unique insights into the adaptation of classifiers under concept drift, showing how access (or lack of thereof) to an adequate number of labeled instances is of crucial importance in the dynamic settings. We formed a set of recommendations on how and when to use the proposed strategies in order to improve learning rates from sparsely labeled non-stationary data streams at no additional cost. 

Our future works will concentrate on using instance exploitation to leverage ensemble learning and control ensemble diversity under concept drift, as well as to improve learning from sparsely labeled imbalanced data streams.

\section*{References}

\bibliographystyle{elsarticle-num}
\bibliography{refs}

\end{document}